\documentclass{article}

\usepackage{iclr2026_conference}
\iclrfinalcopy   

\usepackage[T1]{fontenc}
\usepackage[utf8]{inputenc}
\usepackage{times}      
\usepackage{microtype}

\usepackage{graphicx}
\usepackage{booktabs}
\usepackage{amsmath}
\usepackage{amssymb}
\usepackage{multirow}
\usepackage{xcolor}
\usepackage{caption}
\usepackage[section]{placeins}
\usepackage{float}
\usepackage[htt]{hyphenat}
\usepackage{hyperref}
\usepackage{url}



\graphicspath{{figures/}}

\title{Analyzing Quality-Latency-Resource Trade-offs in a Technical
Documentation RAG Assistant Using LoRA Adaptation}

\author{Evgenii Palnikov$^{1}$, Elizaveta Gavrilova$^{1}$ \\[6pt]
\normalfont $^{1}$HSE University
}

\begin{document}
\maketitle
\fancyhead[L]{\NoHyper Preprint\endNoHyper}
\fancyfoot[C]{\NoHyper\thepage\endNoHyper}
\thispagestyle{fancy}

\begin{abstract}
We study quality--latency--resource trade-offs in a documentation-grounded
retrieval-augmented generation (RAG) system that uses Low-Rank Adaptation
(LoRA) of the generator. We build a manually verified benchmark of
5{,}144~question--answer pairs over the official Kubernetes documentation and
combine it with a fixed hybrid-retrieval pipeline (BGE-M3 dense, BGE-M3 native
sparse, Reciprocal Rank Fusion, cross-encoder reranking). Over this benchmark we
ablate 20~LoRA configurations on \texttt{Llama-3.2-3B-Instruct} and
\texttt{Llama-3.1-8B-Instruct} across rank and target-module choices, and
evaluate each on token-level F1, LLM-judged groundedness and correctness
(pass@4), inference latency, inference memory, and training cost, all reported
with bootstrap 95\% confidence intervals. Pareto analysis shows that LoRA
adapters acting only on the q and v attention projections consistently
dominate the front, while the 3B/8B choice mainly defines operating regime.
A param-matched control comparison further indicates that the q/v advantage
is structural rather than purely parametric. The benchmark, selected adapters,
and code are available at \url{https://github.com/EugPal/rag-lora-tradeoffs}.

\end{abstract}

\section{Introduction}
\label{sec:intro}

Retrieval-augmented generation (RAG) has become a standard way to ground
large language models (LLMs) in external corpora when their parametric
memory is insufficient or stale \citep{lewis2020rag,gao2023survey,wu2024survey}.
For \emph{documentation-oriented} assistants -- where users expect answers
backed by specific configuration flags, API versions, or feature gates --
the bar is even higher: a deployable system must combine factual accuracy
and visible support from the retrieved context with acceptable latency,
inference memory, and training cost. In practice these objectives are in
direct tension. Larger generators improve answer quality but increase
inference memory and latency; richer retrieval and reranking add overhead;
more expressive LoRA \citep{hu2022lora} adapters change the cost of
training. Picking a working point therefore requires reasoning about a
\emph{multi-objective} trade-off rather than a single quality metric.

Existing work largely studies these dimensions in isolation. RAG surveys
focus on retrieval/fusion/reranking design choices
\citep{gao2023survey,yu2024evaluation}; parameter-efficient fine-tuning
surveys focus on the LoRA family in general purpose tasks
\citep{han2024peft,wang2025peft}; RAG-evaluation work focuses on quality
and groundedness on standard QA sets \citep{es2024ragas,yu2024evaluation}.
A few recent papers combine RAG with LoRA, but they typically treat LoRA
as a single technique -- a single rank, a single target-module set -- and
compare it against alternatives like DoRA or full fine-tuning
\citep{tahir2024jora,zhao2024ramole,baqar2025raglora}. This leaves an open
question: \emph{how does the choice of LoRA configuration -- rank, target
modules, base-model size -- interact with retrieval quality, latency, and
training cost when the rest of the RAG pipeline is held fixed?}

We address this question by constructing a documentation-grounded RAG
benchmark and analysing a structured space of LoRA configurations on it. Our
benchmark consists of $5{,}144$ manually verified question--answer pairs
over the official Kubernetes documentation, paired with a fixed hybrid
retrieval pipeline (BGE-M3 \citep{chen2024m3embedding} dense + BGE-M3 native
sparse + Reciprocal Rank Fusion \mbox{\citep{cormack2009rrf}} + a cross-encoder
reranker). Over this pipeline we
ablate 20~LoRA configurations on \texttt{Llama-3.2-3B-Instruct} and
\texttt{Llama-3.1-8B-Instruct} (5 ranks $\times$ 2 target-module sets per
base model) against the corresponding non-adapted baselines. Each
configuration is evaluated on (i)~token-level F1, (ii)~LLM-as-a-judge
groundedness and correctness (pass@4), (iii)~inference latency,
(iv)~inference VRAM, and (v)~training time and VRAM, with bootstrap 95\%
confidence intervals on all point estimates and paired bootstrap on key
$\Delta F_1$ comparisons. To probe robustness, we re-run the full grid under
10~retrieval/prompting ablation regimes.

Three findings emerge. First, on every Pareto front we examine --
F1-vs-latency, F1-vs-inference-memory, and F1-vs-training-cost --
non-dominated points are dominated by LoRA adapters that act only on the
$q$ and $v$ attention projections. \texttt{full\_attention} adapters
($q,k,v,o$) never appear on the training Pareto front and win at most
$2/10$ ablation regimes on groundedness. Second, the 3B/8B choice mainly
determines the operating regime rather than the achievable quality: the
strongest LoRA-adapted 3B configuration is statistically indistinguishable
from the unadapted 8B baseline on $F_1$ ($\Delta F_1\in[-0.021,+0.026]$),
while costing roughly $9$\,GB less inference VRAM. Third, a \emph{param-matched}
control comparison -- in which the total number of trainable LoRA
parameters is held constant across the $q/v$ and full-attention schemes --
shows that the $q/v$ advantage is structural and not a consequence of having
more parameters.

Our contributions are: (i) a manually verified, license-clean RAG
benchmark over Kubernetes documentation, designed for multi-criteria
evaluation; (ii) a systematic LoRA-configuration study within a fixed
hybrid-retrieval pipeline, with statistical inference on all reported
trade-offs; (iii) a param-matched control comparison that isolates the
structural effect of the target-module choice from the effect of parameter
count; (iv) the released benchmark and selected adapters, intended as a
reproducible starting point for follow-up work on retrieval-augmented
fine-tuning of documentation assistants.

\section{Related Work}
\label{sec:related}

\paragraph{RAG and retrieval components.}
Retrieval-augmented generation \citep{lewis2020rag} has been thoroughly
surveyed in recent years \citep{gao2023survey,wu2024survey}: modern RAG is
not a single architecture but a family of designs that differ in how
retrieval, fusion, reranking, and context use are organised. On the
retrieval side, dense passage retrieval was canonicalised by
\citet{karpukhin2020dpr}, classical sparse retrieval by BM25
\citep{robertson1994bm25, robertson1995okapi, robertson2009bm25}, and
training-aware extensions such as RocketQA \citep{qu2021rocketqa} and
ColBERTv2 \citep{santhanam2022colbertv2} show that retrieval quality
depends as much on the training scheme as on the index type. Recent
multi-functional embedders -- in particular BGE-M3
\citep{chen2024m3embedding} -- are widely adopted in hybrid pipelines
combined with cross-encoder reranking
\citep{nogueira2019passagereranking, yoon2024listt5} and rank-level fusion
\citep{cormack2009rrf}. For documentation-grounded assistants the overall
quality is therefore a property of the whole pipeline, not of the
generator alone.

\paragraph{Parameter-efficient fine-tuning.}
Full fine-tuning of instruction-tuned LLMs is expensive in both compute
and memory; PEFT surveys \citep{han2024peft, wang2025peft} document a
large literature of methods that change model behaviour while keeping the
backbone frozen. Among these, LoRA \citep{hu2022lora} introduces low-rank
adapters on linear projections and has spawned a family of follow-ups,
including QLoRA \citep{dettmers2023qlora} and LoRA+
\citep{hayou2024loraplus}, plus throughput-focused training infrastructure
such as ASPEN \citep{ye2023aspen}. Collectively, this line shows that LoRA
can deliver domain adaptation without full fine-tuning. It does not, however,
answer the question of which LoRA \emph{configuration} -- rank, target
modules, base-model size -- to pick when the adapter is plugged into a
retrieval-augmented system.

\paragraph{RAG evaluation, groundedness, and quality--cost trade-offs.}
The evaluation literature \citep{yu2024evaluation, es2024ragas} stresses
that RAG quality must be measured not only on the final answer but on
retrieval quality, groundedness of the answer in retrieved evidence, and
generation properties. Closer to the trade-off framing of this paper,
\citet{baek2026lorabayesian} treat LoRA-configuration selection itself as
a non-trivial optimisation problem inside the LoRA hyperparameter space,
and \citet{baqar2025raglora} compare RAG, LoRA, and DoRA jointly from an
accuracy-and-faithfulness viewpoint.

\paragraph{Combining RAG and LoRA.}
A small but growing body of work studies RAG and LoRA together. JORA
\citep{tahir2024jora} contributes infrastructure for retrieval-augmented
LoRA fine-tuning; RAMoLE \citep{zhao2024ramole} uses retrieval to select
between multiple LoRA experts at inference time. Among empirical studies,
\citet{baqar2025raglora} is closest in spirit to this work -- it
contrasts RAG, LoRA, and DoRA on question answering -- but, like other
work in this group, treats LoRA as a single technique to be plugged in
rather than as a configuration space to be searched over.

\paragraph{Research gap.}
Across the four sub-areas above, work on RAG focuses on retrieval and
pipeline design; work on LoRA/PEFT focuses on efficient adaptation in the
abstract; work on RAG evaluation establishes that groundedness and
multi-criteria analysis are necessary. What remains underexplored is the
\emph{systematic empirical analysis of how different LoRA configurations
inside a fixed RAG architecture for technical documentation interact with
answer quality, latency, inference memory, and training cost}. This is
the gap addressed by the present paper.

\section{Task and Benchmark}
\label{sec:task}

\subsection{Task formulation}
\label{sec:task-formulation}

We consider documentation-grounded question answering over a fixed corpus
$D=\{d_1,\dots,d_n\}$ of text chunks. A fixed retrieval-and-reranking module
$R$ maps every question $q$ to a context $c=R(q,D)$. The research variable is
the generator configuration $x\in\mathcal{X}$, characterised by the base
model, the LoRA rank, and the set of target modules. For each configuration
$x$ the system produces an answer $a_x(q,c)$. We associate with $x$ a quality
score $Q(x)=F_1(x)$ and a cost vector
$C(x)=\bigl(L_{\mathrm{inf}}(x), M_{\mathrm{inf}}(x), T_{\mathrm{train}}(x),
M_{\mathrm{train}}(x)\bigr)$, where $L_{\mathrm{inf}}$ is mean inference
latency, $M_{\mathrm{inf}}$ is peak inference VRAM, $T_{\mathrm{train}}$ is
total training time, and $M_{\mathrm{train}}$ is peak training VRAM. The goal
is not a single $x^\ast=\arg\max Q(x)$ but the Pareto-optimal set
\citep{lewis2020rag,yu2024evaluation}.

\subsection{Corpus and QA construction}
\label{sec:task-corpus}

The corpus is the official Kubernetes documentation, cleaned and segmented
into semantically coherent chunks suitable for retrieval and answer
attribution. The QA set was built in two stages. We hand-wrote 500 QA pairs
directly from the documentation, then drafted further candidates one-by-one
with an LLM agent based on GPT-5.4 -- not via bulk synthetic augmentation.
All candidates went through manual verification by the author: pairs were
discarded on ambiguous questions, missing supporting evidence, duplicates,
or factual errors in the reference answer. Of $5{,}467$ candidates, $323$
($\approx 5.9\%$) were rejected and $\approx 20\%$ of the remaining pairs
were edited; the rest were accepted as-is. The final benchmark contains
$5{,}144$ verified pairs ($500$ hand-written + $4{,}644$ LLM-drafted and
manually verified), partitioned into train/eval/test as in
Table~\ref{tab:splits}.

\begin{table}[H]
\centering
\small
\begin{tabular}{lrrr}
\toprule
Split & Rows & Exact & Normal \\
\midrule
Train & 3{,}614 & 1{,}449 & 2{,}165 \\
Eval  &   745 &   361 &   384 \\
Test  &   785 &   475 &   310 \\
\bottomrule
\end{tabular}
\caption{Split sizes and distribution of answer types (\textsc{exact} vs.\
\textsc{normal}).}
\label{tab:splits}
\end{table}

\paragraph{Same-family risk.}
The QA-generation model (GPT-5.4) and the LLM judge used in
\S\ref{sec:eval} (\texttt{gpt-5.4-mini}) belong to the same model family,
which raises a same-family-bias concern. There is no direct train/test leak:
the judge is not trained on the gold answers, and all gold answers were
manually verified. We discuss this residual risk explicitly as a limitation
(\S\ref{sec:discussion}).

\subsection{Answer types}
\label{sec:task-answer-types}

We annotate each QA pair as either \textsc{exact} or \textsc{normal}.
\textsc{exact} pairs require near-literal answers -- flags, paths, field
names, API versions, feature gates -- while \textsc{normal} pairs are
factual answers that tolerate paraphrasing as long as the meaning is
preserved. This typology was designed for this benchmark; it is used as
a supervision signal at training time and as a stratification axis at
evaluation time, but it is \emph{not} passed to the generator at inference,
which receives only the question and the retrieved context. This separation
of training- and test-time conditioning follows the evaluation protocols in
prior RAG work \citep{yu2024evaluation,es2024ragas}.

\section{RAG Pipeline}
\label{sec:pipeline}

The pipeline follows the standard three-step structure -- retrieval,
reranking, generation -- consistent with modern RAG surveys
\citep{lewis2020rag, gao2023survey, wu2024survey}.

\paragraph{Retrieval.}
The main configuration uses a hybrid retrieval contour. Dense retrieval
runs over a FAISS \citep{douze2024faiss, johnson2019billionscale} index
of BGE-M3 \citep{chen2024m3embedding} embeddings; sparse retrieval uses
the native sparse channel of BGE-M3. Dense and sparse candidate lists
are merged via Reciprocal Rank Fusion \citep{cormack2009rrf}. Classical
BM25 \citep{robertson2009bm25} is used only as the sparse component in a
separate ablation regime (\S\ref{sec:results-ablation}), not in the main
pipeline.

\paragraph{Reranking.}
The fused candidate list passes through a pretrained cross-encoder
reranker, \texttt{BAAI/bge-reranker-v2-m3}, building on the standard
cross-encoder and listwise reranking literature
\citep{nogueira2019passagereranking, yoon2024listt5} and on retrieval
extensions of large language models \citep{li2023llmembedder}.

\paragraph{Generator and prompting.}
The generator is one of two instruction-tuned models from the Llama~3
family \citep{dubey2024llama3}: \texttt{Llama-3.2-3B-Instruct} and
\texttt{Llama-3.1-8B-Instruct}. Both are decoder-only transformers with
RoPE positional encoding, grouped-query attention, and SwiGLU MLP
blocks. At inference we use a \emph{neutral} prompting mode, with no
explicit instruction to ``answer only from the context'' -- the role
of strict grounding is left to LoRA fine-tuning and to retrieval quality
rather than to prompt engineering. This separates the contribution of
adaptation from that of prompt-side anchoring and keeps the comparison
between adapters honest. At training time the model retains the option
to condition on the answer-type label (\textsc{exact}/\textsc{normal},
\S\ref{sec:task-answer-types}), but the inference prompt does not carry
that label. This split between training and test-time conditioning
follows established practice in RAG evaluation
\citep{yu2024evaluation, es2024ragas}.

\paragraph{Inference context.}
The main inference regime uses \texttt{eval\_top\_k}$=2$ retrieved
chunks, matching \texttt{embed\_top\_k}$=2$ at training time. For
sensitivity analysis we additionally run the full grid with
\texttt{eval\_top\_k}$\in\{1,4\}$ (\S\ref{sec:results-stats}).

\section{LoRA Configurations and Experimental Design}
\label{sec:lora}

\subsection{Configuration space}
\label{sec:lora-space}

Full fine-tuning of the chosen Llama-3 models is too expensive for the
configuration search we want to run. We therefore use Low-Rank
Adaptation \citep{hu2022lora}, which is well-suited for ablations
because it changes the generator's behaviour while keeping the backbone
frozen \citep{dettmers2023qlora, han2024peft}.

We train 20 LoRA adapters on top of two base models. For each base model
we vary two factors: the LoRA rank $r\in\{4, 8, 16, 32, 64\}$ and the
set of adapted modules. The latter takes two values:
\texttt{qv\_only} adapts the $q$ and $v$ attention projections only;
\texttt{full\_attention} adapts all four major attention projections
($q$, $k$, $v$, $o$). The two unadapted base models, \texttt{3B baseline}
and \texttt{8B baseline}, serve as controls evaluated under the same
inference protocol.

\subsection{Fixed training hyperparameters}
\label{sec:lora-hparams}

All training runs use a common optimisation setup:
\texttt{num\_train\_epochs}$=8$, \texttt{learning\_rate}$=2\times10^{-5}$,
AdamW \citep{kingma2015adam, loshchilov2019adamw} with cosine schedule
\citep{loshchilov2017sgdr} and a linear warm-up over $\approx 3\%$ of the
total steps, in \texttt{bf16} mixed precision. We use
\texttt{embed\_top\_k}$=2$ retrieved chunks per training example.
LoRA-side hyperparameters are also fixed across runs:
\texttt{bias}$=$\texttt{none}, \texttt{task\_type}$=$\texttt{CAUSAL\_LM},
and \texttt{lora\_dropout}$=0.05$. We tie \texttt{lora\_alpha} to rank
via the rule $\alpha=2r$. With these constraints, the comparison
isolates three sources of variation: the base-model size, the rank, and
the target-module choice.

The training prompt uses a mixed-context strategy. When supporting
chunks are pre-annotated for an example they are used directly;
otherwise the context is materialised from the retrieval contour
(\S\ref{sec:pipeline}). This is methodologically close to
retrieval-augmented fine-tuning setups such as JORA
\citep{tahir2024jora}, and narrows the gap between idealised training
and the actual inference regime.

\subsection{Hardware and inference stack}
\label{sec:lora-hardware}

All runtime cost metrics (\textit{latency}, \textit{inference VRAM},
\textit{training time}, \textit{training VRAM}) are measured on a
\emph{single} hardware configuration, to remove device-related variance
from the comparison. The compute node runs in Yandex DataSphere: one
NVIDIA A100 (40\,GB), 28 vCPU, and 114\,GB of host RAM. The same node is
used both for LoRA training and for inference, so training and inference
metrics are directly comparable across configurations.

The inference stack is Hugging Face Transformers with PEFT: the base
model is loaded in its native precision without quantisation
(\texttt{--no-quant-generator} and \texttt{--no-quant-judge} are set),
the LoRA adapter is attached separately via
\texttt{PeftModel.from\_pretrained}, and the mixed-precision setting
matches training (\texttt{bf16}). The retrieval contour uses
\texttt{BAAI/bge-m3} for dense embeddings and
\texttt{BAAI/bge-reranker-v2-m3} for reranking
(\texttt{reranker\_batch\_size}$=16$, \texttt{retrieve\_top\_n}$=20$).
Generation is greedy (no sampling) with a fixed maximum answer length
shared across configurations.

Latency is measured per-sample (effective generator batch size of 1) as
the mean wall-clock time per test example over the full test split
($n=785$), counting all pipeline stages: query embedding, dense and
sparse retrieval, reranking, prompt construction, and generation.
Inference VRAM is logged as peak device memory via
\texttt{torch.cuda.max\_memory\_allocated()} at the end of each run.

\subsection{Hypotheses}
\label{sec:lora-hypotheses}

The experimental design tests four working hypotheses:
(i) scaling the base model improves quality but raises both latency and
inference memory;
(ii) increasing rank expands adaptation capacity, but with diminishing
returns at high $r$;
(iii) broader target-module coverage (\texttt{full\_attention}) is not
necessarily preferable once training and inference costs are factored in;
and
(iv) among the resulting configurations, the practically useful working
points are not the per-metric maxima but the Pareto-optimal set.

\section{Evaluation Methodology}
\label{sec:eval}

\subsection{Quality metrics}
\label{sec:eval-quality}

The primary quality metric is \emph{token-level $F_1$} between the
generated and the gold answer \citep{yu2024evaluation}. All headline
results and per-regime tables and plots in
\S\ref{sec:results} and Appendix~\ref{app:tables} are computed on the
held-out test split ($n=785$); the eval split ($n=745$) is used for
configuration selection during the experimental loop. Along with the
point estimate of $F_1$ we report a non-parametric bootstrap 95\%
confidence interval ($1{,}000$ resamples on the test split). For paired
comparisons of $\Delta F_1$ between configurations we use the paired
bootstrap on the same test split, which is appropriate for ranking close
configurations on a shared sample.

We do not report embedding-based semantic metrics such as BERTScore
\citep{zhang2020bertscore}. For documentation-grounded question
answering with a high share of \textsc{exact} questions, token $F_1$ is
the most directly interpretable similarity metric; the semantic side of
quality is instead captured by the judge-based scores described next.

\subsection{Judge-based groundedness and correctness}
\label{sec:eval-judge}

We add two judge-based quality axes computed by an external LLM judge,
\texttt{gpt-5.4-mini}. The judge receives only the triple
\textit{(question, context, answer)} and is blind to the generator
identity -- no \texttt{model\_id} or configuration name is provided --
so that scores are not contaminated by adapter or model branding. It
assigns two independent ratings: \emph{correctness}, the content-level
accuracy of the answer with respect to the provided context, and
\emph{groundedness}, the degree to which the answer is supported by the
context with no unsupported additions
\citep{yu2024evaluation, es2024ragas, baqar2025raglora}.

We aggregate these into two pass-at-$k$ scores,
\textit{correctness\_pass@4} and \textit{groundedness\_pass@4}: the
fraction of answers that the judge rates $\geq 4$ on the corresponding
scale. These judge-based scores are used as a \emph{second} axis of
quality alongside $F_1$, not as a replacement. As
\S\ref{sec:results-groundedness} shows, the configuration that maximises
$F_1$ is often not the configuration that maximises grounding, which is
itself an informative finding.

\subsection{Cost metrics}
\label{sec:eval-cost}

We measure four cost quantities. The two inference-side metrics are mean
latency $L_{\mathrm{inf}}$ and peak inference VRAM $M_{\mathrm{inf}}$
(\S\ref{sec:lora-hardware}). The two training-side metrics are total
training time $T_{\mathrm{train}}$ and peak training VRAM
$M_{\mathrm{train}}$. The four together form the cost vector used in the
Pareto analysis (\S\ref{sec:task-formulation}). This setup follows the
quality--cost framing used in RAG and LoRA evaluation literature
\citep{es2024ragas, baek2026lorabayesian, baqar2025raglora}.

\subsection{Pareto analysis}
\label{sec:eval-pareto}

Comparison across the resulting configuration set is done with a
multi-objective Pareto analysis. A configuration $x$ is Pareto-optimal
if there is no other configuration $x'$ that is at least as good on
quality and at least as good on every cost dimension, and strictly
better on at least one of them (\S\ref{sec:task-formulation}). In the
two-dimensional fronts of \S\ref{sec:results} the quality coordinate is
$F_1$ and the cost coordinate alternates between mean inference latency,
peak inference VRAM, training time, and training VRAM. This protocol
focuses the discussion on the non-dominated set rather than on a single
``best'' configuration; the latter view is incompatible with the actual
practical question of choosing a working point under a cost budget.

\section{Results}
\label{sec:results}

All results below are computed on the held-out test split ($n=785$) with the
fixed hybrid-retrieval pipeline described in \S\ref{sec:pipeline} and the
fixed training and inference setup described in \S\ref{sec:lora}. Unless
noted otherwise, F1 denotes token-level F1 and \emph{grnd@4} (resp.
\emph{corr@4}) denotes the LLM-judge groundedness (resp.\ correctness)
pass@4 score (\S\ref{sec:eval}). 95\% bootstrap confidence intervals (CIs)
are computed with $1{,}000$ resamples; for paired comparisons of $\Delta F_1$
between configurations we use the paired bootstrap on the same test split.

\subsection{Quality vs.\ inference cost}
\label{sec:results-quality-cost}

\begin{figure}[!htbp]
\centering
\includegraphics[width=0.95\linewidth]{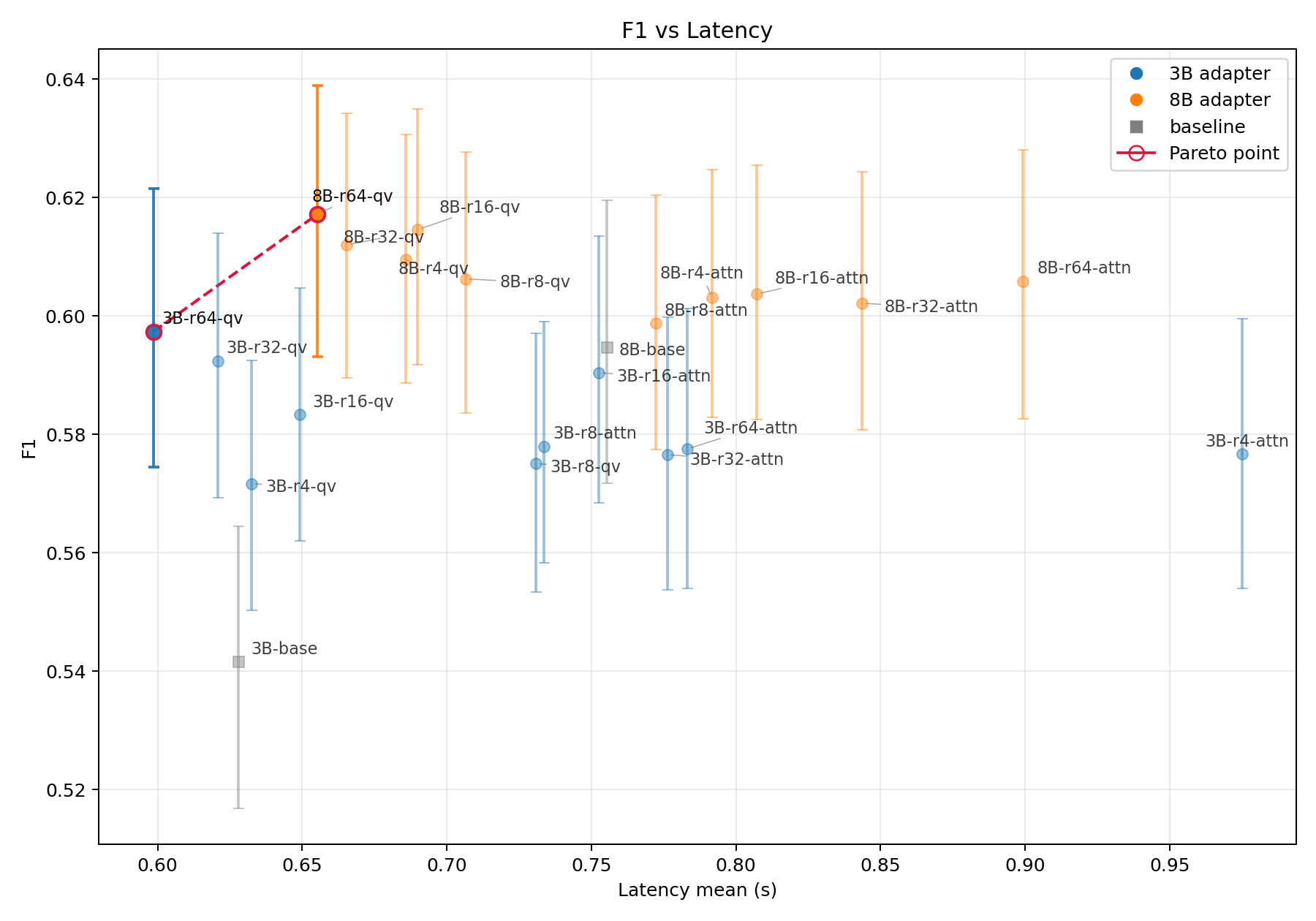}
\caption{Token-level F1 vs.\ end-to-end inference latency (per query) in the
base regime. Error bars are 95\% bootstrap CIs on $F_1$. The Pareto front
(red dashed) connects the two non-dominated configurations,
\texttt{3B r64 qv\_only} and \texttt{8B r64 qv\_only}; all
\texttt{full\_attention} adapters lie strictly inside the front.}
\label{fig:f1-vs-latency}
\end{figure}

On the F1-vs-latency front the non-dominated configurations are
\texttt{3B r64 qv\_only} ($F_1=0.597$\,[0.574, 0.622]) and
\texttt{8B r64 qv\_only} ($F_1=0.617$\,[0.593, 0.639]); see
Table~\ref{tab:runtime-front}. The 8B point gains $\Delta F_1=+0.020$ over
the 3B point with paired-bootstrap 95\%~CI $[+0.0005, +0.0410]$ -- a small
but statistically supported margin -- at the cost of an additional
$\sim$9\,GB of inference memory and a $0.057$\,s latency increase. Notably,
\texttt{3B r64 qv\_only} is statistically indistinguishable from the
unadapted \texttt{8B baseline} ($0.595$\,[0.572, 0.620]):
$\Delta F_1\in[-0.021, +0.026]$. The runtime-front choice is therefore not
between two qualities but between two operating regimes.

\begin{table}[H]
\centering
\small
\begin{tabular}{lcrr}
\toprule
Config. & $F_1$ [95\% CI] & Lat.\ (s) & VRAM (GB) \\
\midrule
\texttt{3B r64 qv\_only} & 0.597\,[0.574, 0.622] & 0.598 & 12.76 \\
\texttt{8B r64 qv\_only} & 0.617\,[0.593, 0.639] & 0.655 & 21.93 \\
\bottomrule
\end{tabular}
\caption{Non-dominated points on the F1-vs-latency front in the base regime.}
\label{tab:runtime-front}
\end{table}

The F1-vs-inference-VRAM view cleanly separates the 3B and 8B families
(VRAM difference $\sim$9\,GB), whereas within each family the VRAM spread is
only $0.3$--$0.4$\,GB -- well below the 95\% CI width of $F_1$. Inference
memory therefore acts as a discrete family selector rather than a continuous
trade-off knob.

\subsection{Training cost}
\label{sec:results-training-cost}

\begin{figure}[!htbp]
\centering
\includegraphics[width=0.95\linewidth]{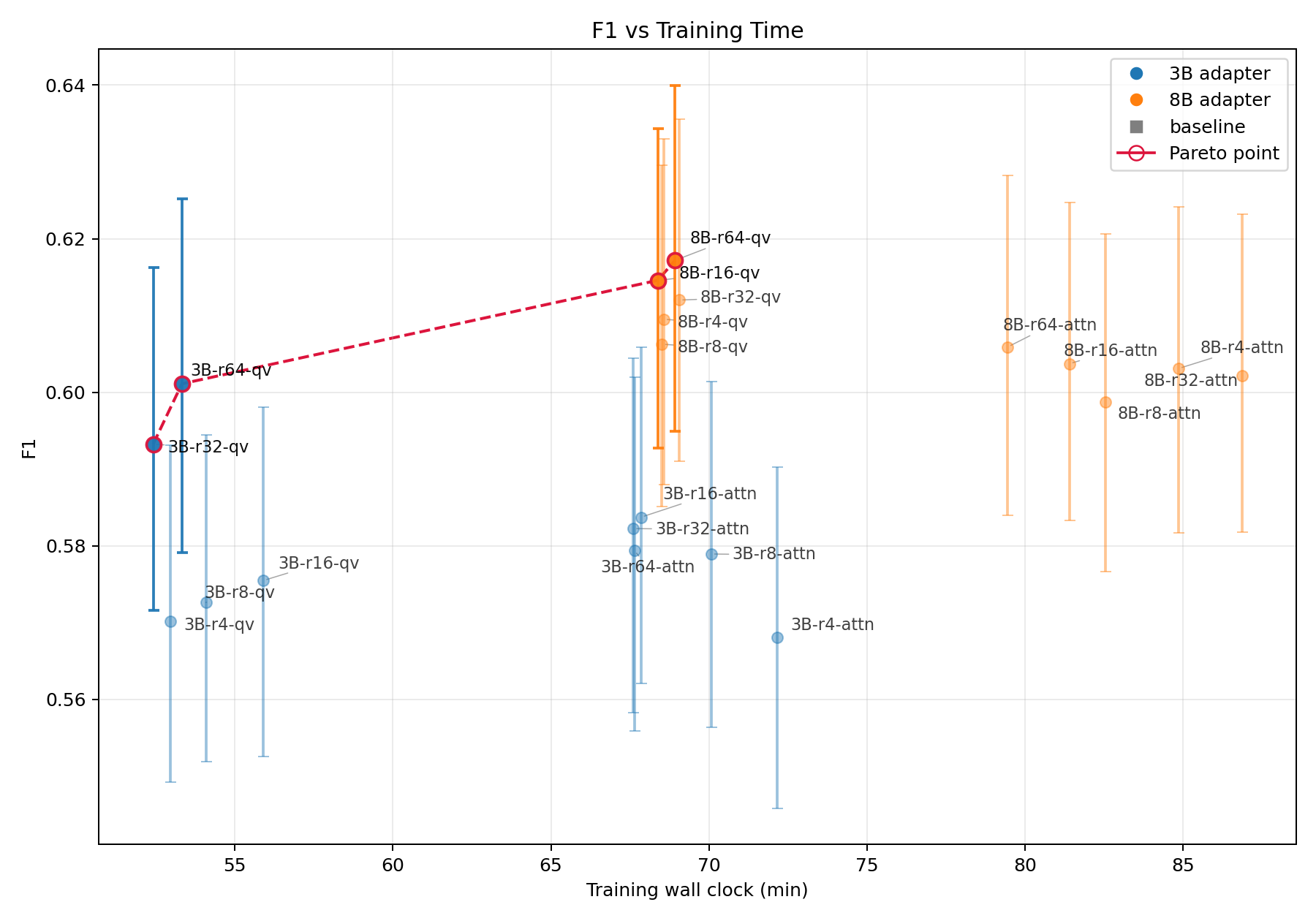}
\caption{$F_1$ vs.\ training wall-clock time (minutes) over all 20 LoRA
configurations. The Pareto front (red dashed) is dominated by
\texttt{qv\_only} adapters; no \texttt{full\_attention} configuration appears
on it. Error bars are 95\% bootstrap CIs on $F_1$.}
\label{fig:f1-vs-training-time}
\end{figure}

\begin{figure}[!htbp]
\centering
\includegraphics[width=\linewidth]{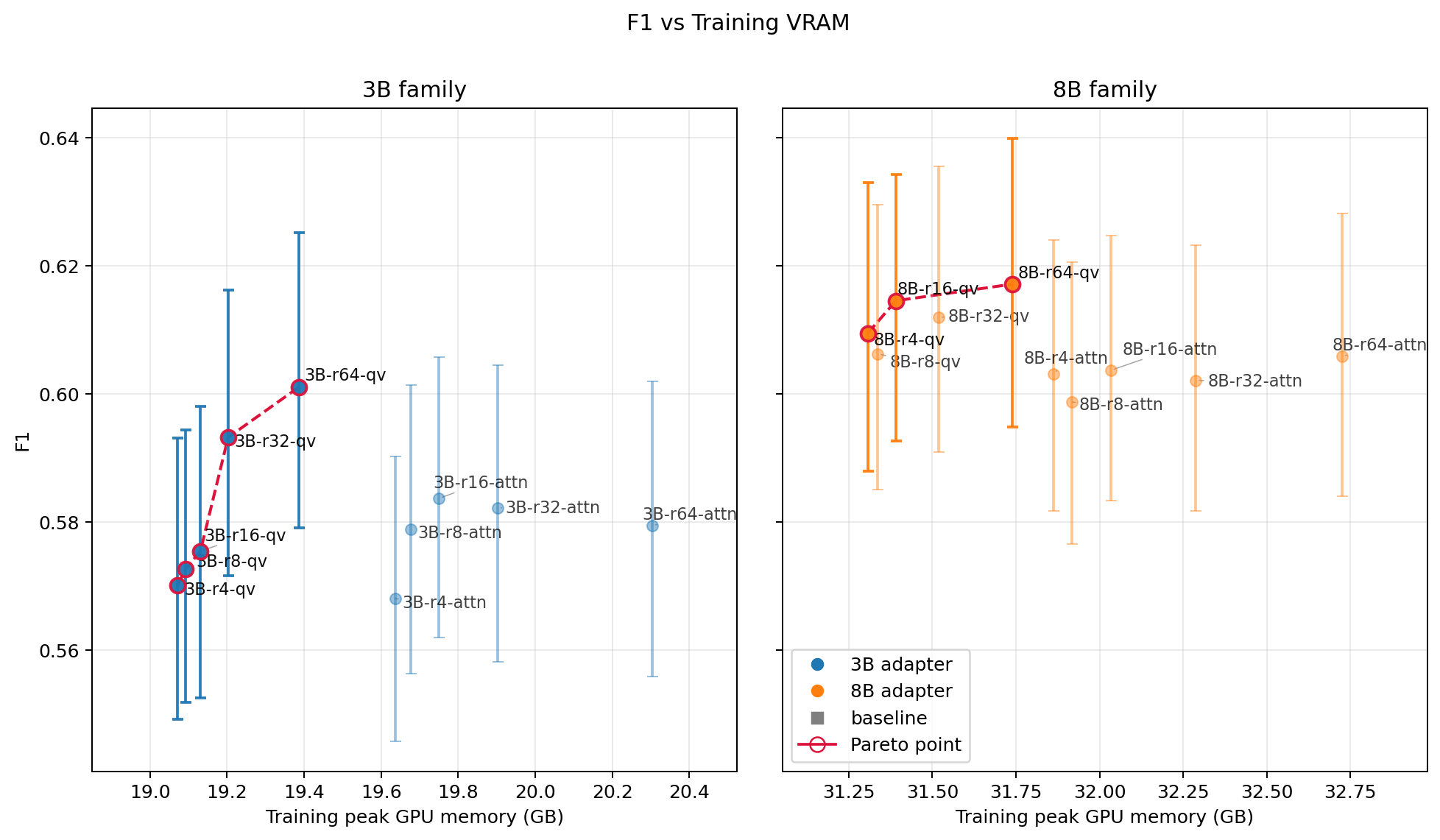}
\caption{$F_1$ vs.\ peak training GPU memory, split by base-model family
(3B / 8B). Within each family, the training Pareto front is again populated
entirely by \texttt{qv\_only} adapters. Error bars are 95\% bootstrap CIs on
$F_1$.}
\label{fig:f1-vs-training-vram}
\end{figure}

The cheapest training point is \texttt{3B r4 qv\_only} ($52.95$\,min,
$19.07$\,GB). Stronger training-side trade-offs cluster around
\texttt{qv\_only} configurations at $r{=}32$ and $r{=}64$ for the 3B
family and $r{=}16$ and $r{=}64$ for the 8B family. \emph{No}
\texttt{full\_attention} configuration appears on the training Pareto front
(Table~\ref{tab:training-front}). Within the 8B family the upper
\texttt{qv\_only} points ($r{=}16$ and $r{=}64$) have largely overlapping
$F_1$ CIs, so the choice between them is driven by secondary criteria, not
by a robust quality gap.

\begin{table}[H]
\centering
\small
\begin{tabular}{lccc}
\toprule
Config. & $F_1$ [95\% CI] & Train (min) & VRAM (GB) \\
\midrule
\texttt{3B r4 qv\_only}  & 0.572\,[0.550, 0.592] & 52.95 & 19.07 \\
\texttt{3B r8 qv\_only}  & 0.573\,[0.552, 0.594] & 54.08 & 19.09 \\
\texttt{3B r16 qv\_only} & 0.583\,[0.562, 0.605] & 55.90 & 19.13 \\
\texttt{3B r32 qv\_only} & 0.592\,[0.569, 0.614] & 52.41 & 19.20 \\
\texttt{3B r64 qv\_only} & 0.597\,[0.574, 0.622] & 53.32 & 19.39 \\
\texttt{8B r4 qv\_only}  & 0.610\,[0.588, 0.633] & 68.57 & 31.31 \\
\texttt{8B r16 qv\_only} & 0.615\,[0.593, 0.634] & 68.38 & 31.39 \\
\texttt{8B r64 qv\_only} & 0.617\,[0.595, 0.640] & 68.93 & 31.74 \\
\bottomrule
\end{tabular}
\caption{Non-dominated points on the training Pareto front (time and VRAM).
All training-front points use the \texttt{qv\_only} adaptation scheme.}
\label{tab:training-front}
\end{table}

\subsection{Groundedness as a second quality axis}
\label{sec:results-groundedness}

\begin{figure}[!htbp]
\centering
\includegraphics[width=0.95\linewidth]{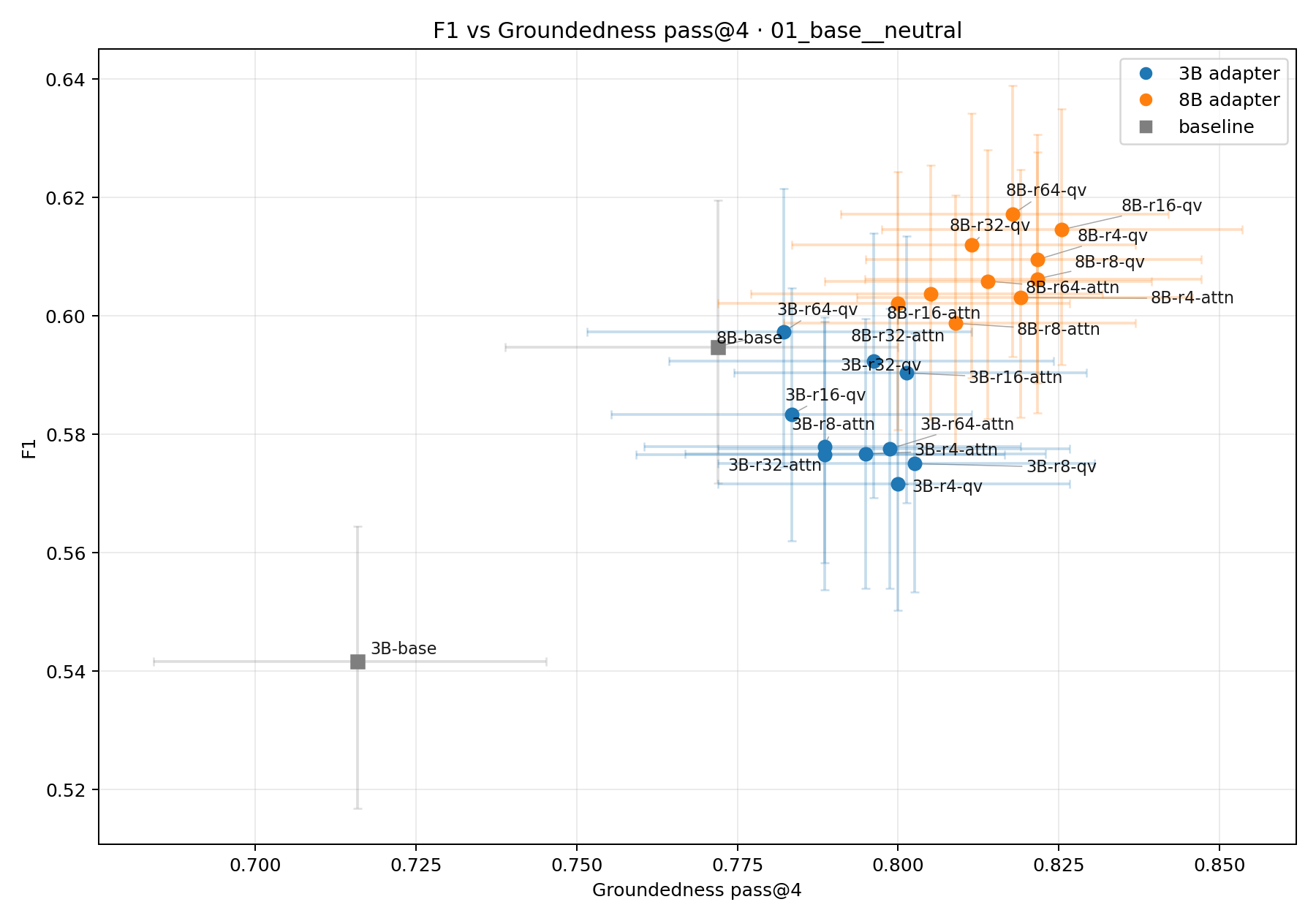}
\caption{$F_1$ vs.\ LLM-judge groundedness pass@4 in the base regime. Error
bars are 95\% bootstrap CIs on both axes. Adapted configurations cluster
tightly in the upper right, while the unadapted baselines are clearly
separated on \emph{grnd@4}.}
\label{fig:f1-vs-grnd}
\end{figure}

The configuration with the highest point F1, \texttt{8B r64 qv\_only}
($F_1=0.617$, \emph{grnd@4}$=0.818$), does \emph{not} maximise groundedness.
The grnd@4 maximum ($0.825$) is reached by \texttt{8B r16 qv\_only}
($F_1=0.615$); the two are statistically indistinguishable on $F_1$
($\Delta F_1\in[-0.011, +0.016]$). LoRA-adapted configurations also shift
grnd@4 upward by $0.05$--$0.08$ over the unadapted baselines at comparable
latency -- a margin that exceeds the typical CI width ($\approx 0.03$).
Within already-adapted configurations, grnd@4 spread mostly falls inside
the CIs and separation is driven by the 3B/8B choice. Task quality and
grounding are therefore not interchangeable: for documentation QA, the
choice of optimal configuration depends on whether the deployment prioritises
exact-match agreement, supportedness, or a compromise between them.

\subsection{Robustness across retrieval and prompting}
\label{sec:results-ablation}

\begin{figure}[!htbp]
\centering
\includegraphics[width=0.9\linewidth]{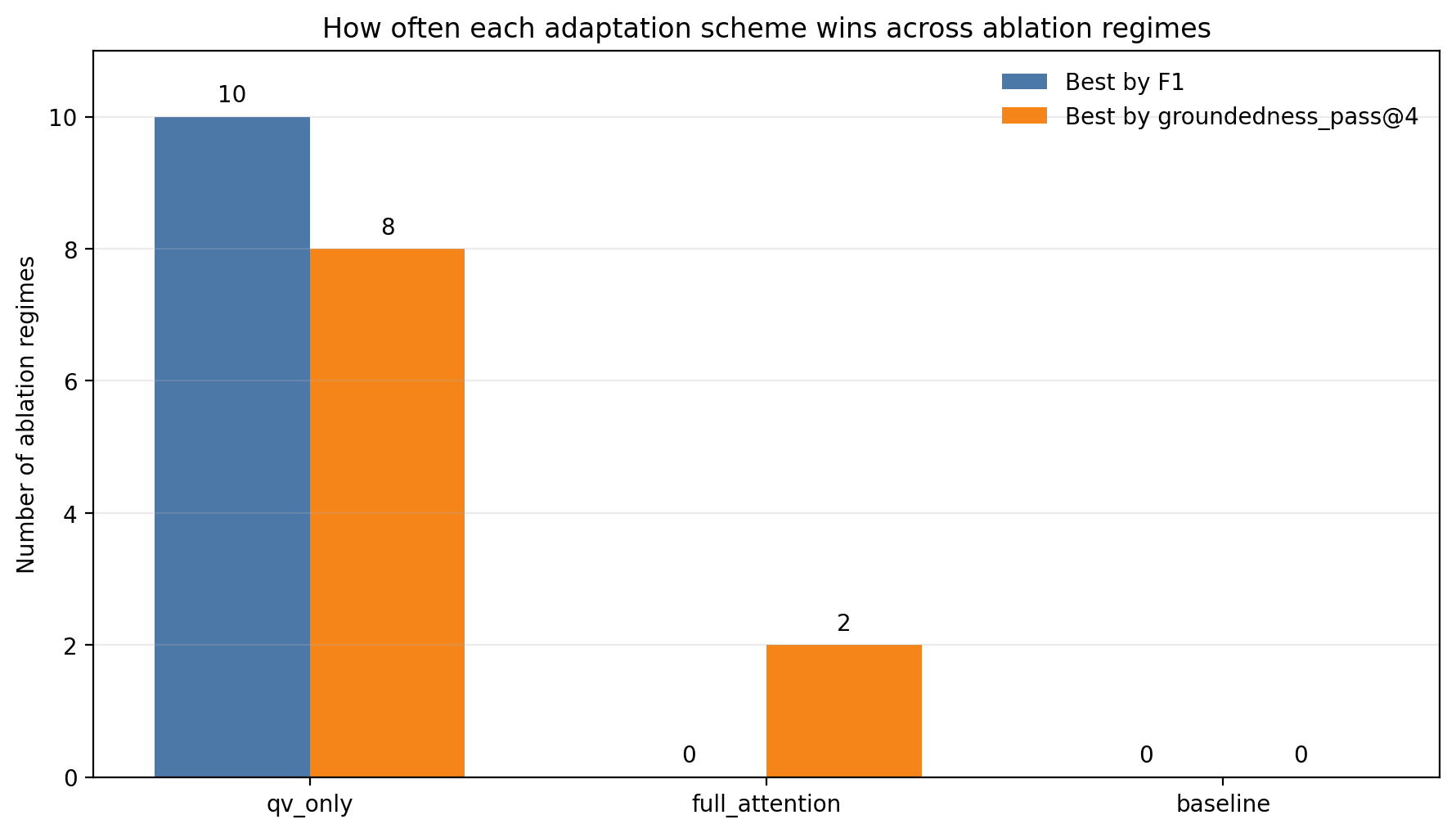}
\caption{Frequency with which each adaptation scheme yields the best
configuration across the 10 retrieval/prompting ablation regimes, separately
for $F_1$ (blue) and \emph{grnd@4} (orange).}
\label{fig:ablation-wins}
\end{figure}

To test whether the LoRA conclusions are an artefact of one pipeline
choice, we re-run the full grid under 10 retrieval/prompting regimes:
five retrieval variants (\texttt{base}, \texttt{reranker\_off},
\texttt{dense\_only}, \texttt{sparse\_only}, \texttt{hybrid\_bm25}) crossed
with two prompting modes (\texttt{neutral}, \texttt{explicit\_grounded}).
Two main observations hold across all 10 modes
(Table~\ref{tab:ablation-summary}, full per-regime tables in
Appendix~\ref{app:tables}). First, the best-F1 configuration is
\texttt{8B r64 qv\_only} in \emph{all} 10 modes, and the best-grnd@4
configuration concentrates on \texttt{qv\_only} adapters in 8 of 10 modes
(\texttt{full\_attention} appears only in two \texttt{sparse}-based modes).
Second, the F1-optimal and grnd@4-optimal points never coincide
(\texttt{same\_point}=\texttt{no} in 10/10 modes). The choice of working
configuration is therefore a function of the retrieval--prompting setup
\emph{and} the quality criterion, not of LoRA alone.

\begin{table}[H]
\centering
\small
\begin{tabular}{lcc}
\toprule
Criterion & \texttt{qv\_only} wins & \texttt{full\_attention} wins \\
\midrule
Best $F_1$        & 10/10 & 0/10 \\
Best \emph{grnd@4} &  8/10 & 2/10 \\
\bottomrule
\end{tabular}
\caption{Frequency with which each adaptation scheme yields the best
configuration across the 10 retrieval/prompting ablation regimes.}
\label{tab:ablation-summary}
\end{table}

\subsection{Statistical robustness of $\Delta F_1$}
\label{sec:results-stats}

Table~\ref{tab:paired-bootstrap} reports paired-bootstrap CIs for six key
comparisons. The reliably supported effects are
(i)~\texttt{3B r64 qv\_only} over \texttt{3B baseline},
(ii)~\texttt{3B r64 qv\_only} over the param-matched
\texttt{3B r64 full\_attention} adapter, and
(iii)~the family-level $\Delta F_1$ of \texttt{qv\_only} over
\texttt{full\_attention} averaged over the eight param-matched pairs
analysed in \S\ref{sec:control}. In contrast, the gaps between
\texttt{3B r64 qv\_only} and the \texttt{8B baseline}, and between the top
two 8B \texttt{qv\_only} points, are not distinguishable from sampling
noise at $n=785$. The headline of this work is therefore not any single
local maximum on $F_1$, but the structural advantage of the
\texttt{qv\_only} adaptation scheme.

\begin{table}[H]
\centering
\small
\begin{tabular}{lcc}
\toprule
Comparison & $\Delta F_1$ & 95\% CI \\
\midrule
\texttt{3B r64 qv\_only} - \texttt{3B baseline}          & $+0.056$ & $[+0.033, +0.078]$ \\
\texttt{3B r64 qv\_only} - \texttt{8B baseline}          & $+0.003$ & $[-0.021, +0.026]$ \\
\texttt{8B r64 qv\_only} - \texttt{3B r64 qv\_only}      & $+0.020$ & $[+0.0005, +0.0410]$ \\
\texttt{8B r64 qv\_only} - \texttt{8B r16 qv\_only}      & $+0.003$ & $[-0.011, +0.016]$ \\
\texttt{3B r64 qv\_only} - \texttt{3B r64 full\_attention} & $+0.020$ & $[+0.004, +0.036]$ \\
\texttt{qv\_only} - \texttt{full\_attention}$^{*}$         & $+0.007$ & $[+0.001, +0.012]$ \\
\bottomrule
\end{tabular}

\vspace{0.3em}
{\footnotesize\textsuperscript{*}\,mean $\Delta F_1$ over the 8~param-matched
pairs (Table~\ref{tab:param-matched}).\par}
\caption{Paired-bootstrap 95\% CIs on $\Delta F_1$ ($n=785$, $1{,}000$
resamples).}
\label{tab:paired-bootstrap}
\end{table}

\section{Param-Matched Control Comparison}
\label{sec:control}

Up to this point, the $F_1$ advantage of \texttt{qv\_only} adapters
(\S\ref{sec:results-quality-cost}--\S\ref{sec:results-ablation}) could
have two natural explanations. The first is \emph{structural}: adapting
exactly the two projections that select and re-weight the retrieved
context is sufficient inside this RAG architecture, while extending
adaptation to $k$ and $o$ projections grows the adapter without buying
proportional quality. The second is \emph{parametric}: with the same
rank $r$, \texttt{full\_attention} adapts twice as many projections as
\texttt{qv\_only} and thus splits a fixed-shape low-rank budget more
thinly across attention sub-spaces, so the per-projection capacity may
simply be too small.

To distinguish the two, we run a param-matched control. Because
\texttt{full\_attention} adapts four projections instead of two, equal
total LoRA parameter counts are achieved by halving the rank for
\texttt{full\_attention}. We therefore pair each \texttt{qv\_only}
configuration at rank $r$ with the \texttt{full\_attention} configuration
at rank $r/2$, computing $\Delta F_1=F_1(\textsc{qv\_only}) -
F_1(\textsc{full\_attention})$ on the test split with paired bootstrap
95\% confidence intervals. Results are reported in
Table~\ref{tab:param-matched}.

\begin{table}[H]
\centering
\small
\begin{tabular}{cccccc}
\toprule
Family & Param budget & \texttt{qv\_only} & \texttt{full\_attention} & $\Delta F_1$ & 95\% CI \\
\midrule
3B & 256$d$ & $r{=}64$ & $r{=}32$ & $+0.021$ & $[+0.005, +0.037]$ \\
3B & 128$d$ & $r{=}32$ & $r{=}16$ & $+0.002$ & $[-0.013, +0.017]$ \\
3B &  64$d$ & $r{=}16$ & $r{=}8$  & $+0.005$ & $[-0.008, +0.020]$ \\
3B &  32$d$ & $r{=}8$  & $r{=}4$  & $-0.002$ & $[-0.014, +0.011]$ \\
8B & 256$d$ & $r{=}64$ & $r{=}32$ & $+0.015$ & $[-0.001, +0.030]$ \\
8B & 128$d$ & $r{=}32$ & $r{=}16$ & $+0.008$ & $[-0.006, +0.023]$ \\
8B &  64$d$ & $r{=}16$ & $r{=}8$  & $+0.016$ & $[+0.003, +0.027]$ \\
8B &  32$d$ & $r{=}8$  & $r{=}4$  & $+0.003$ & $[-0.009, +0.017]$ \\
\bottomrule
\end{tabular}
\caption{Param-matched comparison between \texttt{qv\_only} and
\texttt{full\_attention} at equal total LoRA parameter count. $d$ is
the per-projection rank dimensionality budget. Paired-bootstrap 95\% CIs
on $\Delta F_1$.}
\label{tab:param-matched}
\end{table}

Two observations follow. First, at equal parameter budget,
\texttt{qv\_only} is significantly better in 2 of 8 pairs (one in each
family), statistically indistinguishable from \texttt{full\_attention}
in the remaining 6, and significantly worse in none. The advantage
therefore survives the parameter-count control: it is not the case that
\texttt{full\_attention} would dominate once you give it the same budget
as \texttt{qv\_only}. Second, the only family-budget combination at
which \texttt{full\_attention} comes close is 3B at the smallest budget
($32d$, $r{=}8$ vs.\ $r{=}4$), where the CI of $\Delta F_1$ straddles
zero on the negative side. This is consistent with the interpretation
that at very low rank \texttt{qv\_only} is itself capacity-constrained,
and \texttt{full\_attention} can match it by spreading the same budget
over more projections; once rank grows beyond $r{=}8$ for either scheme,
the structural advantage of \texttt{qv\_only} re-emerges.

\section{Discussion}
\label{sec:discussion}

\subsection{Why \texttt{qv\_only} wins on the Pareto front}
\label{sec:discussion-qv}

Across every Pareto front in \S\ref{sec:results} the non-dominated
points concentrate around \texttt{qv\_only} adapters, and the
paired-bootstrap test against the param-matched
\texttt{full\_attention} alternative gives a mean
$\Delta F_1=+0.007$\,$[+0.001, +0.012]$
(\S\ref{sec:results-stats}, \S\ref{sec:control}). The most plausible
explanation is that, inside a fixed retrieval contour, the adapter's
job is not to globally restructure attention but to refine how the
generator \emph{selects from} and \emph{re-weights} the already-supplied
context. The $q$ and $v$ projections are precisely the two attention
projections that control these two operations; adapting also $k$ and
$o$ increases the adapter size and the training cost faster than it
buys robust gain on the front.

\subsection{Backbone size sets the regime, not the verdict}
\label{sec:discussion-backbone}

The 3B/8B split looks structural but should not be read as the
headline. A larger 8B backbone does set the upper $F_1$ ceiling, but it
almost automatically commits the system to a more expensive operating
regime -- roughly $9$\,GB of additional inference VRAM and $\sim$15
extra training minutes per adapter. Crucially, the strongest adapted 3B
configuration is statistically indistinguishable from the unadapted 8B
baseline ($\Delta F_1\in[-0.021, +0.026]$). So scaling the backbone is
not the only way to recover that level of quality: targeted LoRA
adaptation closes the gap at a fraction of the cost. The backbone
choice is therefore best understood as selecting an operating regime
(small vs.\ large inference budget), with the within-regime quality
controlled mainly by the adapter scheme.

\subsection{Rank effect is sub-linear but practically useful}
\label{sec:discussion-rank}

Increasing $r$ helps $F_1$ more than it hurts inference cost in this
pipeline. For 3B \texttt{qv\_only}, going from $r{=}4$ to $r{=}64$ moves
$F_1$ from $0.572$ to $0.597$, while latency stays in $0.60$--$0.63$\,s
and inference VRAM moves only from $12.67$ to $12.76$\,GB -- well
inside the noise of measurement. The dominant inference cost in this
pipeline is the base model and the retrieval contour, not the LoRA
adapter itself. The same monotone-but-saturating pattern holds for the
8B family, but the upper rungs ($r{=}16$ vs.\ $r{=}64$) are within
paired-bootstrap noise, so within the top 8B configurations the rank
should be picked by secondary criteria (e.g., groundedness, or training
cost), not by point F1.

\subsection{Retrieval and prompting shift the optimum but not the family}
\label{sec:discussion-ablation}

The 10-mode ablation shows that switching off the reranker, going
dense- or sparse-only, or replacing the native sparse channel with BM25,
together with the choice of neutral vs.\ explicit-grounded prompting,
\emph{does} move the per-mode optimum point inside the same
configuration grid. What it does \emph{not} do is overturn the qualitative
pattern: \texttt{8B r64 qv\_only} is the best-$F_1$ configuration in
$10/10$ modes, and \texttt{qv\_only} adapters win on
\textit{grnd@4} in 8 of 10. Retrieval and prompting therefore decide how
favourable the operating environment is, not which adaptation scheme
should be used in it.

\subsection{Sensitivity to context budget (\texorpdfstring{$\mathrm{top}_k$}{top\_k})}
\label{sec:discussion-topk}

\begin{figure}[!htbp]
\centering
\includegraphics[width=0.85\linewidth]{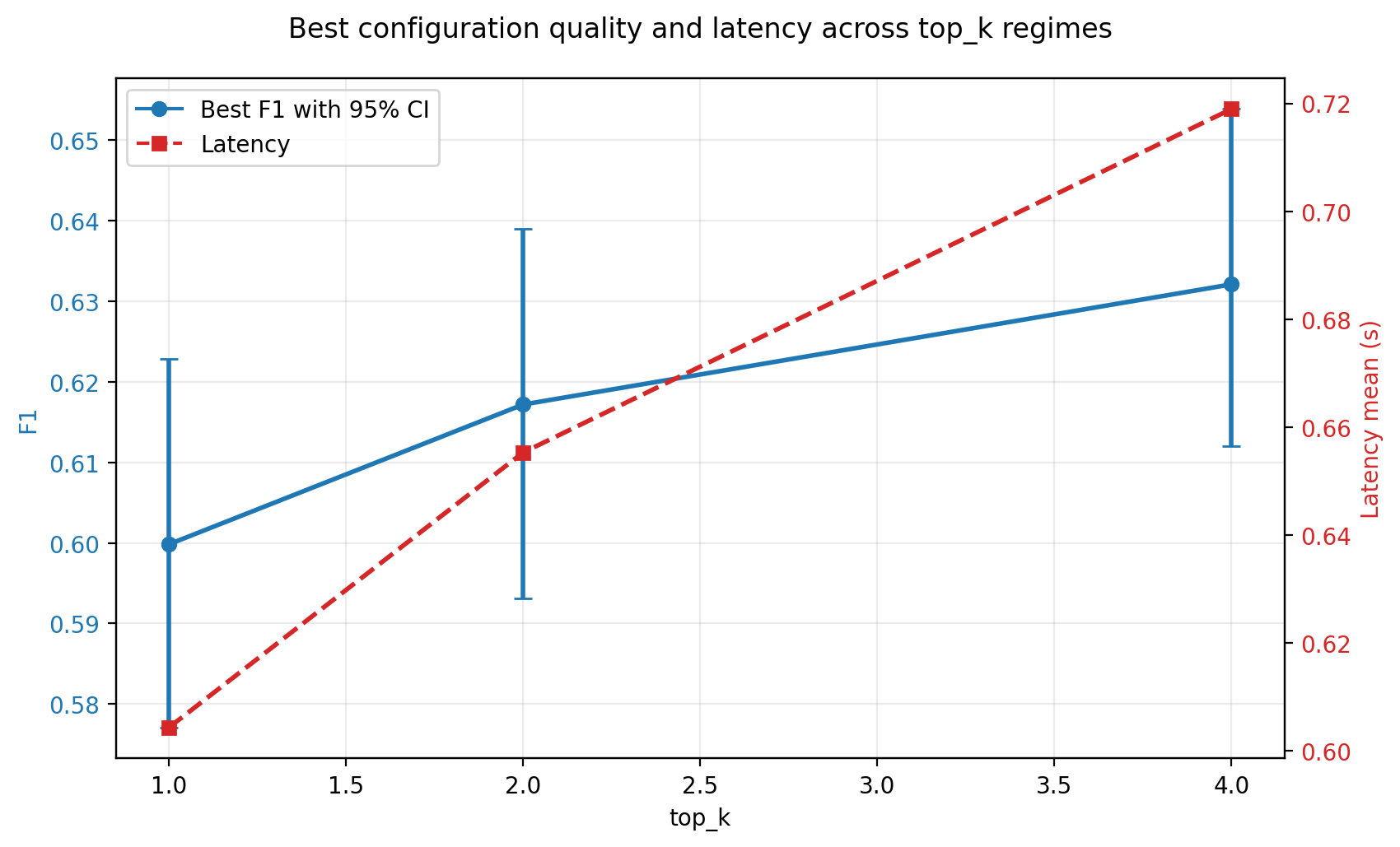}
\caption{Effect of the retrieval cutoff $k$ on the best $F_1$ (left axis,
blue; with 95\% bootstrap CI) and on mean end-to-end inference latency
(right axis, red). The chosen working point $k\!=\!2$ corresponds to the
``knee'' of the curve, beyond which marginal $F_1$ gains are dominated by
marginal latency cost.}
\label{fig:topk-summary}
\end{figure}

A separate generalisation run varies the number of retrieved chunks at
inference time, holding all adapters fixed. Increasing
\texttt{eval\_top\_k} from $1$ to $4$ raises the best $F_1$ from $0.600$
to $0.632$ but adds $\sim$0.12\,s of latency
(Figure~\ref{fig:topk-summary}, Table~\ref{tab:topk-summary}). The best-$F_1$
configuration is \texttt{8B r64 qv\_only} for all three values of
\texttt{top\_k}; the runtime Pareto front collapses to a single point at
\texttt{top\_k}$=1$ and gains \texttt{3B r64 qv\_only} as a second
non-dominated point at \texttt{top\_k}$\in\{2,4\}$. The conclusion is
that broadening retrieval is not a free quality lever, but the
structural ranking of LoRA configurations survives the change.

\begin{table}[H]
\centering
\small
\begin{tabular}{cccrc}
\toprule
\texttt{top\_k} & Best-$F_1$ config & $F_1$ [95\% CI] & Lat.\ (s) & Runtime front \\
\midrule
$1$ & \texttt{8B r64 qv\_only} & 0.600\,[0.577, 0.623] & 0.604 & \{\texttt{8B r64 qv\_only}\} \\
$2$ & \texttt{8B r64 qv\_only} & 0.617\,[0.593, 0.639] & 0.655 & \{\texttt{3B r64 qv\_only}, \texttt{8B r64 qv\_only}\} \\
$4$ & \texttt{8B r64 qv\_only} & 0.632\,[0.612, 0.654] & 0.719 & \{\texttt{3B r64 qv\_only}, \texttt{8B r64 qv\_only}\} \\
\bottomrule
\end{tabular}
\caption{Sensitivity to retrieval context budget at inference time, in
the base regime. All adapters held fixed.}
\label{tab:topk-summary}
\end{table}

\subsection{Error analysis}
\label{sec:discussion-errors}

To make the discussion of residual errors quantitative, we sampled 100
mis-predictions on the test split ($50$ each from
\texttt{3B r64 qv\_only} and \texttt{8B r64 qv\_only},
$F_1<1$, \texttt{seed}$=42$) and labelled them into four operational
classes: \emph{retrieval miss}, \emph{overclaiming}, \emph{incomplete
answer}, and \emph{exact/precision failure}
(Table~\ref{tab:error-types}). \emph{Exact/precision} errors dominate
($53\%$ overall, $66\%$ for the 8B adapter): a substantial share of the
remaining gap is not about missing knowledge but about reproducing a
literal command, flag, version, or field name verbatim. Incomplete
answers ($24\%$) and retrieval misses ($19\%$) follow; pure overclaiming
is rare in this sample ($4\%$). This taxonomy suggests two natural
follow-up directions: post-hoc constrained decoding for the exact-class
sub-distribution, and groundedness-aware re-training for the
incomplete-answer class.

\begin{table}[H]
\centering
\small
\begin{tabular}{lccc}
\toprule
Error type & \texttt{3B r64 qv\_only} & \texttt{8B r64 qv\_only} & Total \\
\midrule
retrieval miss          & 11 (22\%) &  8 (16\%) & 19 (19\%) \\
overclaiming            &  2 (4\%)  &  2 (4\%)  &  4 (4\%)  \\
incomplete answer       & 17 (34\%) &  7 (14\%) & 24 (24\%) \\
exact/precision failure & 20 (40\%) & 33 (66\%) & 53 (53\%) \\
\bottomrule
\end{tabular}
\caption{Manual error-type taxonomy on a balanced 100-sample subset of
mis-predictions ($F_1<1$, fixed random seed).}
\label{tab:error-types}
\end{table}

\section{Conclusion}
\label{sec:conclusion}

We presented a systematic study of LoRA configurations inside a fixed
documentation-grounded RAG pipeline. Three findings stand out. First,
on every Pareto front we examined -- $F_1$ against latency, inference
VRAM, training time, and training VRAM -- the non-dominated points
concentrate around adapters that act on the $q$ and $v$ attention
projections only; full-attention adapters do not appear on the training
Pareto front in this study, and the param-matched control comparison
confirms that the advantage of \texttt{qv\_only} is structural rather
than just a parameter-count artefact. Second, the 3B/8B choice
determines the operating regime (small vs.\ large inference budget),
but a strong LoRA-adapted 3B configuration reaches statistically
comparable $F_1$ to the unadapted 8B baseline at roughly $9$\,GB less
inference VRAM, so backbone size and adaptation interact as
\emph{complements}, not substitutes. Third, judge-based groundedness is
not interchangeable with $F_1$: the configuration that maximises $F_1$
is not the configuration that maximises grounding, and at fixed $F_1$
the difference between them is robust across the 10 retrieval/prompting
ablation regimes.

Practically, the paper produces three recommended working points: a
lightweight one (\texttt{3B r64 qv\_only}) when the deployment is
inference-cost-constrained but expects high quality; the F1-maximising
one (\texttt{8B r64 qv\_only}) when an 8B budget is available; and a
grounding-maximising 8B alternative (\texttt{8B r16 qv\_only}) when
supportedness is the priority.

To support reproducible follow-up work, we will release the manually
verified Kubernetes QA benchmark and the selected LoRA adapters
together with this paper. The most natural extensions of this study
are multi-seed validation for the close-CI comparisons, a
human-agreement study for the LLM judge, a port of the ablation
protocol to a second documentation domain, and a re-measurement of the
runtime axis on alternative inference stacks.

\bibliographystyle{iclr2026_conference}
\bibliography{references}

\newpage
\tableofcontents
\newpage

\appendix
\section{Limitations}
\label{sec:limitations}

\paragraph{Single training seed.}
All LoRA adapters were trained with a single fixed seed
(\texttt{seed}$=42$), which controls adapter initialisation, batch
order, and dropout masks. The bootstrap CIs of \S\ref{sec:results}
capture sampling variance over the test set (fixed model, varying test)
but \emph{not} training-time variance from re-running the same
configuration with a different seed. In the LLM fine-tuning literature,
seed-to-seed dispersion on task metrics can reach a few $F_1$ points
\citep{baek2026lorabayesian}, which is particularly relevant for the
$0.002$--$0.005$ $F_1$ gaps reported here. We therefore phrase claims
about close configurations (e.g.,
\texttt{8B r64 qv\_only} vs.\ \texttt{8B r16 qv\_only}) as
``statistically comparable'' rather than as one strictly dominating the
other.

\paragraph{Single LLM judge, no human-agreement calibration.}
The \textit{grnd@4} and \textit{corr@4} scores are produced by a single
judge (\texttt{gpt-5.4-mini}) under a fixed protocol. The judge is blind
to the generator (no \texttt{model\_id}), which removes the most direct
form of generator-side bias, but we did not run a formal
human-agreement study (e.g., Cohen's $\kappa$ on an expert-annotated
sample) and we did not cross-validate the judge against an alternative
judge model. The judge-based metrics should therefore be read as
indicative scores from one automatic system rather than as final values
of correctness or groundedness. Importantly, all main conclusions of
the paper rely on the judge-free metric $F_1$; the judge-based scores
are used only as a secondary axis to differentiate configurations that
are close on $F_1$. A residual concern is that the judge model
(\texttt{gpt-5.4-mini}) and the model used in QA-candidate drafting
(GPT-5.4, \S\ref{sec:task-corpus}) come from the same family, which may
introduce same-family bias against Llama-style answers. By construction
this bias is uniform across all compared LoRA configurations, so the
relative ordering on the Pareto front is preserved, but the absolute
judge scores should be interpreted with this caveat.

\paragraph{Single domain.}
The empirical results, including absolute metric values and the shape
of the Pareto fronts, are tied to the Kubernetes-documentation corpus
and the curated QA set built on it. The domain has characteristic
properties -- many \textsc{exact} questions, technical terms,
commands, flags, API versions -- that affect both retrieval behaviour
and the distribution of error types (\S\ref{sec:discussion-errors}).
Absolute numbers should not be transported to other documentation
domains (medical regulations, legal corpora, internal company docs)
without re-running the protocol. The \emph{qualitative} findings --
\texttt{qv\_only} advantage, sub-linear rank effect, and the
family-as-regime split -- are conjecturally transferable, but
confirming this requires an equivalent ablation in a different domain.

\paragraph{Single retriever and reranker.}
The retrieval stack (BGE-M3 dense + BGE-M3 native sparse, merged via RRF,
followed by the \texttt{bge-reranker-v2-m3} cross-encoder) is held fixed
across all experiments.
Replacing the embedder or reranker, or moving to a late-interaction
architecture such as ColBERTv2 \citep{santhanam2022colbertv2}, can shift
both absolute metrics and the position of the per-regime optimum.
\S\ref{sec:results} is therefore best read as a statement about LoRA
under this specific retrieval stack, not as a universal claim about RAG
systems.

\paragraph{Single inference hardware and stack.}
Latency, inference VRAM, training time, and training VRAM are all
measured on a single hardware setup (\S\ref{sec:lora-hardware}).
Different batching, alternative inference engines (vLLM, TensorRT-LLM),
different precisions (\texttt{fp16}, \texttt{int8}, \texttt{int4}), or
different parallelism schemes can shift absolute values and possibly
the relative ordering on the runtime axis. The shape of the
quality-versus-cost curves is robust, but absolute numbers should be
re-measured before transferring to a production deployment.

\section{Domain and Problem Statement}
\label{app:domain}

\subsection{Domain}

The domain of the work is question answering over technical documentation. Unlike open-domain conversational systems, here the user expects not a generally plausible answer, but accurate and verifiable information grounded in the documentation. Such a setting fits the logic of Retrieval-Augmented Generation (RAG), in which the answer must rely on an external corpus rather than only on the parametric memory of the model \cite{lewis2020rag, gao2023survey, wu2024survey, yu2024evaluation}.

A particular feature of this domain is its high sensitivity to the wording of the answer. For some questions a brief factual restatement is acceptable, while for others an almost literal form of the answer is essential: an exact flag, path, field name, API version, or specific parameter value. The system must therefore be able to handle both ordinary documentation-grounded answers and cases that require near-literal reproduction; for documentation question answering this is directly tied to the requirements of factual accuracy and degree of support by the retrieved context \cite{yu2024evaluation, es2024ragas}.

\subsection{Problem statement}

We consider the task of building a RAG system that, given a user question about the Kubernetes documentation, must retrieve the relevant fragments of the corpus and generate an answer based only on the retrieved context. Such a formulation is in line with Retrieval-Augmented Generation for tasks that require access to an external knowledge base and with documentation question-answering systems \cite{lewis2020rag, gao2023survey, wu2024survey, yu2024evaluation}.

Let $D = \{d_1, \ldots, d_n\}$ denote the documentation corpus, represented as a set of text chunks, and let $R$ be the fixed retrieval-and-reranking pipeline that for a question $q$ selects the context $c = R(q, D)$. Let $x \in X$ denote the generator configuration inside this fixed retrieval pipeline, that is, the choice of base model, LoRA rank, and the scope of adapted modules. For each configuration $x$, the system maps a question $q$ to an answer $a_x(q, c)$, where $c$ is determined solely by the retrieval and reranking procedures, and the main research focus shifts to the generator adaptation and to comparing several adapter configurations \cite{karpukhin2020dpr, robertson1994bm25, robertson1995okapi, robertson2009bm25, zhao2024dense, qu2021rocketqa, santhanam2022colbertv2, chen2024m3embedding, nogueira2019passagereranking, yoon2024listt5, hu2022lora, dettmers2023qlora, hayou2024loraplus, han2024peft, wang2025peft, ye2023aspen}.

For each configuration we introduce a quality function $Q(x) = F1(x)$ and a cost vector $C(x) = (L_{\mathrm{inf}}(x), M_{\mathrm{inf}}(x), T_{\mathrm{train}}(x), M_{\mathrm{train}}(x))$, where $L_{\mathrm{inf}}$ is the mean inference latency, $M_{\mathrm{inf}}$ is the peak inference memory consumption, $T_{\mathrm{train}}$ is the total training time, and $M_{\mathrm{train}}$ is the peak training memory consumption. For a specific deployment scenario the choice of configuration can be written as a constrained optimization problem of the form $x^* = \arg\max Q(x)$ subject to constraints on one or several components of $C(x)$.

However, the main interest of this work is not in a single solution $x^*$, but in the set of Pareto-optimal configurations:
\[
\begin{aligned}
P = \{ x \in X \mid {} & \nexists x' \in X : Q(x') \geq Q(x), \\
& C_j(x') \leq C_j(x)\ \text{for all } j,\ \text{and at least one inequality is strict} \}.
\end{aligned}
\]

This formulation matches the practical logic of the task, because the applied value of a documentation-oriented RAG assistant is determined not only by the maximum answer quality, but also by conflicting constraints on latency, memory, and training cost; this logic is consistent with works on multi-criteria evaluation of RAG and LoRA \cite{yu2024evaluation, es2024ragas, baek2026lorabayesian, baqar2025raglora}.

\subsection{Task specifics}

The task under study has a number of features that make it non-trivial compared with ordinary generative question answering. First, answers are heterogeneous in nature: some questions require literal reproduction of a value, while others demand a concise but complete and correct description of a fact. Second, generation quality strongly depends on the quality of retrieval and reranking, because the generator is restricted to the provided context. An error at the retrieval stage directly affects the final answer even with a strong generator.

Third, for a practical system computational constraints are critical. Even if a more complex configuration improves quality, it may be unsuitable in real conditions due to increased latency or memory consumption. Fourth, tuning LoRA inside a RAG system does not reduce to standard fine-tuning of a language model: one must account for the interaction between the retrieval stage, the prompt-formation mode, and the generator adaptation scheme. This is precisely why the work uses a Pareto-based approach, which makes it possible to analyse not a single best point but a set of non-dominated configurations.

\section{Data and Corpus}
\label{app:data}

\subsection{Data source}

The data source is the Kubernetes documentation. The corpus was pre-cleaned and converted into a set of text fragments suitable for retrieval and for grounding the answer in the supporting context. During corpus preparation, the documentation pages were brought into a structured textual representation and then segmented into semantically meaningful chunks. This segmentation follows the general logic of building corpora suitable for retrieval in RAG systems \cite{gao2023survey, wu2024survey, yu2024evaluation, karpukhin2020dpr, chen2024m3embedding}.

\subsection{Construction of the QA sets}

The question-answer set was assembled in a combined way: 500 QA pairs were written by the author manually, while the remaining part was prepared using an AI agent based on the GPT-5.4 model (through the agent interface in the Cursor environment). Generation was performed one pair at a time: each question-answer pair was created individually from the Kubernetes documentation rather than obtained through bulk automated augmentation. The resulting set then underwent a full manual verification pass by the author, including verification of question correctness, reference answers, and their correspondence to the source fragments of the documentation. During verification, successive cleanup passes were applied to the whole set; a pair was discarded if the question was malformed or ambiguous, if no answer was present in the documentation corpus, if it was a duplicate of another pair, or if the reference answer contained a factual error. As a result of the cleanup passes, 323 pairs out of an initial pool of 5467 QA candidates were rejected (about 5.9\%), and approximately 20\% of the remaining pairs required editing of the wording or rephrasing of the reference answer to match the format of short documentation answers; the remaining pairs were accepted without changes. This logic is consistent with modern approaches to evaluating RAG systems, where both the quality of the answer and its support by the retrieved context matter \cite{yu2024evaluation, es2024ragas}. The final verified set of 5144 pairs (500 written manually from scratch and 4644 automatically generated and manually verified) was split into training, validation, and test parts. In the current configuration of the work we use the train, eval, and test splits, intended respectively for training adapters, selecting and comparing configurations, and the final evaluation.

In addition, it is worth recording the relationship between the model used to generate QA candidates (GPT-5.4) and the model used as the judge in \S\ref{sec:eval} (\texttt{gpt-5.4-mini}). Although both models belong to the same family, no direct train/test leak arises between them: the judge was not trained on the gold answers, and the reference answers themselves underwent a full manual verification by the author. The residual risk of a same-family bias is discussed separately in Appendix~\ref{sec:limitations} as a limitation of the judge-based metrics.

Thus the work produces not only a question-answer set, but an integrated evaluation benchmark: the documentation corpus, the QA set, the train/eval/test split, the metric system, and a unified protocol for ablation and Pareto experiments. This makes the resulting set usable not only for training and testing individual adapters, but also for the reproducible comparison of retrieval and LoRA configurations within a common experimental setting.

\subsection{Answer types}

In the QA set two answer types are distinguished: \texttt{exact} and \texttt{normal}. The \texttt{exact} class covers cases in which the almost-literal form of the answer matters, for example flags, paths, field names, API version, feature gates, and other short literal values. The \texttt{normal} class covers ordinary factual answers, grounded in the documentation and admitting careful rephrasing while preserving meaning. This distinction is introduced in the present work as part of our own methodology for annotating and analysing answers.

This annotation was used both for manual analysis of the QA set and during adapter training. It is important to emphasize that in the current main inference regime the generator does not receive the answer-type label (\texttt{exact} or \texttt{normal}) as an input signal and operates in a neutral prompt-formation mode. The answer-type annotation is therefore used in the work primarily as a supervision signal during training and as a source of stratified quality metrics, but not as an explicit condition at test-time inference; this scheme methodologically aligns with the separation of training and testing regimes in RAG-evaluation works \cite{yu2024evaluation, es2024ragas}.

\subsection{Dataset characteristics}

The final sizes of the splits are given in Table~\ref{tab:app-dataset_splits}.

\begin{table}[H]
\centering
\begin{tabular}{cccc}
\toprule
Split & Rows & Exact & Normal \\
\midrule
Train & 3614 & 1449 & 2165 \\
Eval & 745 & 361 & 384 \\
Test & 785 & 475 & 310 \\
\bottomrule
\end{tabular}
\caption{Sizes of the training, validation, and test parts of the QA set}
\label{tab:app-dataset_splits}
\end{table}

Table~\ref{tab:app-dataset_splits} records the sizes of the training, validation, and test parts of the set, as well as the distribution of answer types within each part. These characteristics are used further as a description of the experimental base on which adapter training and final configuration evaluation are performed.

\section{RAG System Architecture}
\label{app:pipeline}

\subsection{Overall solution scheme}

The system under study is a Retrieval-Augmented Generation pipeline for question answering over technical documentation. Given a user question, the system first retrieves a set of candidates from the documentation corpus, then reranks the retrieved fragments, and finally passes the selected context to the generator. This three-step scheme of retrieval, reranking, and generation matches modern views on RAG systems \cite{lewis2020rag, gao2023survey, wu2024survey, yu2024evaluation}.

\subsection{Retrieval component}

In the current main configuration we use a hybrid retrieval pipeline: dense retrieval based on a FAISS index is combined with native sparse retrieval based on BGE-M3, and the results are merged using Reciprocal Rank Fusion. This scheme draws on works on dense retrieval, BM25, hybrid retrieval, and result fusion \cite{karpukhin2020dpr, robertson1994bm25, robertson1995okapi, robertson2009bm25, zhao2024dense, qu2021rocketqa, santhanam2022colbertv2, chen2024m3embedding, douze2024faiss, johnson2019billionscale, cormack2009rrf, li2023llmembedder}.

\subsection{Ranking and reranking}

After the initial retrieval step the candidates additionally pass through reranking. In the system under study we use the pretrained reranker \texttt{BAAI/bge-reranker-v2-m3}. The very idea of reranking goes back to classical works on cross-encoder and listwise reranking \cite{nogueira2019passagereranking, yoon2024listt5}; for the BGE family the work on dense retrieval based on large language models \cite{li2023llmembedder} is also relevant.

\subsection{Generator and prompting}

We consider two instruction-tuned models of the Llama 3 family as generators: \texttt{meta-llama/Llama-3.2-3B-Instruct} and \texttt{meta-llama/Llama-3.1-8B-Instruct}. Both models belong to successive releases of the Llama 3 family \cite{dubey2024llama3} -- versions 3.2 and 3.1, respectively; they share the common architectural line of decoder-only transformers with RoPE, GQA attention, and SwiGLU blocks described in \cite{dubey2024llama3}. In the main inference regime we use a neutral prompt-formation mode that minimizes explicit control over answer style.

This prompt-formation mode allows adapters to be evaluated in a single inference regime and thus compared on a fairer basis. At training time, the option to condition on the answer type is preserved, which is important for handling both literal and ordinary factual answers; such a separation of training and testing regimes methodologically aligns with works on RAG evaluation \cite{yu2024evaluation, es2024ragas}.

\section{LoRA Adaptation and Experimental Design}
\label{app:lora}

\subsection{Motivation for using LoRA}

Full fine-tuning of modern instruction-tuned language models is costly both in memory and in time. We therefore use parameter-efficient fine-tuning, namely LoRA. This choice rests directly on works on LoRA, QLoRA, and general PEFT surveys \cite{hu2022lora, dettmers2023qlora, hayou2024loraplus, han2024peft, wang2025peft}. For our research task, in which several adaptation configurations have to be compared, this approach is particularly convenient.

\subsection{Trained configurations}

The current set of training artefacts comprises two baseline models and 20 trained LoRA configurations. For each base model we considered LoRA configurations with several rank values and two attention-module coverage schemes: adaptation of the query and value projections only (\texttt{q\_proj}, \texttt{v\_proj}) and adaptation of all main projections in the attention block (\texttt{q\_proj}, \texttt{k\_proj}, \texttt{v\_proj}, \texttt{o\_proj}).

\begin{table}[H]
\centering
\resizebox{\textwidth}{!}{
\begin{tabular}{cccc}
\toprule
Base model & Ranks & Adaptation schemes & Number of LoRA configs \\
\midrule
3B & \texttt{4, 8, 16, 32, 64} & \texttt{qv\_only}, \texttt{full\_attention} & 10 \\
8B & \texttt{4, 8, 16, 32, 64} & \texttt{qv\_only}, \texttt{full\_attention} & 10 \\
\bottomrule
\end{tabular}
}
\caption{Composition of the trained LoRA configurations: base models, rank options, and adaptation schemes}
\label{tab:app-lora_configurations}
\end{table}

The baseline configurations \texttt{3B baseline} and \texttt{8B baseline} were used as LoRA-free reference points and were compared against the trained adapters on the same evaluation metrics.

\subsubsection{Baseline evaluation protocol}

For the baseline configurations we used the same inference protocol as for the main LoRA-adapter comparisons. The evaluation was performed in the neutral prompting mode, without any additional explicit requirement of a grounded-answer style in the prompt. On the retrieval side we used the same base pipeline as in the main \texttt{01\_base\_\_neutral} regime: dense retrieval, native sparse retrieval, candidate fusion via reciprocal rank fusion, and subsequent reranking with the pretrained reranker.

The number of context fragments at inference time was also kept fixed and matched the main comparison regime: \texttt{top\_k = 2}. Thanks to this, the comparison of \texttt{3B baseline} / \texttt{8B baseline} with the LoRA configurations was performed in the same retrieval and prompting pipeline in which the main results of \S\ref{sec:results} are interpreted, rather than in a separate special regime. This is important for the correctness of the subsequent comparisons, including the discussion of the closeness between \texttt{3B r64 qv\_only} and \texttt{8B baseline} on \texttt{F1}.

\subsection{Fixed training hyperparameters}

\begin{table}[H]
\centering
\begin{tabular}{ccc}
\toprule
Parameter & Value & Comment \\
\midrule
\texttt{embed\_top\_k} & \texttt{2} & Number of context fragments in a training example \\
\texttt{num\_train\_epochs} & \texttt{8} & Total number of epochs \\
\texttt{learning\_rate} & \texttt{2e-5} & Single value used for all runs \\
\bottomrule
\end{tabular}
\caption{Fixed training hyperparameters shared across all LoRA configurations}
\label{tab:app-training_hyperparams}
\end{table}

In the current set of training runs a single scheme was used to form the training context. Across all configurations the number of retrieved context fragments was fixed as \texttt{embed\_top\_k = 2}, that is, every training example was built on top of two context fragments.

All runs in the current set used \texttt{num\_train\_epochs = 8} and \texttt{learning\_rate = 2e-5}. The base optimization scheme follows the standard PEFT setup: the AdamW optimizer \cite{kingma2015adam, loshchilov2019adamw} with a cosine schedule on the learning rate \cite{loshchilov2017sgdr} and a linear warm-up over approximately ${\sim}3\%$ of the total training steps; mixed precision is set to \texttt{bf16}. Compared configurations were therefore trained in a comparable regime, and any differences in quality and cost can be attributed primarily to the adapter parameters rather than to changes in the optimization regime.

\subsection{LoRA adaptation parameters}

For all runs we used \texttt{bias = none}, \texttt{task\_type = CAUSAL\_LM}, and \texttt{lora\_dropout = 0.05}. The \texttt{lora\_alpha} coefficient was chosen proportional to the rank via the rule \texttt{alpha = 2r}. As a result, when comparing configurations we varied primarily the model scale, the rank, and the scope of adapted modules, while the remaining LoRA parameters stayed fixed.

\begin{table}[H]
\centering
\begin{tabular}{cc}
\toprule
Configuration aspect & Value \\
\midrule
\texttt{r = 4} & \texttt{lora\_alpha = 8} \\
\texttt{r = 8} & \texttt{lora\_alpha = 16} \\
\texttt{r = 16} & \texttt{lora\_alpha = 32} \\
\texttt{r = 32} & \texttt{lora\_alpha = 64} \\
\texttt{r = 64} & \texttt{lora\_alpha = 128} \\
\texttt{lora\_dropout} & \texttt{0.05} \\
\texttt{qv\_only} & \texttt{q\_proj}, \texttt{v\_proj} \\
\texttt{full\_attention} & \texttt{q\_proj}, \texttt{k\_proj}, \texttt{v\_proj}, \texttt{o\_proj} \\
\bottomrule
\end{tabular}
\caption{LoRA adaptation parameters: correspondence of rank to \texttt{lora\_alpha}, dropout, and the attention-module coverage for the two compared schemes}
\label{tab:app-lora_params}
\end{table}

In terms of the scope of adapted modules, two main schemes were used in the current experimental set. The \texttt{qv\_only} configuration adapts only the \texttt{q\_proj} and \texttt{v\_proj} projections. The \texttt{full\_attention} configuration extends adaptation to all main attention projections: \texttt{q\_proj}, \texttt{k\_proj}, \texttt{v\_proj}, and \texttt{o\_proj}. This comparison makes it possible to assess whether wider attention-block coverage justifies the additional training and inference cost inside the same RAG architecture.

\subsection{Training-context formation regime}
\label{app:training-context}

Adapter training does not rely on a single context regime. We use a mixed scheme for forming the training example. If the example has pre-labelled supporting chunks, exactly those are fed as the supporting context. If no such labelling exists, the context is formed by the retrieval procedure. This scheme is methodologically close to the retrieval-augmented fine-tuning settings described in JORA \cite{tahir2024jora} and helps reduce the gap between idealized training and the actual inference regime.

\subsection{Hardware configuration and inference stack}
\label{app:inference_setup}

All runtime cost metrics (\texttt{Latency}, \texttt{Inference VRAM}, \texttt{Training Time}, \texttt{Training VRAM}) were measured in a single compute environment, so as to remove any contribution of hardware to the relative differences between configurations. The measurements were performed in Yandex DataSphere on a single compute node with the following characteristics: 1$\times$ NVIDIA A100 40~GB, 28~vCPU, and 114~GB of RAM. The same node was used both for training the LoRA adapters and for subsequent inference; thanks to this \texttt{Training Time}, \texttt{Training VRAM}, \texttt{Latency}, and \texttt{Inference VRAM} remain comparable across the entire set of experiments.

The inference stack is based on Hugging Face Transformers and PEFT \cite{dettmers2023qlora, hayou2024loraplus}: the base model (\texttt{meta-llama/Llama-3.2-3B-Instruct} or \texttt{meta-llama/Llama-3.1-8B-Instruct}) is loaded in the original precision without quantization (the flags \texttt{-{}-no-quant-generator} and \texttt{-{}-no-quant-judge}), and the LoRA adapter is loaded separately on top of the base model via \texttt{PeftModel.from\_pretrained}; the mixed-precision setting matches \texttt{bf16}, as during training (see \S\ref{app:training-context}). The retrieval pipeline uses \texttt{BAAI/bge-m3} for dense embeddings and \texttt{BAAI/bge-reranker-v2-m3} for final reranking (\texttt{reranker\_batch\_size = 16}, \texttt{retrieve\_top\_n = 20}); in the main comparison regime the generator operates with a fixed \texttt{eval\_top\_k = 2} retrieved fragments, which matches the training-context formation scheme (\texttt{embed\_top\_k = 2}, see \S\ref{app:training-context}); additionally, for sensitivity control to the parameter, runs were performed with \texttt{eval\_top\_k} $\in \{1, 4\}$. Generation is performed in greedy decoding (no sampling) with the maximum answer length fixed in the code and identical across all configurations.

Latency is measured per sample (the effective generator batch size equals one), as the mean time of the complete processing of a single test example -- from receiving the query to returning the generated answer -- over the entire test split \texttt{n = 785}, and includes all pipeline stages: building the query embedding, dense and sparse retrieval, reranking, prompt formation, and answer generation. \texttt{Inference VRAM} is recorded as the peak GPU memory consumption on the device, obtained via \texttt{torch.cuda.max\_memory\_allocated} at the end of the run.

\subsection{Experimental hypotheses}

The work tests the following hypotheses:

\begin{itemize}
\item scaling the model up improves quality but raises latency and memory consumption;
\item increasing the rank expands the adaptation capacity, but does not always yield a proportional gain in quality;
\item a wider coverage of adapted modules is not necessarily preferable once training and inference cost are accounted for;
\item among the set of configurations there exist practical working points that do not maximize quality but offer the best balance between quality and cost.
\end{itemize}

\section{Evaluation Methodology}
\label{app:eval}

\subsection{Quality metrics}

The main quality metric in the work is token-level F1, used to compare configurations by the degree of agreement between the answer and the reference \cite{yu2024evaluation}. The final \texttt{F1} values, the judge-based metrics, the summary tables of \S\ref{sec:results} and the corresponding plots are computed on the final test split (\texttt{n = 785}). The eval split (\texttt{n = 745}) is used for selection and intermediate comparison of configurations during the experimental cycle. In addition to point estimates we also compute 95\% bootstrap confidence intervals on the test set (1000 resamples). For key configuration comparisons we use the paired bootstrap for the difference $\Delta F1$, which separates stable differences from narrow gaps lying within sampling noise. Semantic embedding-based metrics such as BERTScore \cite{zhang2020bertscore} are not used in the main setting of the work: for documentation-oriented question answering with a high share of exact questions, token-level F1 provides a more direct and interpretable comparison with the reference, while the semantic dimension of quality is covered separately by the judge-based correctness and groundedness scores (see Section~5.2).

\subsection{Judge-based groundedness and correctness}

In addition, we use judge-based evaluation of groundedness and correctness. In this scheme an external judge language model \texttt{gpt-5.4-mini} receives the question, the retrieved context, and the generated answer, after which it assigns two independent scores: correctness as the degree of semantic correctness of the answer relative to the provided context, and groundedness as the degree to which the answer is supported by the retrieved fragments without unsupported additions. Evaluation is performed in a blind regime: the judge receives only the triple \texttt{(question, context, answer)} and does not receive the \texttt{model\_id}, the configuration name, or any other generator metadata. This evaluation is used not as a replacement for the main \texttt{F1} metric, but as an additional axis of quality that supports the analysis of supportedness of the answer by the retrieved context and of its faithfulness to that context \cite{yu2024evaluation, es2024ragas, baqar2025raglora}.

The main aggregates of the judge-based evaluation are \texttt{correctness\_pass@4} and \texttt{groundedness\_pass@4}, that is, the fraction of answers that received a score of at least 4 on the corresponding scale. These metrics are especially useful when several configurations have similar \texttt{F1} values but differ in the degree of context support. In particular, the subsequent ablation experiments show that the configuration optimal in \texttt{F1} does not necessarily coincide with the configuration optimal in groundedness, which indicates a stable trade-off between task quality and supportedness.

\subsection{Cost metrics}

To analyse computational cost we use both deployment and training metrics. The former include the mean answer latency and the peak GPU memory consumption at inference. The latter include the total training time and the peak GPU memory consumption at training. Such a set reflects the idea of jointly analysing quality and cost, emphasized in works on the efficiency evaluation of RAG and LoRA \cite{es2024ragas, baek2026lorabayesian, baqar2025raglora}.

\subsection{Pareto analysis}

To compare configurations in a multi-criteria setting we use Pareto analysis. A configuration is considered non-dominated if no other configuration is no worse in quality and at the same time no worse in cost, while being strictly better in at least one dimension. In two-dimensional fronts the role of the quality function is played by \texttt{F1}, while the role of cost is taken in turn by the mean inference latency, inference memory, training time, and training memory. This approach makes it possible to identify practical working points and to avoid simplistic model selection based on a single metric; the multi-criteria logic of such a comparison is consistent with works in which quality and cost are analysed jointly \cite{baek2026lorabayesian, baqar2025raglora}.

\section{Experimental Results}
\label{app:results}

In this appendix the results of comparing baseline systems and LoRA adapters are interpreted as multi-criteria trade-offs between quality, latency, memory, and training cost. All summary results in Tables~\ref{tab:app-runtime_front}, \ref{tab:app-training_front}, \ref{tab:app-ablation_summary} and \ref{tab:app-paired_f1_deltas}, and in the corresponding plots, are computed on the final test split (\texttt{n = 785}).

\subsection{Quality and latency}

\begin{figure}[H]
\centering
\includegraphics[width=0.9\textwidth]{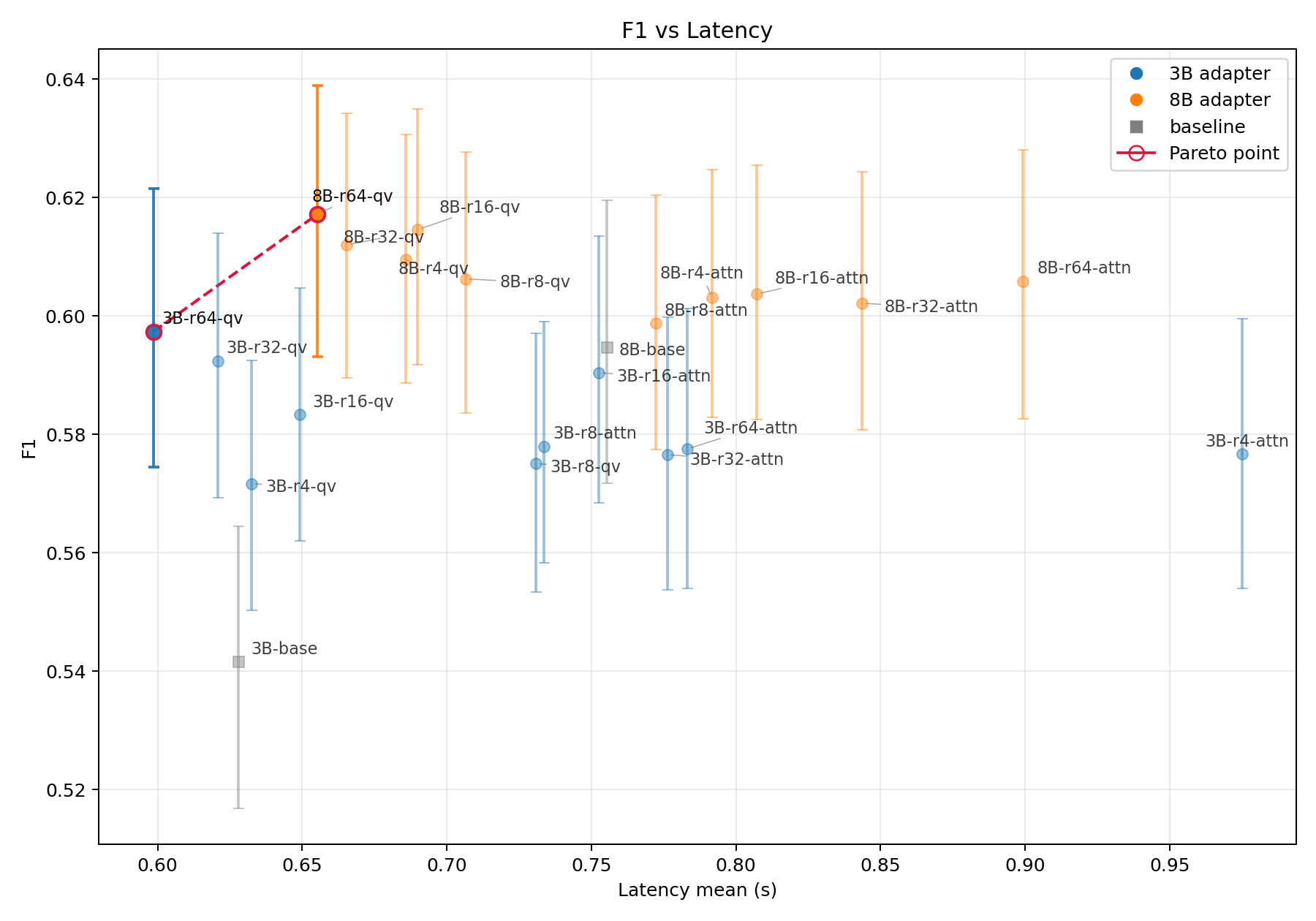}
\caption{F1 vs Latency}
\label{fig:app-f1_vs_latency}
\end{figure}

On the main response-time front the non-dominated configurations are \texttt{3B r64 qv\_only} and \texttt{8B r64 qv\_only}. The first represents the strongest lightweight point, and the second yields the highest quality in the current set of experiments. These points are listed in Table~\ref{tab:app-runtime_front}.

\begin{table}[H]
\centering
\begin{tabular}{ccccc}
\toprule
Configuration & F1 [95\% CI] & EM & Latency, s & Inference VRAM, GB \\
\midrule
\texttt{3B r64 qv\_only} & 0.597 [0.574, 0.622] & 0.250 & 0.598 & 12.760 \\
\texttt{8B r64 qv\_only} & 0.617 [0.593, 0.639] & 0.250 & 0.655 & 21.926 \\
\bottomrule
\end{tabular}
\caption{Non-dominated points on the quality--latency front}
\label{tab:app-runtime_front}
\end{table}

Table~\ref{tab:app-runtime_front} shows that the main response-time front (the runtime front) in the base regime consists of two points, between which the main practical trade-off lies. The transition from \texttt{3B r64 qv\_only} to \texttt{8B r64 qv\_only} brings approximately a $0.020$ gain in \texttt{F1}, and the paired bootstrap for $\Delta F1$ gives a 95\% CI of \texttt{[+0.0005, +0.0410]}. This indicates a small but test-set-supported advantage of the 8B configuration in \texttt{F1}, which at the same time comes with a transition to a substantially more expensive inference-memory regime and a latency increase of approximately $0.057$~s. Importantly, \texttt{3B r64 qv\_only} with \texttt{F1 = 0.597 [0.574, 0.622]} turns out to be statistically comparable to the unadapted \texttt{8B baseline} configuration (\texttt{0.595 [0.572, 0.620]}): for $\Delta F1$ the 95\% CI is \texttt{[-0.021, +0.026]}. The choice between these points is therefore determined not only by quality, but also by whether the system needs a lightweight 3B regime or a more expensive 8B regime with the highest point quality.

\subsection{Quality and inference memory}

\begin{figure}[H]
\centering
\includegraphics[width=0.9\textwidth]{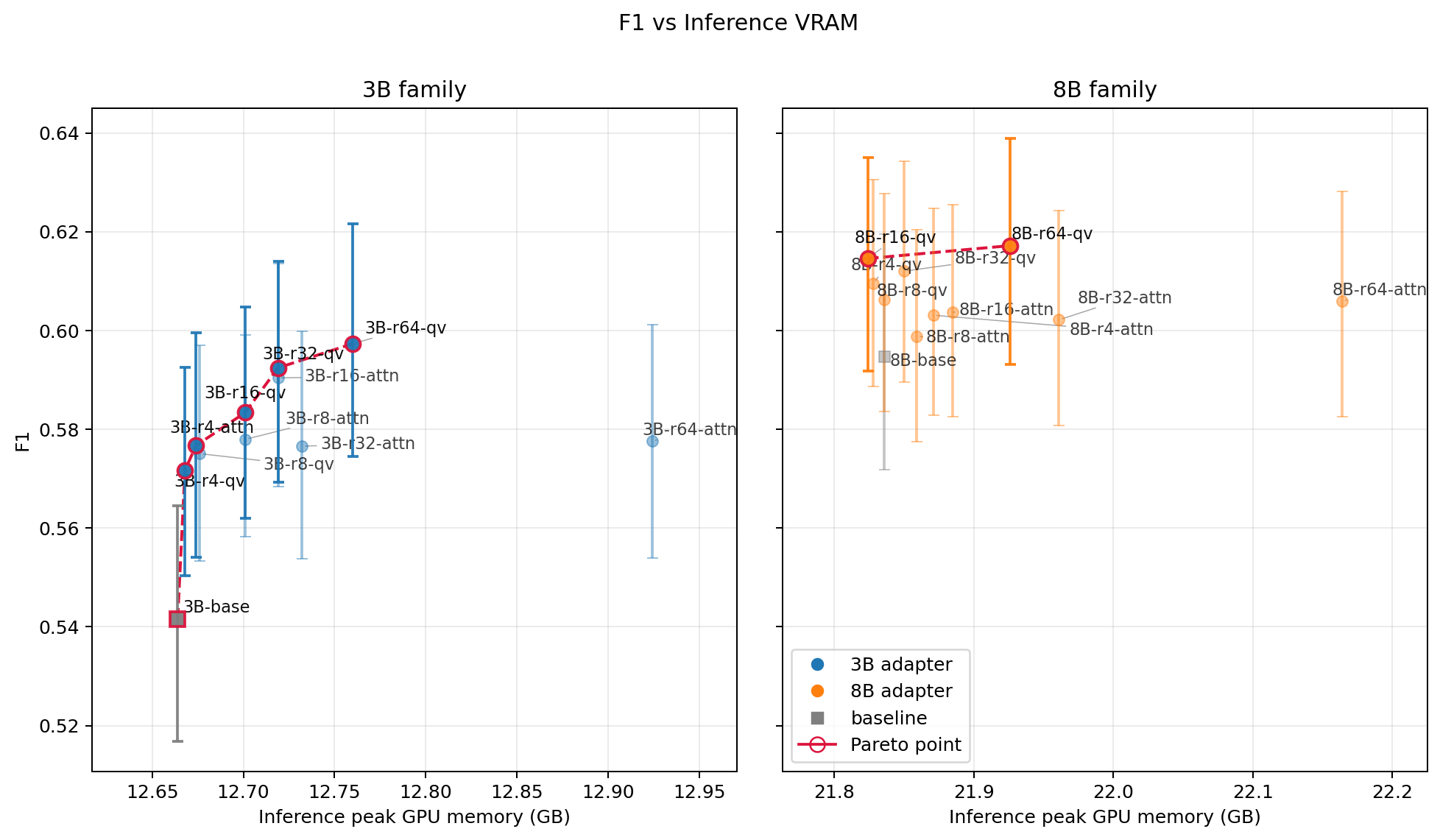}
\caption{F1 vs Inference VRAM}
\label{fig:app-f1_vs_inference_vram}
\end{figure}

The inference-memory plot shows a clear separation between the 3B and 8B families. The 3B configurations occupy a more compact area in memory and therefore remain most attractive for constrained hardware.

\subsection{Training cost}

\begin{figure}[H]
\centering
\includegraphics[width=0.9\textwidth]{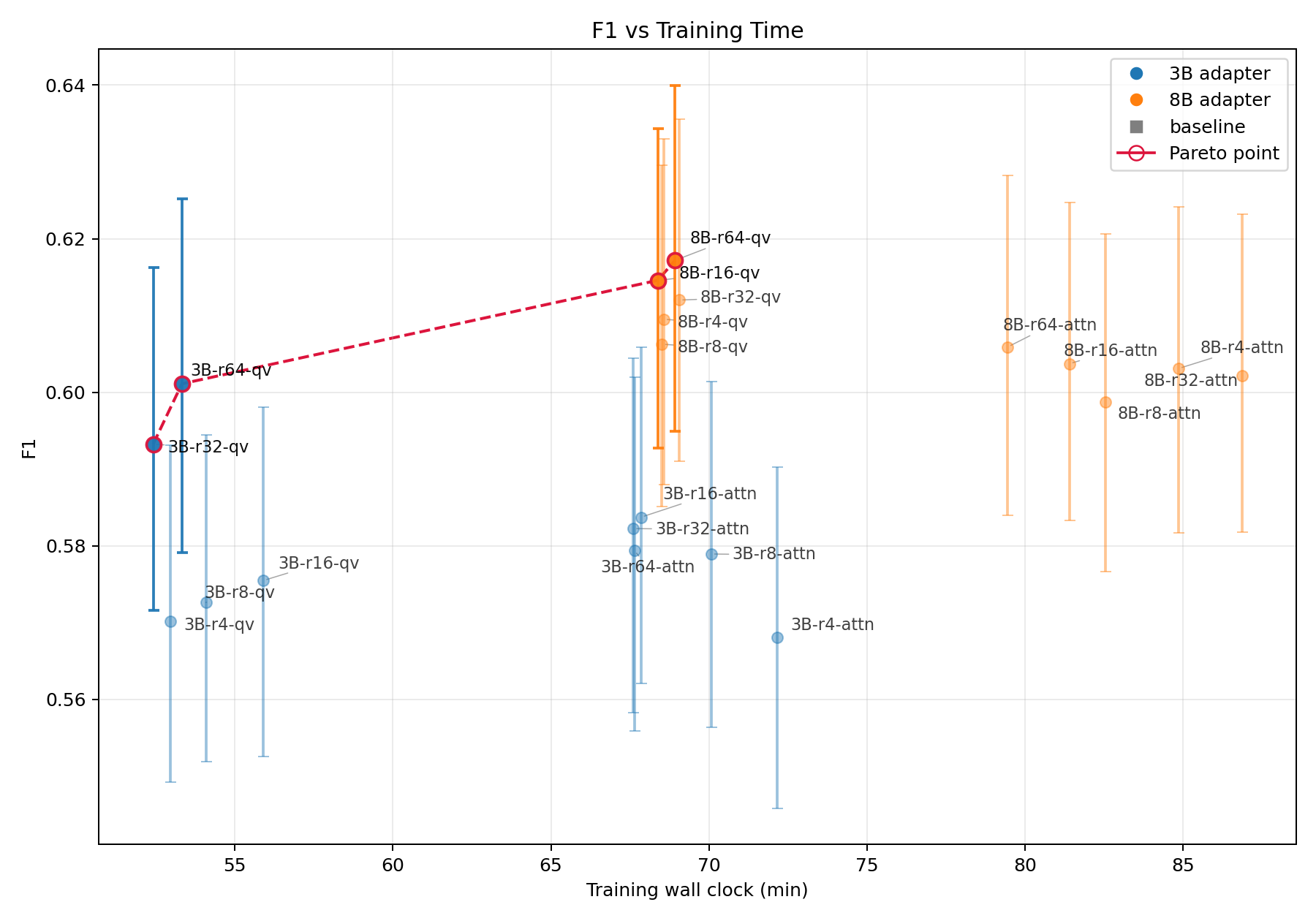}
\caption{F1 vs Training Time}
\label{fig:app-f1_vs_training_time}
\end{figure}

\begin{figure}[H]
\centering
\includegraphics[width=0.9\textwidth]{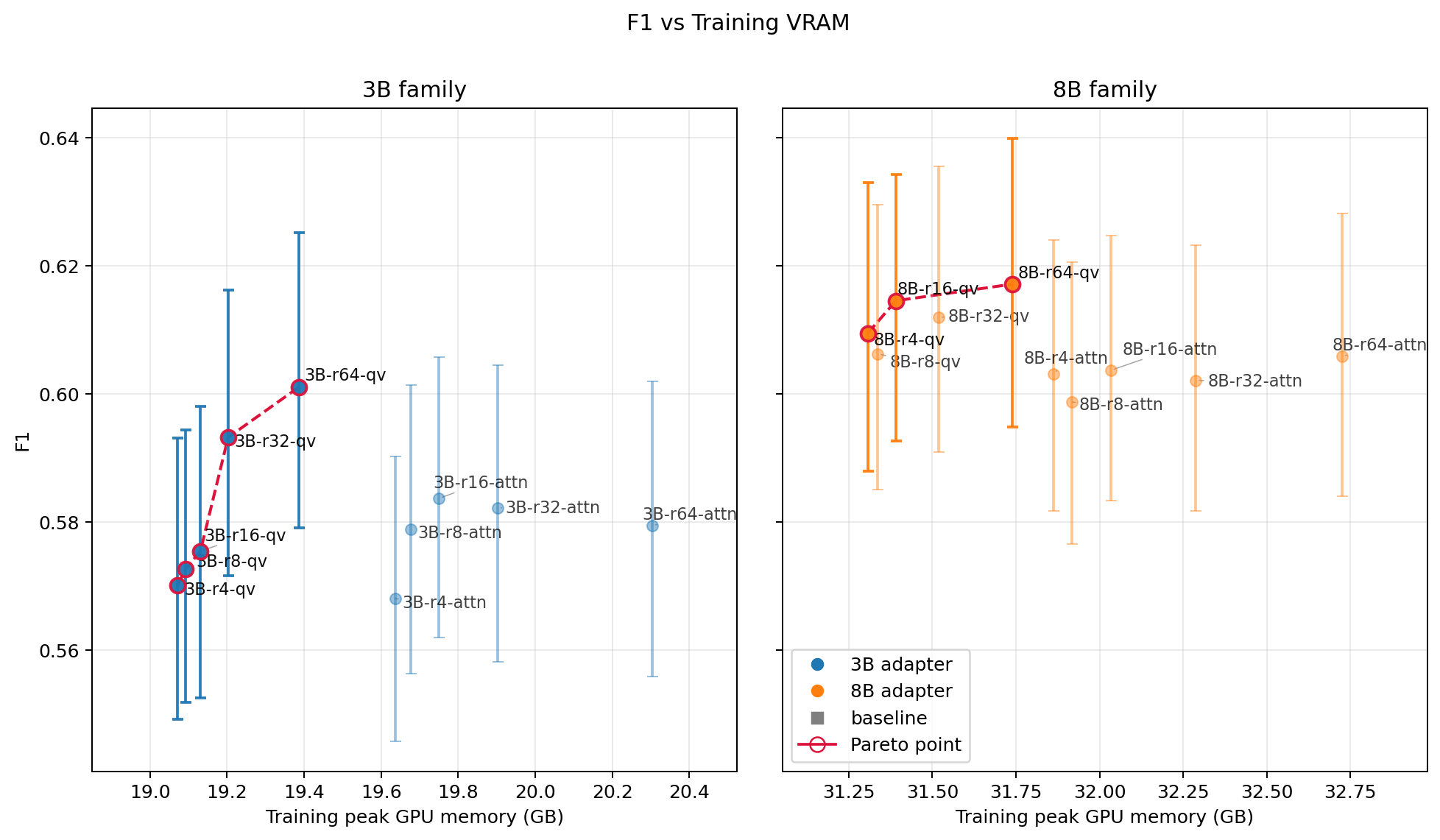}
\caption{F1 vs Training VRAM}
\label{fig:app-f1_vs_training_vram}
\end{figure}

In terms of training time and training memory the cheapest point remains \texttt{3B r4 qv\_only}, however the current training Pareto fronts are no longer limited to the lowest ranks. The strongest training-side trade-offs form around \texttt{qv\_only} configurations with ranks \texttt{r = 32} and \texttt{r = 64} in the 3B family and \texttt{r = 16}/\texttt{r = 64} in the 8B family. The composition of the non-dominated training-front points is given in Table~\ref{tab:app-training_front}.

\begin{table}[H]
\centering
\resizebox{\textwidth}{!}{
\begin{tabular}{ccccc}
\toprule
Configuration & F1 [95\% CI] & Training time, min & Train VRAM, GB & Front \\
\midrule
\texttt{3B r4 qv\_only} & 0.572 [0.550, 0.592] & 52.95 & 19.072 & vram \\
\texttt{3B r8 qv\_only} & 0.573 [0.552, 0.594] & 54.08 & 19.092 & vram \\
\texttt{3B r16 qv\_only} & 0.583 [0.562, 0.605] & 55.90 & 19.131 & vram \\
\texttt{3B r32 qv\_only} & 0.592 [0.569, 0.614] & 52.41 & 19.203 & time, vram \\
\texttt{3B r64 qv\_only} & 0.597 [0.574, 0.622] & 53.32 & 19.387 & time, vram \\
\texttt{8B r4 qv\_only} & 0.610 [0.588, 0.633] & 68.57 & 31.307 & vram \\
\texttt{8B r16 qv\_only} & 0.615 [0.593, 0.634] & 68.38 & 31.391 & time, vram \\
\texttt{8B r64 qv\_only} & 0.617 [0.595, 0.640] & 68.93 & 31.740 & time, vram \\
\bottomrule
\end{tabular}
}
\caption{Non-dominated points by training criteria}
\label{tab:app-training_front}
\end{table}

Table~\ref{tab:app-training_front} shows that inside the 3B family the move to \texttt{r = 32} and \texttt{r = 64} is accompanied by a noticeable improvement in \texttt{F1}, while training time and memory grow only moderately compared to the lower ranks. At the same time, \texttt{full\_attention} configurations do not appear at all on the current training Pareto fronts, which further emphasizes the structural advantage of \texttt{qv\_only}. For the 8B family the differences in training time between \texttt{qv\_only} configurations are also small relative to the overall backbone cost, while the upper points \texttt{r = 16} and \texttt{r = 64} have almost fully overlapping \texttt{F1} intervals, so the choice between them is driven more by additional criteria than by a stable quality gap.

\subsection{Judge-based groundedness as a second quality axis}

\begin{figure}[H]
\centering
\includegraphics[width=0.9\textwidth]{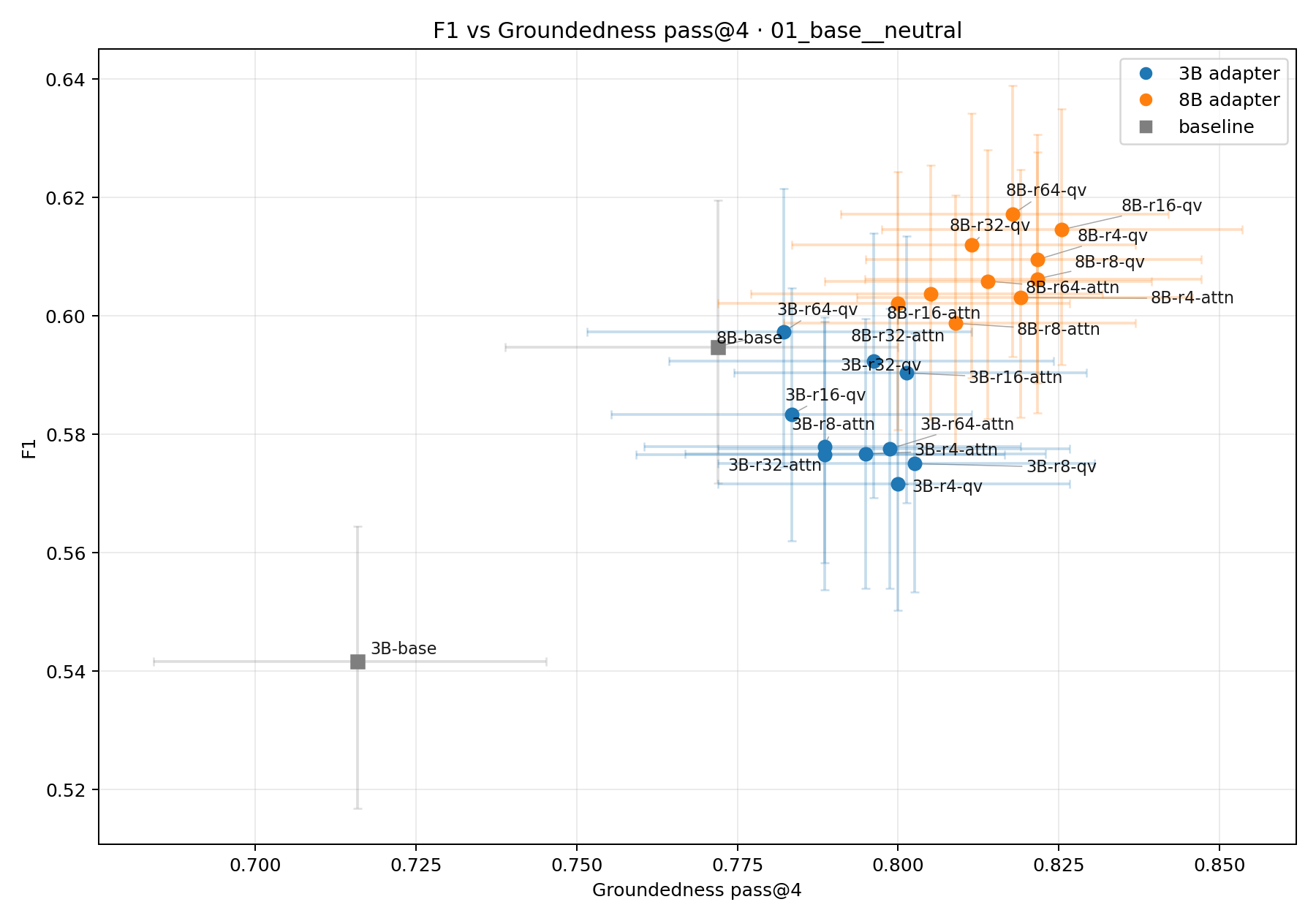}
\caption{F1 vs Groundedness pass@4}
\label{fig:app-f1_vs_groundedness_pass4}
\end{figure}

Alongside the main task metric \texttt{F1}, the work additionally considered a judge-based groundedness evaluation, aggregated as \texttt{groundedness\_pass@4}. This metric measures how often the model produces answers that are sufficiently grounded in the retrieved context and contain no unsupported additions. Unlike \texttt{F1}, it captures not just agreement with the reference answer but also the degree of supportedness of the result inside the retrieval-augmented pipeline.

The plot shows that configurations with the best point \texttt{F1} do not necessarily coincide with configurations optimal in \texttt{groundedness\_pass@4}. In the base regime the best-\texttt{F1} configuration \texttt{8B r64 qv\_only} attains \texttt{F1 = 0.617 [0.593, 0.639]}, while the groundedness maximum (\texttt{0.825}) is achieved by \texttt{8B r16 qv\_only} at \texttt{F1 = 0.615 [0.592, 0.635]}. For the paired difference $\Delta F1$ between them the 95\% CI is \texttt{[-0.011, +0.016]}, that is, on \texttt{F1} these two points are statistically indistinguishable on the current test set. This means that a gain in task quality is not equivalent to a gain in the factual support of the answer by the context. Consequently, for documentation-oriented question answering the quality of the system cannot be fully described by \texttt{F1} alone: some configurations reach higher groundedness at a statistically comparable level of the main metric.

\begin{figure}[H]
\centering
\includegraphics[width=0.9\textwidth]{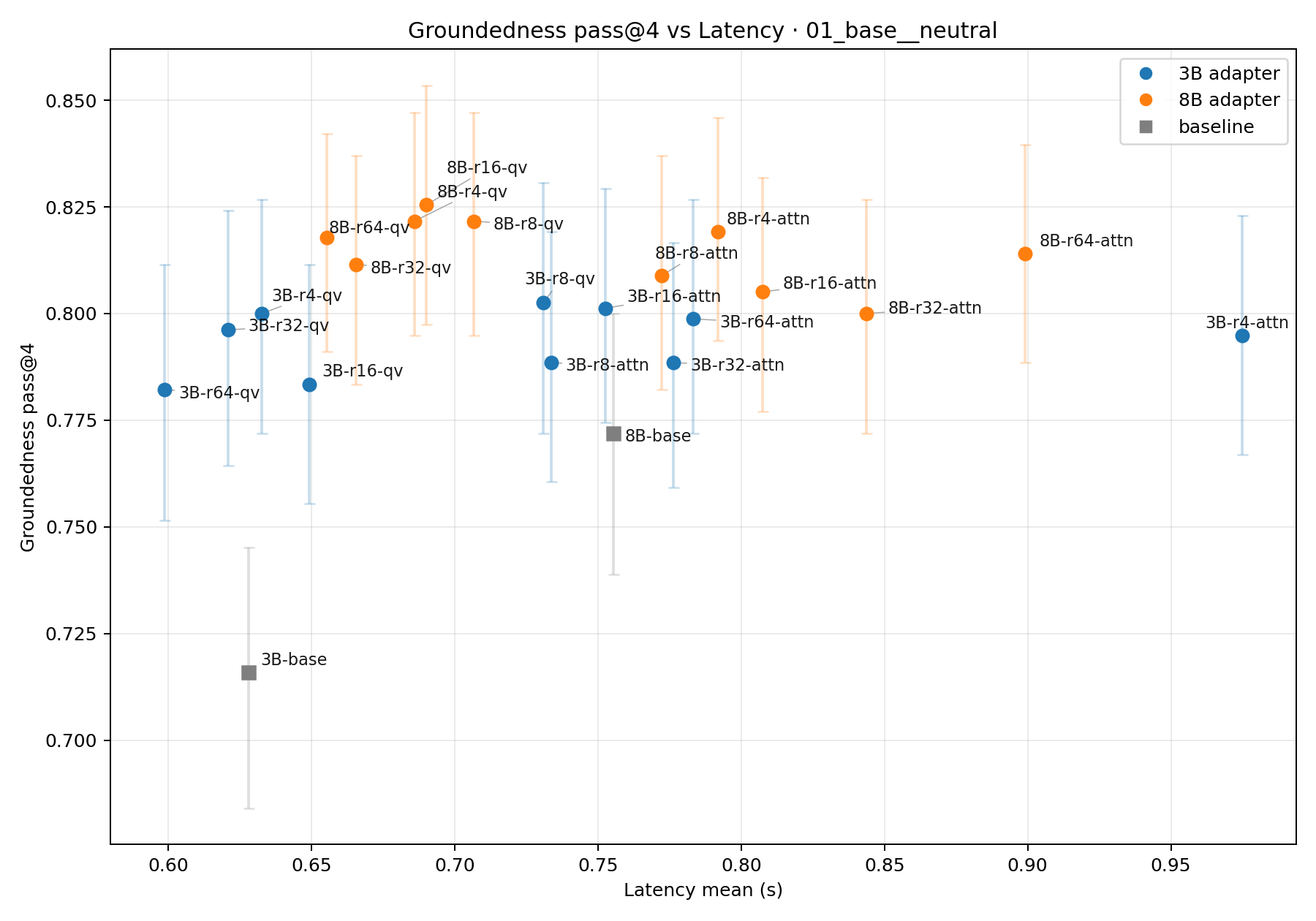}
\caption{Groundedness pass@4 vs Latency}
\label{fig:app-groundedness_pass4_vs_latency}
\end{figure}

The comparison of \texttt{groundedness\_pass@4} against latency further shows that the LoRA-free baseline configurations (\texttt{3B-base}, \texttt{8B-base}) systematically lie below all adapted points: at comparable latency, LoRA fine-tuning shifts \texttt{groundedness\_pass@4} upward by 0.05--0.08, and in absolute value this shift exceeds the width of the 95\% CI ($\approx 0.03$). Within already-adapted configurations the \texttt{groundedness\_pass@4} spread mostly falls inside the CI band, and the separation is driven mainly by the 3B/8B split. Latency and groundedness therefore do not form a non-trivial front beyond the one already visible on \texttt{F1}, but fine-tuning remains a necessary condition for a high level of supportedness. The resource view by inference VRAM carries no additional information either: within each family the VRAM spread does not exceed 0.3--0.4~GB, while the corresponding \texttt{groundedness} spread fully lies within the 95\% CI band, so the corresponding projection is not shown.

Thus the judge-based analysis confirms the overall multi-criteria logic of the work. Practically useful configurations differ not only in \texttt{F1} and resource cost, but also in the degree of supportedness, so the choice of working point depends on what exactly is considered the priority in the applied scenario: maximum agreement with the reference, stricter groundedness, or a trade-off between them.

\subsection{Ablation experiments on retrieval and prompting}

To check the robustness of the conclusions about LoRA configurations, additional ablation experiments were conducted in which not only the adapter parameters but also other parts of the pipeline were varied: the retrieval pipeline and the prompt-formation mode. We considered ten regimes obtained by combining five retrieval settings (\texttt{base}, \texttt{reranker\_off}, \texttt{dense\_only}, \texttt{sparse\_only}, \texttt{hybrid\_bm25}) and two prompting modes (\texttt{neutral}, \texttt{explicit\_grounded}). For each regime two reference points were extracted from the 22 configurations: the configuration with the maximum \texttt{F1} and the configuration with the maximum \texttt{groundedness\_pass@4}.

The base regime (\texttt{base}) used the full retrieval pipeline, including dense retrieval, native sparse retrieval, their fusion via reciprocal rank fusion, and subsequent reranking of the candidates by the pretrained reranker. The \texttt{reranker\_off} regime excludes only the final reranking step while keeping hybrid retrieval. The \texttt{dense\_only} regime disables the sparse branch and uses only dense retrieval, while \texttt{sparse\_only}, conversely, excludes the dense component and uses only sparse retrieval. The \texttt{hybrid\_bm25} regime preserves the hybrid scheme but replaces native sparse retrieval with classical BM25, which makes it possible to compare two variants of the sparse component inside the same retrieval pipeline. On top of the retrieval ablations, two prompting modes were considered: \texttt{neutral}, corresponding to a neutral generation mode without explicitly requiring the answer to rely on the context, and \texttt{explicit\_grounded}, in which the prompt explicitly emphasizes the need to answer strictly based on the retrieved material.

\begin{figure}[H]
\centering
\includegraphics[width=0.85\textwidth]{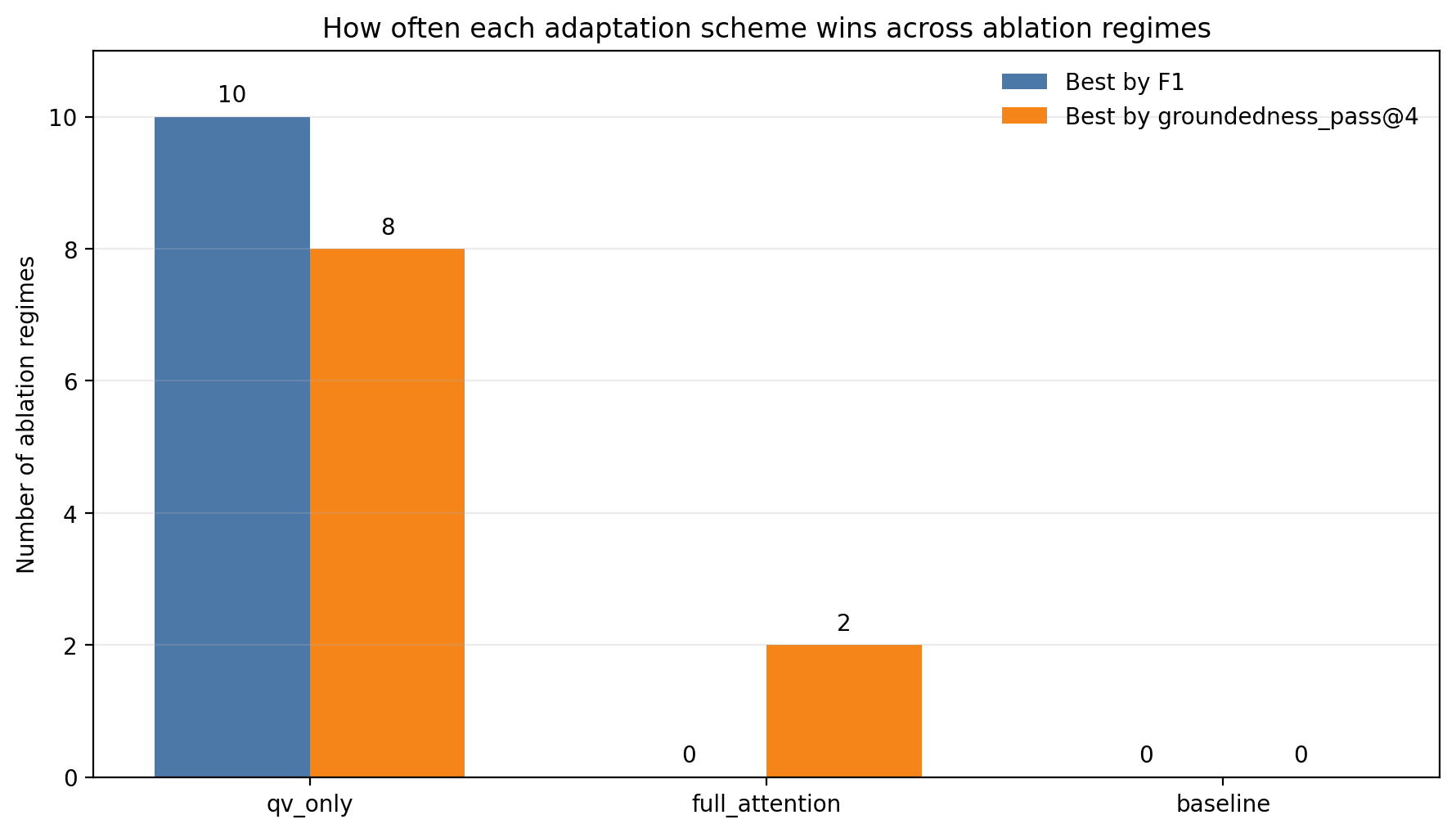}
\caption{Distribution of optimal configurations by adaptation scheme (\texttt{qv\_only}, \texttt{full\_attention}, \texttt{baseline}) across the ten ablation regimes: number of regimes in which the scheme gives the best point on \texttt{F1} and on \texttt{groundedness\_pass@4}.}
\label{fig:app-ablation_scheme_wins}
\end{figure}

The summary results are given in Table~\ref{tab:app-ablation_summary}. The table allows us to see whether the main conclusions about LoRA are preserved when retrieval/prompting is varied, how the optimal working point shifts within a regime, and whether the points optimal in task quality and in groundedness coincide. An aggregated view by adaptation scheme, around which the optimal points concentrate, is additionally shown in Figure~\ref{fig:app-ablation_scheme_wins}. The full per-regime plots and the detailed tables with all configurations are moved to the appendices in order not to overload the main exposition.

\begin{table}[H]
\centering
\footnotesize
\setlength{\tabcolsep}{3pt}
\resizebox{\textwidth}{!}{
\begin{tabular}{ccccccccccccccc}
\toprule
retrieval & prompt & best F1 & F1 [95\% CI] & grnd@4 & corr@4 & lat. & inf. VRAM & best grnd & F1 [95\% CI] & grnd@4 & corr@4 & lat. & inf. VRAM & same \\
 &  & config &  &  &  & (s) & (GB) & config &  &  &  & (s) & (GB) & point \\
\midrule
base & neutral & 8B r64 qv\_only & 0.617 [0.593, 0.639] & 0.818 & 0.827 & 0.655 & 21.926 & 8B r16 qv\_only & 0.615 [0.592, 0.635] & 0.825 & 0.836 & 0.690 & 21.824 & no \\
base & explicit & 8B r64 qv\_only & 0.623 [0.599, 0.645] & 0.822 & 0.824 & 0.664 & 21.953 & 8B r16 qv\_only & 0.610 [0.589, 0.632] & 0.823 & 0.834 & 0.662 & 21.852 & no \\
no\_rerank & neutral & 8B r64 qv\_only & 0.617 [0.596, 0.638] & 0.819 & 0.824 & 0.658 & 21.926 & 8B r16 qv\_only & 0.615 [0.592, 0.637] & 0.836 & 0.842 & 0.685 & 21.824 & no \\
no\_rerank & explicit & 8B r64 qv\_only & 0.623 [0.601, 0.645] & 0.817 & 0.824 & 0.649 & 21.953 & 8B r16 qv\_only & 0.610 [0.589, 0.632] & 0.829 & 0.834 & 0.675 & 21.852 & no \\
dense & neutral & 8B r64 qv\_only & 0.617 [0.595, 0.640] & 0.818 & 0.828 & 0.650 & 21.926 & 8B r8 qv\_only & 0.606 [0.585, 0.628] & 0.824 & 0.831 & 0.673 & 21.836 & no \\
dense & explicit & 8B r64 qv\_only & 0.623 [0.600, 0.645] & 0.824 & 0.827 & 0.661 & 21.953 & 8B r16 qv\_only & 0.610 [0.588, 0.633] & 0.827 & 0.832 & 0.679 & 21.852 & no \\
sparse & neutral & 8B r64 qv\_only & 0.617 [0.595, 0.640] & 0.820 & 0.823 & 0.661 & 21.926 & 8B r4 full\_attention & 0.603 [0.580, 0.626] & 0.827 & 0.832 & 0.801 & 21.871 & no \\
sparse & explicit & 8B r64 qv\_only & 0.623 [0.600, 0.645] & 0.828 & 0.832 & 0.646 & 21.953 & 8B r16 full\_attention & 0.601 [0.580, 0.623] & 0.829 & 0.834 & 0.766 & 21.877 & no \\
bm25\_hybrid & neutral & 8B r64 qv\_only & 0.617 [0.597, 0.639] & 0.815 & 0.824 & 0.659 & 21.926 & 8B r8 qv\_only & 0.606 [0.586, 0.627] & 0.827 & 0.836 & 0.646 & 21.836 & no \\
bm25\_hybrid & explicit & 8B r64 qv\_only & 0.623 [0.603, 0.645] & 0.823 & 0.829 & 0.664 & 21.953 & 8B r16 qv\_only & 0.610 [0.590, 0.631] & 0.834 & 0.838 & 0.691 & 21.852 & no \\
\bottomrule
\end{tabular}
}
\caption{Ablation summary results by \texttt{F1} and \texttt{groundedness\_pass@4}}
\label{tab:app-ablation_summary}
\end{table}

Table~\ref{tab:app-ablation_summary} shows that the main conclusions about LoRA do not disappear when retrieval and prompting are changed. At the level of the entire ablation series the same structural pattern is consistently preserved: the points with the maximum point \texttt{F1} and the points optimal in groundedness in most regimes concentrate around \texttt{qv\_only} adapters, while \texttt{full\_attention} only occasionally turns out to be optimal in groundedness. At the same time, the interval estimates show that within individual regimes some of the best-\texttt{F1} configurations are statistically close to their nearest alternatives. Under moderate changes of the retrieval pipeline and the prompting policy what is therefore reproduced is not so much the advantage of a particular backbone, but the advantage of the \texttt{qv\_only} adaptation scheme itself.

At the same time, the optimal working point within a regime shifts depending on the chosen quality criterion. In all ten regimes the points with the maximum \texttt{F1} do not coincide with the points with the maximum \texttt{groundedness\_pass@4} (\texttt{same\_point = no}). As a result, retrieval and prompting affect not only the absolute level of quality, but also the position of the trade-off between task performance and groundedness. The ablation results therefore confirm the methodological necessity of considering LoRA configurations not as a single source of variation, but as part of a more general retrieval-augmented system with conflicting criteria of quality and cost.

\subsection{Statistical robustness of \texttt{F1} differences}

To separate stable \texttt{F1} differences from narrow gaps, for the key comparisons we computed non-parametric 95\% bootstrap CIs on the test set (\texttt{n = 785}, 1000 resamples). For individual configuration pairs we additionally used the paired bootstrap for the difference $\Delta F1$, since it correctly assesses whether the observed gap crosses the boundary of statistical noise.

\begin{table}[H]
\centering
\footnotesize
\setlength{\tabcolsep}{4pt}
\resizebox{\textwidth}{!}{
\begin{tabular}{lccc}
\toprule
Comparison & $\Delta F1$ & 95\% CI on $\Delta F1$ & Significant? \\
\midrule
\texttt{3B r64 qv\_only} - \texttt{3B baseline} & +0.056 & [+0.033, +0.078] & yes \\
\texttt{3B r64 qv\_only} - \texttt{8B baseline} & +0.003 & [-0.021, +0.026] & no \\
\texttt{8B r64 qv\_only} - \texttt{3B r64 qv\_only} & +0.020 & [+0.0005, +0.0410] & yes \\
\texttt{8B r64 qv\_only} - \texttt{8B r16 qv\_only} & +0.003 & [-0.011, +0.016] & no \\
\texttt{3B r64 qv\_only} - \texttt{3B r64 full\_attention} & +0.020 & [+0.004, +0.036] & yes \\
\texttt{8B r64 qv\_only} - \texttt{8B r64 full\_attention} & +0.011 & [-0.003, +0.025] & no \\
\midrule
\texttt{qv\_only} - \texttt{full\_attention}\textsuperscript{*} & +0.0067 & [+0.0010, +0.0124] & yes \\
\bottomrule
\end{tabular}
}
\vspace{0.3em}
{\footnotesize\noindent\textsuperscript{*} mean $\Delta F1$ over 8 param-matched pairs from Table~\ref{tab:app-param_matched_pairs}.\par}
\caption{Paired bootstrap estimates of key \texttt{F1} differences}
\label{tab:app-paired_f1_deltas}
\end{table}

Table~\ref{tab:app-paired_f1_deltas} shows that the statistically stable effects on the current test set are primarily three: the improvement of \texttt{3B r64 qv\_only} over \texttt{3B baseline}, the superiority of \texttt{3B r64 qv\_only} over \texttt{3B r64 full\_attention}, and the positive average difference of \texttt{qv\_only} over \texttt{full\_attention} across matched rank pairs. On the contrary, the difference between \texttt{3B r64 qv\_only} and \texttt{8B baseline}, as well as between \texttt{8B r64 qv\_only} and \texttt{8B r16 qv\_only}, does not allow us to claim strict statistical superiority. The main stable conclusion of the work is therefore not connected to any local maximum on \texttt{F1}, but to the fact that LoRA adaptation in the successful \texttt{qv\_only} scheme systematically forms stronger working points and gives a confirmed gain in at least some of the key comparisons.

\section{Discussion}
\label{app:discussion}

\subsection{Why \texttt{qv\_only} adapters dominate}

The obtained results show that the strongest response-time fronts are formed not by the wider \texttt{full\_attention} adapters, but by the narrower \texttt{qv\_only} solutions. This is confirmed by the bootstrap analysis: the average difference \texttt{qv\_only - full\_attention} over matched rank pairs is positive and statistically significant (\texttt{$\Delta F1 = +0.0067$}, 95\% CI \texttt{[+0.0010, +0.0124]}), with the effect being most pronounced in the 8B family. The most plausible explanation is that, for documentation-oriented question answering inside a fixed retrieval pipeline, the main task of the adapter is not a deep reorganization of the entire attention block, but a more accurate tuning of how the model selects and uses the context already given to it. In this setting, selective adaptation of the \texttt{q\_proj} and \texttt{v\_proj} projections turns out to be sufficient, while extending LoRA to \texttt{k\_proj} and \texttt{o\_proj} increases the adapter size and training cost faster than it brings a stable benefit on the Pareto fronts.

The comparison of \texttt{3B r64 qv\_only} and \texttt{3B r64 full\_attention} is particularly telling. In the current set of results the broader module coverage not only fails to improve quality, but also loses in \texttt{F1}; for $\Delta F1$ between these configurations the 95\% CI is \texttt{[+0.004, +0.036]}. This means that in the current set of experiments \texttt{full\_attention} acts not as a universally better strategy, but as a more expensive way to obtain a less stable result.

\subsubsection{Param-matched comparison of \texttt{qv\_only} vs \texttt{full\_attention}}
\label{subsubsec:app-param_matched_qv_vs_full}

An additional control check is the param-matched analysis, in which pairs of configurations with the same total number of LoRA parameters are compared. Since \texttt{full\_attention} adapts four attention projections while \texttt{qv\_only} only two, the same parameter budget is reached at half the rank for \texttt{full\_attention}. For every such pair on the final test split (\texttt{n = 785}) we computed $\Delta F1 = F1(\texttt{qv\_only}) - F1(\texttt{full\_attention})$ and the paired bootstrap 95\% CI. The summary is given in Table~\ref{tab:app-param_matched_pairs}.

\begin{table}[H]
\centering
\footnotesize
\setlength{\tabcolsep}{4pt}
\resizebox{\textwidth}{!}{%
\begin{tabular}{lcccccc}
\toprule
Family & Param budget & \texttt{qv\_only} & \texttt{full\_attention} & $\Delta F1$ & 95\% CI & Significant? \\
\midrule
3B & 256d & \texttt{r = 64} & \texttt{r = 32} & +0.0207 & [+0.0052, +0.0373] & yes \\
3B & 128d & \texttt{r = 32} & \texttt{r = 16} & +0.0021 & [-0.0133, +0.0174] & no \\
3B & 64d & \texttt{r = 16} & \texttt{r = 8} & +0.0054 & [-0.0083, +0.0203] & no \\
3B & 32d & \texttt{r = 8} & \texttt{r = 4} & -0.0016 & [-0.0137, +0.0109] & no \\
8B & 256d & \texttt{r = 64} & \texttt{r = 32} & +0.0150 & [-0.0013, +0.0301] & no \\
8B & 128d & \texttt{r = 32} & \texttt{r = 16} & +0.0083 & [-0.0056, +0.0225] & no \\
8B & 64d & \texttt{r = 16} & \texttt{r = 8} & +0.0158 & [+0.0034, +0.0265] & yes \\
8B & 32d & \texttt{r = 8} & \texttt{r = 4} & +0.0032 & [-0.0093, +0.0170] & no \\
\bottomrule
\end{tabular}%
}
\caption{Param-matched comparison of \texttt{qv\_only} and \texttt{full\_attention} by paired bootstrap \texttt{F1} differences}
\label{tab:app-param_matched_pairs}
\end{table}

Table~\ref{tab:app-param_matched_pairs} shows that at the same parameter budget \texttt{qv\_only} is significantly better in 2 of 8 pairs, is statistically comparable in the remaining 6, and in no pair is significantly worse than \texttt{full\_attention}. It follows that the claim of a structural advantage of \texttt{qv\_only} does not reduce to the fact that this scheme simply ``gets more parameters per projection'' at a fixed rank: even in the param-matched setting it is at least no worse and in some comparisons yields a confirmed gain in \texttt{F1}.
\subsection{Why the backbone size sets the operating regime but does not cancel the LoRA effect}

The split between 3B and 8B configurations looks structural, but it should not be interpreted as the main result of the work. The larger 8B model does possess a greater parametric capacity and therefore more often sets the upper bound on \texttt{F1}, but it almost automatically moves the system into a more expensive regime in terms of inference and training memory. The 3B family, on the contrary, forms an area of efficient solutions: with a noticeably smaller resource profile it retains quality high enough to remain a practical choice for constrained hardware. It is particularly telling that the strong adapted configuration \texttt{3B r64 qv\_only} turns out to be statistically comparable to the unadapted \texttt{8B baseline} model: the point difference in \texttt{F1} is only \texttt{+0.003}, while the 95\% CI for $\Delta F1$ is \texttt{[-0.021, +0.026]}. The difference between 3B and 8B therefore cannot be reduced to a simple rule of ``more parameters means better'': domain-specific adaptation can compensate for the advantage of a larger LoRA-free backbone even where a strict conclusion about superiority in \texttt{F1} can no longer be drawn.

Thus the backbone scale primarily sets the operating regime of the system, but does not answer the question of which adaptation scheme is successful. This question is decided inside each family separately, and this is where the main effect of the work appears: in both the 3B and the 8B family the \texttt{qv\_only} adapters turn out to be more competitive in the quality-cost ratio than \texttt{full\_attention}. The difference between 3B and 8B should therefore be understood as a background against which a more important regularity, related to the structure of LoRA adaptation itself, becomes visible.

\subsection{Why the effect of rank growth is non-linear}

Increasing the rank does not produce the same effect for all configurations, however in both families an important regularity is observed: increasing \texttt{r} improves quality more strongly than it raises the runtime cost. For 3B \texttt{qv\_only} adapters the move from \texttt{r = 4} to \texttt{r = 64} raises \texttt{F1} from \texttt{0.572} to \texttt{0.597}, while the latency of these two extreme points changes only from \texttt{0.633} to \texttt{0.598}~s, and the inference memory from \texttt{12.668} to \texttt{12.760}~GB. This means that in the current pipeline the main inference cost is determined not so much by the adapter size as by the base model and the retrieval pipeline, and the additional adaptation capacity coming with rank goes mainly into quality improvements.

The same logic is partially reproduced in the larger family as well: inside \texttt{qv\_only} the move from \texttt{8B r4} to \texttt{8B r64} raises \texttt{F1} from \texttt{0.610} to \texttt{0.617}, while latency stays in the approximate range of \texttt{0.66-0.69}~s and inference memory changes very little inside the 8B regime. However, between \texttt{8B r16 qv\_only} and \texttt{8B r64 qv\_only} the paired bootstrap gives a 95\% CI of \texttt{[-0.011, +0.016]}, so inside the upper part of the 8B family rank growth is more accurately interpreted as a weak positive trend rather than a strictly established advantage of the higher rank. In other words, a high rank brings the greatest benefit precisely where a structurally successful LoRA adaptation scheme has already been chosen, but inside the 8B family the choice between the top ranks requires accounting for additional criteria.

\subsection{Why retrieval and prompting shift the optimum point but do not cancel the conclusions about LoRA}

The ablation experiments show that the retrieval and prompting components affect not only the absolute values of the metrics, but also the position of the optimal working point inside the same set of adapters. Switching off the reranker, moving to dense-only or sparse-only retrieval, and explicit changes to the prompting policy can shift the local optimum in \texttt{F1} or in groundedness, since they change both the quality of the supplied context and the freedom of the generator in producing the answer.

However, these changes do not destroy the main conclusion about LoRA configurations; they only modify the conditions in which it shows itself. In other words, retrieval and prompting determine how favourable the environment for the generator turns out to be, but inside this environment the advantage more often stays with \texttt{qv\_only} adapters. The ablation results should therefore be read not as a refutation of the main LoRA conclusions, but as a confirmation that the adapter, retrieval, and prompting form a single pipeline, inside which a structurally successful adaptation scheme remains a stable source of advantage.

\subsection{Generalization analysis by \texttt{top\_k}}

As a separate generalization analysis we checked the robustness of the main conclusions with respect to the number of chunks passed to the generator at inference time. In addition to the main regime \texttt{top\_k = 2}, additional experiments were performed for \texttt{top\_k = 1} and \texttt{top\_k = 4} in the \texttt{01\_base\_\_neutral} regime for all 22 configurations without any further adapter training. This cross-section makes it possible to check that the obtained conclusions do not reduce to one particular value of the retrieval-context budget.

\begin{figure}[H]
\centering
\includegraphics[width=0.9\textwidth]{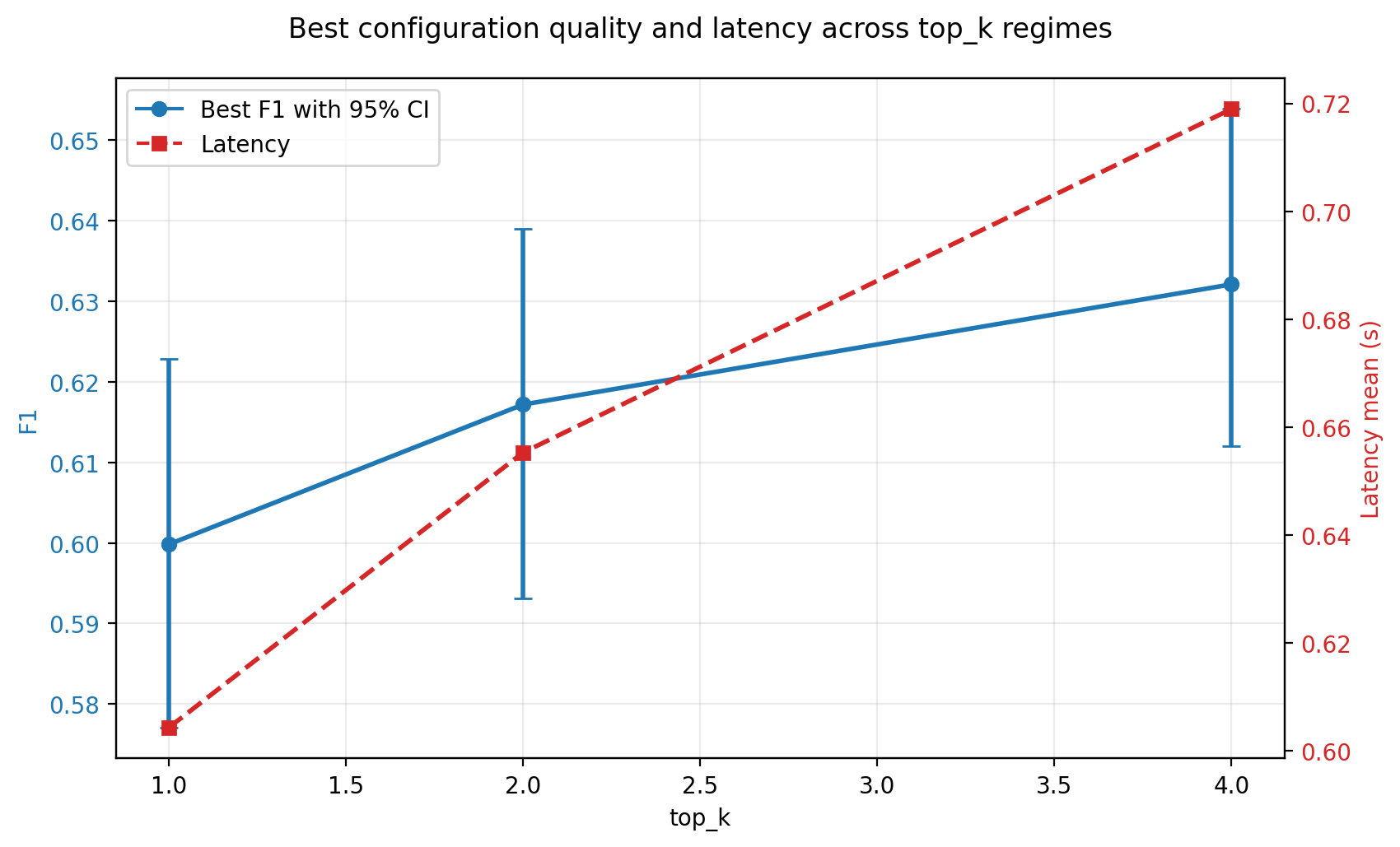}
\caption{Change of the best quality and latency for different values of \texttt{top\_k}}
\label{fig:app-topk_tradeoff_summary}
\end{figure}

The plot for \texttt{top\_k = 1, 2, 4} shows that increasing the retrieval-context budget consistently raises the maximum of \texttt{F1} (\texttt{0.600 -> 0.617 -> 0.632}), but at the same time raises latency (\texttt{0.604 -> 0.655 -> 0.719}~s). This means that increasing \texttt{top\_k} does not reduce to a free improvement in quality: a wider context does help the best configuration, but it requires a more expensive inference-time regime.

A comparison of the concrete best points is given in Table~\ref{tab:app-topk_summary}.

\begin{table}[H]
\centering
\footnotesize
\setlength{\tabcolsep}{4pt}
\resizebox{\textwidth}{!}{
\begin{tabular}{cccccc}
\toprule
\texttt{top\_k} & best F1 config & F1 [95\% CI] & lat. (s) & inf. VRAM (GB) & runtime Pareto front \\
\midrule
\texttt{1} & \texttt{8B r64 qv\_only} & 0.600 [0.577, 0.623] & 0.604 & 21.820 & \texttt{8B r64 qv\_only} \\
\texttt{2} & \texttt{8B r64 qv\_only} & 0.617 [0.593, 0.639] & 0.655 & 21.926 & \texttt{3B r64 qv\_only}, \texttt{8B r64 qv\_only} \\
\texttt{4} & \texttt{8B r64 qv\_only} & 0.632 [0.612, 0.654] & 0.719 & 22.078 & \texttt{3B r64 qv\_only}, \texttt{8B r64 qv\_only} \\
\bottomrule
\end{tabular}
}
\caption{Comparison of the best configurations for different values of \texttt{top\_k}}
\label{tab:app-topk_summary}
\end{table}

Table~\ref{tab:app-topk_summary} shows that for all three values of \texttt{top\_k} the best configuration in \texttt{F1} remains the same: \texttt{8B r64 qv\_only}. At the same time, the structure of the runtime trade-off changes only mildly: for \texttt{top\_k = 1} the quality-latency front collapses to the single point \texttt{8B r64 qv\_only}, while for \texttt{top\_k = 2} and \texttt{top\_k = 4} it simultaneously contains the lightweight point \texttt{3B r64 qv\_only} and the highest-quality point \texttt{8B r64 qv\_only}. The generalization analysis therefore shows that varying \texttt{top\_k} affects primarily the absolute values of quality and cost, but does not cancel the main conclusion about the advantage of LoRA adapters that adapt the \texttt{q}- and \texttt{v}-projections of attention.

\subsection{Practical recommendations}

From the obtained results, several explicit recommendations on the choice of configuration can be drawn.

\begin{itemize}
\item If the priority is a lightweight inference regime with high quality preserved, the preferred configuration is \texttt{3B r64 qv\_only}.
\item If minimizing training cost while keeping acceptable quality is important, a reasonable starting point is \texttt{3B r4 qv\_only}, since it is the cheapest non-dominated configuration in training time and memory.
\item If one needs the best quality-cost compromise inside the light family, the most practical intermediate configuration is \texttt{3B r32 qv\_only}: it already lies on the training Pareto front and is noticeably close to the upper points in quality without moving to the 8B regime.
\item If it is important to obtain quality statistically comparable to \texttt{8B baseline} but with a lighter resource profile, a reasonable working point is \texttt{3B r64 qv\_only}.
\item If inside the 8B family it is more important to keep higher groundedness at statistically comparable quality, \texttt{8B r16 qv\_only} is preferable, since exactly this configuration yields the maximum \texttt{groundedness\_pass@4} in the base regime.
\item If the priority is the maximum point-quality estimate and a substantially larger resource budget is available, \texttt{8B r64 qv\_only} is preferable; however its advantage in \texttt{F1} over \texttt{8B r16 qv\_only} on the current test set is not statistically confirmed.
\item \texttt{full\_attention} configurations are reasonable to consider as a niche option: in the current set of runs they rarely give a stable advantage in \texttt{F1} and more often lose in cost.
\end{itemize}

\subsection{Error analysis}

Alongside the aggregate metrics it is useful to consider the typical error classes that recur across different RAG-system configurations. In the present work it is convenient to group these errors not by individual failing examples, but by the failure mechanism inside the retrieval-augmented pipeline. This allows us to separate errors caused by low-quality context from errors that arise already at the answer-generation stage given relevant context.

To make this section quantitative, we built a reproducible sample of 100 failing answers on the final test split: 50 randomly chosen cases with \texttt{F1 < 1} for \texttt{3B r64 qv\_only} and \texttt{8B r64 qv\_only} each (\texttt{seed = 42}). Each example was then assigned to one of four operational error classes: \texttt{retrieval miss} (the relevant source chunk did not enter the final context), \texttt{overclaiming} (the answer adds extra unsupported statements), \texttt{incomplete answer} (the answer extracts only part of the required information), and \texttt{exact/precision failure} (an error of exact reproduction of a command, field, version, name, or other literal value). The summary is given in Table~\ref{tab:app-error-types-sample}.

\begin{table}[H]
\centering
\footnotesize
\setlength{\tabcolsep}{4pt}
\caption{Quantitative cross-section of error types on a sample of 100 failing answers}
\label{tab:app-error-types-sample}
\begin{tabular}{lccc}
\toprule
Error type & 3B r64 qv\_only & 8B r64 qv\_only & Total \\
\midrule
\texttt{retrieval miss} & 11 (22\%) & 8 (16\%) & 19 (19\%) \\
\texttt{overclaiming} & 2 (4\%) & 2 (4\%) & 4 (4\%) \\
\texttt{incomplete answer} & 17 (34\%) & 7 (14\%) & 24 (24\%) \\
\texttt{exact/precision failure} & 20 (40\%) & 33 (66\%) & 53 (53\%) \\
\bottomrule
\end{tabular}
\end{table}

Table~\ref{tab:app-error-types-sample} shows that in this sample exact-reproduction errors dominate (53\%), and this is especially pronounced for \texttt{8B r64 qv\_only} (66\% against 40\% for \texttt{3B r64 qv\_only}). This indicates that a noticeable share of the remaining errors is connected not with the full absence of relevant context, but with inaccurate reproduction of short commands, field names, versions, and other local facts. The second place is occupied by incomplete answers (24\%), followed by \texttt{retrieval miss} (19\%), while \texttt{overclaiming} is noticeably rarer in this sample (4\%).

Below are some illustrative examples from the same sample.

\begin{itemize}
\item \texttt{retrieval miss}: for the question \texttt{What does kube-proxy provide for Services?} the reference answer is formulated as \texttt{provides load balancing}, while the model answers \texttt{A virtual IP mechanism.}; for the question \texttt{What protocol does kube-proxy not understand?} the reference is \texttt{does not understand HTTP}, while the model returns \texttt{TCP, UDP, AND SCTP STREAM FORWARDING.}.
\item \texttt{overclaiming}: for the question \texttt{What does the term storage version describe in Kubernetes?} the reference is limited to the formula \texttt{how an object is stored in your cluster}, but the model adds an extra part about the object's representation in the API; for the question \texttt{Why might a Windows Pod show ErrImgPull or ImagePullBackOff?} the reference ties the error to running on an incompatible Windows node, while the model substitutes more general reasons such as a wrong image name or a missing \texttt{imagePullSecret}.
\item \texttt{incomplete answer}: for the question \texttt{Which method is easier to use for adding custom resources: CRDs or aggregated APIs?} the reference answer is two-part: \texttt{CRDs are easier to use. Aggregated APIs are more flexible.}, while the model keeps only the first part; for the question \texttt{Does Kubernetes have a User object for storing usernames in its API?} the reference simultaneously states the absence of a \texttt{User} object and the absence of storing usernames in the API, while the model answers only \texttt{No.}.
\item \texttt{exact/precision failure}: for the question \texttt{What is the default API server bind port for kubeadm init phase kubeconfig all?} the reference value is \texttt{6443}, while the model returns \texttt{6444}; for the question \texttt{Which flag forces Windows line endings in kubectl edit?} the reference form is \texttt{--windows-line-endings}, while the model returns the string \texttt{windows-line-enddings}.
\end{itemize}
First, a noticeable but non-dominant share is made up of \texttt{retrieval miss}-type errors, in which the relevant documentation fragment does not enter the final context. In this case the generator is forced to answer from incomplete or indirectly related material, and therefore even strong adapters cannot stably recover the correct answer. Practically such cases show up either as a factual error or as a partially correct but overly general answer.

Second, we observe unsupported-addition (overclaiming) errors. In the current sample there are few of them, but they remain methodologically important: the answer looks plausible and often gets an acceptable \texttt{F1}, but it adds statements that do not directly follow from the retrieved context. This error type is especially sensitive to the judge-based groundedness evaluation and explains why configurations optimal in \texttt{F1} do not coincide with configurations optimal in \texttt{groundedness\_pass@4}.

Third, we see incomplete answers in the presence of relevant context. On the quantitative sample this is the second most frequent error type. Here retrieval has already worked well enough, but the generator extracts from the found material only part of the required information: it omits a caveat, a constraint, a condition, or one element of a multi-part answer. Such errors are especially important for technical documentation, where even a brief omission may change the practical meaning of the answer.

Fourth, for the class of exact questions, format and precision-of-reproduction errors are typical, and exactly this type turns out to be the largest in the quantitative sample. Even with the correct context, the model can give an answer that is close in meaning but not literally correct: it rephrases the field name, inaccurately reproduces a flag, a path, an API version, or a specific value. Such failures show that some of the errors in documentation-oriented question answering are related not to the absence of knowledge as such, but to the difficulty of accurately reproducing local facts.

Finally, a separate class is made up of prompt-sensitive failures. The ablation experiments show that changing the retrieval/prompting policy can shift the optimal working point even for the same LoRA configurations. This effect was not encoded as a separate row in Table~\ref{tab:app-error-types-sample}, since it shows up primarily at the level of between-regime comparison rather than inside a single fixed sample of answers. It follows that a proper error analysis for a RAG system should treat the generator, retrieval, and prompting as interrelated components of one pipeline, not as independent sources of quality.

\section{Per-Regime Detailed Tables}
\label{app:tables}

This appendix contains the detailed metric tables for each of the 10 ablation
regimes referenced in Section~\ref{sec:results}. For every regime we report,
for each of the 12 LoRA configurations and the two non-adapted baselines
(3B/8B), the point estimates of $F_1$, judge \texttt{groundedness\_pass@4},
judge \texttt{correctness\_pass@4} together with their 95\% bootstrap
confidence intervals ($1{,}000$ resamples on the test split, $n = 785$), as
well as the corresponding latency and inference VRAM. The same numbers
underpin all per-regime plots in Appendix~\ref{app:plots} and the aggregate
ablation summary in Section~\ref{sec:results}.

\subsection*{\texttt{01\_base\_\_neutral}}

\begin{table}[H]
\centering
\footnotesize
\setlength{\tabcolsep}{3pt}
\resizebox{\textwidth}{!}{
\begin{tabular}{lccccc}
\toprule
config & F1 [95\% CI] & grnd@4 [95\% CI] & corr@4 [95\% CI] & lat. (s) & inf. VRAM (GB) \\
\midrule
\texttt{3B baseline} & 0.542 [0.517, 0.565] & 0.716 [0.684, 0.745] & 0.717 [0.684, 0.748] & 0.628 & 12.664 \\
\texttt{3B r4 full\_attention} & 0.577 [0.554, 0.600] & 0.795 [0.767, 0.823] & 0.808 [0.781, 0.834] & 0.975 & 12.674 \\
\texttt{3B r4 qv\_only} & 0.572 [0.550, 0.592] & 0.800 [0.772, 0.827] & 0.806 [0.777, 0.833] & 0.633 & 12.668 \\
\texttt{3B r8 full\_attention} & 0.578 [0.558, 0.599] & 0.789 [0.761, 0.819] & 0.808 [0.782, 0.834] & 0.734 & 12.701 \\
\texttt{3B r8 qv\_only} & 0.575 [0.553, 0.597] & 0.803 [0.772, 0.831] & 0.813 [0.787, 0.842] & 0.731 & 12.676 \\
\texttt{3B r16 full\_attention} & 0.590 [0.568, 0.614] & 0.801 [0.774, 0.829] & 0.810 [0.782, 0.838] & 0.753 & 12.719 \\
\texttt{3B r16 qv\_only} & 0.583 [0.562, 0.605] & 0.783 [0.755, 0.811] & 0.790 [0.758, 0.818] & 0.649 & 12.701 \\
\texttt{3B r32 full\_attention} & 0.577 [0.554, 0.600] & 0.789 [0.759, 0.817] & 0.803 [0.775, 0.831] & 0.776 & 12.732 \\
\texttt{3B r32 qv\_only} & 0.592 [0.569, 0.614] & 0.796 [0.764, 0.824] & 0.808 [0.781, 0.834] & 0.621 & 12.719 \\
\texttt{3B r64 full\_attention} & 0.578 [0.554, 0.601] & 0.799 [0.772, 0.827] & 0.818 [0.792, 0.845] & 0.783 & 12.924 \\
\texttt{3B r64 qv\_only} & 0.597 [0.574, 0.622] & 0.782 [0.752, 0.811] & 0.791 [0.762, 0.817] & 0.598 & 12.760 \\
\texttt{8B baseline} & 0.595 [0.572, 0.620] & 0.772 [0.739, 0.800] & 0.775 [0.746, 0.804] & 0.755 & 21.836 \\
\texttt{8B r4 full\_attention} & 0.603 [0.583, 0.625] & 0.819 [0.794, 0.846] & 0.828 [0.801, 0.855] & 0.792 & 21.871 \\
\texttt{8B r4 qv\_only} & 0.610 [0.589, 0.631] & 0.822 [0.795, 0.847] & 0.832 [0.806, 0.859] & 0.686 & 21.828 \\
\texttt{8B r8 full\_attention} & 0.599 [0.578, 0.620] & 0.809 [0.782, 0.837] & 0.815 [0.789, 0.841] & 0.772 & 21.859 \\
\texttt{8B r8 qv\_only} & 0.606 [0.584, 0.628] & 0.822 [0.795, 0.847] & 0.833 [0.806, 0.860] & 0.706 & 21.836 \\
\texttt{8B r16 full\_attention} & 0.604 [0.583, 0.625] & 0.805 [0.777, 0.832] & 0.810 [0.783, 0.836] & 0.807 & 21.885 \\
\texttt{8B r16 qv\_only} & 0.615 [0.592, 0.635] & 0.825 [0.797, 0.854] & 0.836 [0.811, 0.862] & 0.690 & 21.824 \\
\texttt{8B r32 full\_attention} & 0.602 [0.581, 0.624] & 0.800 [0.772, 0.827] & 0.808 [0.780, 0.838] & 0.844 & 21.961 \\
\texttt{8B r32 qv\_only} & 0.612 [0.590, 0.634] & 0.811 [0.783, 0.837] & 0.814 [0.786, 0.841] & 0.665 & 21.850 \\
\texttt{8B r64 full\_attention} & 0.606 [0.583, 0.628] & 0.814 [0.789, 0.839] & 0.815 [0.789, 0.842] & 0.899 & 22.164 \\
\texttt{8B r64 qv\_only} & 0.617 [0.593, 0.639] & 0.818 [0.791, 0.842] & 0.827 [0.797, 0.855] & 0.655 & 21.926 \\
\bottomrule
\end{tabular}
}
\caption{Per-configuration metrics for the \texttt{01\_base\_\_neutral} regime: F1, groundedness, and correctness (pass@4) with 95\% bootstrap CI (1000 resamples on the final test split), inference latency, and peak inference VRAM.}
\label{tab:appA-01-base--neutral}
\end{table}

\clearpage

\subsection*{\texttt{02\_base\_\_explicit\_grounded}}

\begin{table}[H]
\centering
\footnotesize
\setlength{\tabcolsep}{3pt}
\resizebox{\textwidth}{!}{
\begin{tabular}{lccccc}
\toprule
config & F1 [95\% CI] & grnd@4 [95\% CI] & corr@4 [95\% CI] & lat. (s) & inf. VRAM (GB) \\
\midrule
\texttt{3B baseline} & 0.536 [0.512, 0.562] & 0.729 [0.698, 0.761] & 0.724 [0.693, 0.754] & 0.610 & 12.686 \\
\texttt{3B r4 full\_attention} & 0.559 [0.536, 0.583] & 0.778 [0.749, 0.806] & 0.796 [0.768, 0.824] & 0.736 & 12.695 \\
\texttt{3B r4 qv\_only} & 0.566 [0.545, 0.587] & 0.806 [0.777, 0.836] & 0.813 [0.783, 0.839] & 0.653 & 12.689 \\
\texttt{3B r8 full\_attention} & 0.569 [0.546, 0.591] & 0.787 [0.760, 0.817] & 0.803 [0.776, 0.829] & 0.768 & 12.723 \\
\texttt{3B r8 qv\_only} & 0.571 [0.549, 0.593] & 0.778 [0.750, 0.806] & 0.792 [0.762, 0.820] & 0.668 & 12.697 \\
\texttt{3B r16 full\_attention} & 0.577 [0.556, 0.598] & 0.789 [0.759, 0.815] & 0.805 [0.778, 0.832] & 0.777 & 12.740 \\
\texttt{3B r16 qv\_only} & 0.574 [0.551, 0.597] & 0.795 [0.764, 0.822] & 0.805 [0.776, 0.832] & 0.651 & 12.723 \\
\texttt{3B r32 full\_attention} & 0.567 [0.544, 0.591] & 0.786 [0.759, 0.814] & 0.803 [0.773, 0.831] & 0.814 & 12.754 \\
\texttt{3B r32 qv\_only} & 0.576 [0.551, 0.600] & 0.775 [0.744, 0.801] & 0.783 [0.755, 0.811] & 0.669 & 12.740 \\
\texttt{3B r64 full\_attention} & 0.573 [0.549, 0.597] & 0.787 [0.759, 0.815] & 0.808 [0.781, 0.834] & 0.897 & 12.945 \\
\texttt{3B r64 qv\_only} & 0.598 [0.575, 0.619] & 0.776 [0.749, 0.805] & 0.790 [0.759, 0.818] & 0.596 & 12.781 \\
\texttt{8B baseline} & 0.603 [0.578, 0.627] & 0.780 [0.753, 0.806] & 0.778 [0.750, 0.805] & 0.752 & 21.795 \\
\texttt{8B r4 full\_attention} & 0.602 [0.580, 0.623] & 0.819 [0.792, 0.846] & 0.825 [0.799, 0.852] & 0.814 & 21.863 \\
\texttt{8B r4 qv\_only} & 0.602 [0.580, 0.623] & 0.811 [0.783, 0.838] & 0.824 [0.799, 0.851] & 0.687 & 21.855 \\
\texttt{8B r8 full\_attention} & 0.601 [0.579, 0.623] & 0.813 [0.786, 0.837] & 0.819 [0.794, 0.846] & 0.789 & 21.852 \\
\texttt{8B r8 qv\_only} & 0.606 [0.585, 0.627] & 0.820 [0.794, 0.846] & 0.836 [0.810, 0.860] & 0.693 & 21.863 \\
\texttt{8B r16 full\_attention} & 0.601 [0.580, 0.622] & 0.813 [0.783, 0.838] & 0.814 [0.786, 0.842] & 0.843 & 21.877 \\
\texttt{8B r16 qv\_only} & 0.610 [0.589, 0.632] & 0.823 [0.797, 0.848] & 0.834 [0.806, 0.859] & 0.662 & 21.852 \\
\texttt{8B r32 full\_attention} & 0.601 [0.579, 0.622] & 0.804 [0.776, 0.831] & 0.811 [0.781, 0.838] & 0.808 & 21.953 \\
\texttt{8B r32 qv\_only} & 0.610 [0.588, 0.632] & 0.813 [0.786, 0.838] & 0.815 [0.789, 0.841] & 0.648 & 21.877 \\
\texttt{8B r64 full\_attention} & 0.603 [0.580, 0.623] & 0.808 [0.777, 0.833] & 0.815 [0.786, 0.845] & 0.806 & 22.156 \\
\texttt{8B r64 qv\_only} & 0.623 [0.599, 0.645] & 0.822 [0.795, 0.848] & 0.824 [0.796, 0.850] & 0.664 & 21.953 \\
\bottomrule
\end{tabular}
}
\caption{Per-configuration metrics for the \texttt{02\_base\_\_explicit\_grounded} regime: F1, groundedness, and correctness (pass@4) with 95\% bootstrap CI (1000 resamples on the final test split), inference latency, and peak inference VRAM.}
\label{tab:appA-02-base--explicit-grounded}
\end{table}

\clearpage

\subsection*{\texttt{03\_reranker\_off\_\_neutral}}

\begin{table}[H]
\centering
\footnotesize
\setlength{\tabcolsep}{3pt}
\resizebox{\textwidth}{!}{
\begin{tabular}{lccccc}
\toprule
config & F1 [95\% CI] & grnd@4 [95\% CI] & corr@4 [95\% CI] & lat. (s) & inf. VRAM (GB) \\
\midrule
\texttt{3B baseline} & 0.541 [0.516, 0.564] & 0.729 [0.699, 0.761] & 0.718 [0.685, 0.746] & 0.634 & 12.664 \\
\texttt{3B r4 full\_attention} & 0.573 [0.552, 0.595] & 0.780 [0.753, 0.809] & 0.794 [0.767, 0.820] & 0.732 & 12.674 \\
\texttt{3B r4 qv\_only} & 0.571 [0.548, 0.593] & 0.800 [0.772, 0.829] & 0.811 [0.783, 0.838] & 0.636 & 12.668 \\
\texttt{3B r8 full\_attention} & 0.579 [0.557, 0.600] & 0.799 [0.772, 0.824] & 0.808 [0.780, 0.836] & 0.784 & 12.701 \\
\texttt{3B r8 qv\_only} & 0.574 [0.554, 0.597] & 0.799 [0.769, 0.827] & 0.808 [0.778, 0.836] & 0.647 & 12.676 \\
\texttt{3B r16 full\_attention} & 0.586 [0.564, 0.608] & 0.803 [0.775, 0.831] & 0.815 [0.786, 0.838] & 0.798 & 12.719 \\
\texttt{3B r16 qv\_only} & 0.581 [0.558, 0.605] & 0.791 [0.762, 0.817] & 0.799 [0.772, 0.827] & 0.649 & 12.701 \\
\texttt{3B r32 full\_attention} & 0.579 [0.557, 0.601] & 0.794 [0.766, 0.823] & 0.814 [0.785, 0.839] & 0.781 & 12.732 \\
\texttt{3B r32 qv\_only} & 0.589 [0.568, 0.611] & 0.794 [0.766, 0.823] & 0.804 [0.778, 0.829] & 0.626 & 12.719 \\
\texttt{3B r64 full\_attention} & 0.575 [0.549, 0.596] & 0.790 [0.762, 0.819] & 0.805 [0.776, 0.832] & 0.777 & 12.924 \\
\texttt{3B r64 qv\_only} & 0.597 [0.572, 0.620] & 0.786 [0.758, 0.815] & 0.794 [0.766, 0.823] & 0.613 & 12.760 \\
\texttt{8B baseline} & 0.592 [0.569, 0.617] & 0.771 [0.741, 0.801] & 0.766 [0.735, 0.796] & 0.779 & 21.836 \\
\texttt{8B r4 full\_attention} & 0.603 [0.583, 0.625] & 0.823 [0.796, 0.848] & 0.831 [0.804, 0.856] & 0.788 & 21.871 \\
\texttt{8B r4 qv\_only} & 0.610 [0.587, 0.631] & 0.822 [0.794, 0.848] & 0.829 [0.804, 0.855] & 0.656 & 21.828 \\
\texttt{8B r8 full\_attention} & 0.599 [0.577, 0.619] & 0.808 [0.780, 0.834] & 0.815 [0.785, 0.841] & 0.742 & 21.859 \\
\texttt{8B r8 qv\_only} & 0.606 [0.584, 0.627] & 0.828 [0.800, 0.854] & 0.832 [0.805, 0.857] & 0.656 & 21.836 \\
\texttt{8B r16 full\_attention} & 0.604 [0.584, 0.624] & 0.809 [0.782, 0.836] & 0.811 [0.783, 0.838] & 0.809 & 21.885 \\
\texttt{8B r16 qv\_only} & 0.615 [0.592, 0.637] & 0.836 [0.809, 0.861] & 0.842 [0.815, 0.868] & 0.685 & 21.824 \\
\texttt{8B r32 full\_attention} & 0.602 [0.580, 0.622] & 0.797 [0.767, 0.824] & 0.806 [0.778, 0.833] & 0.817 & 21.961 \\
\texttt{8B r32 qv\_only} & 0.612 [0.590, 0.636] & 0.817 [0.791, 0.843] & 0.820 [0.794, 0.848] & 0.641 & 21.850 \\
\texttt{8B r64 full\_attention} & 0.606 [0.585, 0.627] & 0.815 [0.787, 0.842] & 0.817 [0.786, 0.841] & 0.843 & 22.164 \\
\texttt{8B r64 qv\_only} & 0.617 [0.596, 0.638] & 0.819 [0.792, 0.846] & 0.824 [0.796, 0.851] & 0.658 & 21.926 \\
\bottomrule
\end{tabular}
}
\caption{Per-configuration metrics for the \texttt{03\_reranker\_off\_\_neutral} regime: F1, groundedness, and correctness (pass@4) with 95\% bootstrap CI (1000 resamples on the final test split), inference latency, and peak inference VRAM.}
\label{tab:appA-03-reranker-off--neutral}
\end{table}

\clearpage

\subsection*{\texttt{04\_reranker\_off\_\_explicit\_grounded}}

\begin{table}[H]
\centering
\footnotesize
\setlength{\tabcolsep}{3pt}
\resizebox{\textwidth}{!}{
\begin{tabular}{lccccc}
\toprule
config & F1 [95\% CI] & grnd@4 [95\% CI] & corr@4 [95\% CI] & lat. (s) & inf. VRAM (GB) \\
\midrule
\texttt{3B baseline} & 0.537 [0.514, 0.561] & 0.740 [0.710, 0.769] & 0.740 [0.707, 0.772] & 0.665 & 12.686 \\
\texttt{3B r4 full\_attention} & 0.559 [0.537, 0.583] & 0.780 [0.749, 0.810] & 0.786 [0.761, 0.813] & 0.783 & 12.695 \\
\texttt{3B r4 qv\_only} & 0.566 [0.544, 0.587] & 0.800 [0.772, 0.827] & 0.809 [0.781, 0.836] & 0.624 & 12.689 \\
\texttt{3B r8 full\_attention} & 0.569 [0.545, 0.592] & 0.800 [0.775, 0.829] & 0.810 [0.783, 0.837] & 0.760 & 12.723 \\
\texttt{3B r8 qv\_only} & 0.571 [0.547, 0.593] & 0.789 [0.761, 0.817] & 0.803 [0.773, 0.831] & 0.670 & 12.697 \\
\texttt{3B r16 full\_attention} & 0.577 [0.554, 0.598] & 0.794 [0.762, 0.822] & 0.805 [0.778, 0.832] & 0.788 & 12.740 \\
\texttt{3B r16 qv\_only} & 0.574 [0.551, 0.597] & 0.790 [0.763, 0.818] & 0.804 [0.776, 0.832] & 0.631 & 12.723 \\
\texttt{3B r32 full\_attention} & 0.567 [0.544, 0.589] & 0.794 [0.767, 0.822] & 0.804 [0.776, 0.831] & 0.782 & 12.754 \\
\texttt{3B r32 qv\_only} & 0.576 [0.552, 0.597] & 0.785 [0.757, 0.814] & 0.797 [0.769, 0.824] & 0.633 & 12.740 \\
\texttt{3B r64 full\_attention} & 0.573 [0.550, 0.595] & 0.799 [0.772, 0.827] & 0.808 [0.780, 0.834] & 0.881 & 12.945 \\
\texttt{3B r64 qv\_only} & 0.598 [0.575, 0.621] & 0.777 [0.745, 0.805] & 0.785 [0.757, 0.815] & 0.602 & 12.781 \\
\texttt{8B baseline} & 0.599 [0.575, 0.620] & 0.773 [0.744, 0.803] & 0.775 [0.743, 0.805] & 0.780 & 21.795 \\
\texttt{8B r4 full\_attention} & 0.602 [0.581, 0.624] & 0.817 [0.789, 0.842] & 0.827 [0.800, 0.854] & 0.785 & 21.863 \\
\texttt{8B r4 qv\_only} & 0.602 [0.579, 0.624] & 0.815 [0.787, 0.845] & 0.824 [0.796, 0.851] & 0.714 & 21.855 \\
\texttt{8B r8 full\_attention} & 0.601 [0.579, 0.624] & 0.819 [0.791, 0.845] & 0.823 [0.794, 0.851] & 0.786 & 21.852 \\
\texttt{8B r8 qv\_only} & 0.606 [0.584, 0.626] & 0.819 [0.790, 0.846] & 0.827 [0.800, 0.854] & 0.683 & 21.863 \\
\texttt{8B r16 full\_attention} & 0.601 [0.580, 0.624] & 0.823 [0.796, 0.851] & 0.828 [0.804, 0.855] & 0.816 & 21.877 \\
\texttt{8B r16 qv\_only} & 0.610 [0.589, 0.632] & 0.829 [0.801, 0.856] & 0.834 [0.806, 0.860] & 0.675 & 21.852 \\
\texttt{8B r32 full\_attention} & 0.601 [0.581, 0.622] & 0.804 [0.776, 0.831] & 0.810 [0.783, 0.836] & 0.832 & 21.953 \\
\texttt{8B r32 qv\_only} & 0.610 [0.589, 0.631] & 0.810 [0.783, 0.837] & 0.820 [0.792, 0.847] & 0.647 & 21.877 \\
\texttt{8B r64 full\_attention} & 0.603 [0.580, 0.625] & 0.808 [0.777, 0.834] & 0.811 [0.783, 0.837] & 0.829 & 22.156 \\
\texttt{8B r64 qv\_only} & 0.623 [0.601, 0.645] & 0.817 [0.791, 0.843] & 0.824 [0.795, 0.851] & 0.649 & 21.953 \\
\bottomrule
\end{tabular}
}
\caption{Per-configuration metrics for the \texttt{04\_reranker\_off\_\_explicit\_grounded} regime: F1, groundedness, and correctness (pass@4) with 95\% bootstrap CI (1000 resamples on the final test split), inference latency, and peak inference VRAM.}
\label{tab:appA-04-reranker-off--explicit-grounded}
\end{table}

\clearpage

\subsection*{\texttt{05\_dense\_only\_\_neutral}}

\begin{table}[H]
\centering
\footnotesize
\setlength{\tabcolsep}{3pt}
\resizebox{\textwidth}{!}{
\begin{tabular}{lccccc}
\toprule
config & F1 [95\% CI] & grnd@4 [95\% CI] & corr@4 [95\% CI] & lat. (s) & inf. VRAM (GB) \\
\midrule
\texttt{3B baseline} & 0.536 [0.512, 0.559] & 0.724 [0.693, 0.757] & 0.724 [0.694, 0.754] & 0.598 & 11.676 \\
\texttt{3B r4 full\_attention} & 0.573 [0.552, 0.594] & 0.786 [0.757, 0.814] & 0.806 [0.781, 0.833] & 0.725 & 12.674 \\
\texttt{3B r4 qv\_only} & 0.571 [0.546, 0.593] & 0.810 [0.783, 0.838] & 0.814 [0.787, 0.841] & 0.616 & 12.668 \\
\texttt{3B r8 full\_attention} & 0.579 [0.558, 0.601] & 0.797 [0.767, 0.824] & 0.811 [0.785, 0.837] & 0.762 & 12.701 \\
\texttt{3B r8 qv\_only} & 0.574 [0.551, 0.597] & 0.796 [0.769, 0.823] & 0.805 [0.777, 0.833] & 0.660 & 12.676 \\
\texttt{3B r16 full\_attention} & 0.586 [0.563, 0.606] & 0.803 [0.775, 0.828] & 0.813 [0.786, 0.841] & 0.807 & 12.719 \\
\texttt{3B r16 qv\_only} & 0.581 [0.557, 0.604] & 0.787 [0.759, 0.817] & 0.792 [0.763, 0.819] & 0.633 & 12.701 \\
\texttt{3B r32 full\_attention} & 0.579 [0.556, 0.602] & 0.800 [0.772, 0.827] & 0.809 [0.780, 0.833] & 0.759 & 12.732 \\
\texttt{3B r32 qv\_only} & 0.589 [0.566, 0.609] & 0.796 [0.767, 0.825] & 0.806 [0.777, 0.832] & 0.630 & 12.719 \\
\texttt{3B r64 full\_attention} & 0.575 [0.554, 0.598] & 0.782 [0.754, 0.810] & 0.800 [0.773, 0.829] & 0.782 & 12.924 \\
\texttt{3B r64 qv\_only} & 0.597 [0.573, 0.621] & 0.787 [0.760, 0.814] & 0.800 [0.773, 0.827] & 0.620 & 12.760 \\
\texttt{8B baseline} & 0.581 [0.556, 0.605] & 0.764 [0.735, 0.794] & 0.768 [0.740, 0.796] & 0.810 & 20.795 \\
\texttt{8B r4 full\_attention} & 0.603 [0.582, 0.623] & 0.817 [0.791, 0.843] & 0.825 [0.799, 0.850] & 0.787 & 21.871 \\
\texttt{8B r4 qv\_only} & 0.610 [0.588, 0.631] & 0.818 [0.791, 0.843] & 0.827 [0.796, 0.851] & 0.676 & 21.828 \\
\texttt{8B r8 full\_attention} & 0.599 [0.575, 0.621] & 0.817 [0.790, 0.842] & 0.822 [0.795, 0.850] & 0.769 & 21.859 \\
\texttt{8B r8 qv\_only} & 0.606 [0.585, 0.628] & 0.824 [0.796, 0.851] & 0.831 [0.805, 0.855] & 0.673 & 21.836 \\
\texttt{8B r16 full\_attention} & 0.604 [0.582, 0.626] & 0.809 [0.782, 0.834] & 0.813 [0.785, 0.839] & 0.801 & 21.885 \\
\texttt{8B r16 qv\_only} & 0.615 [0.592, 0.635] & 0.817 [0.789, 0.841] & 0.831 [0.805, 0.856] & 0.655 & 21.824 \\
\texttt{8B r32 full\_attention} & 0.602 [0.581, 0.624] & 0.799 [0.769, 0.825] & 0.806 [0.778, 0.832] & 0.812 & 21.961 \\
\texttt{8B r32 qv\_only} & 0.612 [0.591, 0.635] & 0.813 [0.785, 0.841] & 0.818 [0.789, 0.843] & 0.667 & 21.850 \\
\texttt{8B r64 full\_attention} & 0.606 [0.582, 0.627] & 0.815 [0.789, 0.841] & 0.817 [0.790, 0.843] & 0.835 & 22.164 \\
\texttt{8B r64 qv\_only} & 0.617 [0.595, 0.640] & 0.818 [0.791, 0.845] & 0.828 [0.801, 0.854] & 0.650 & 21.926 \\
\bottomrule
\end{tabular}
}
\caption{Per-configuration metrics for the \texttt{05\_dense\_only\_\_neutral} regime: F1, groundedness, and correctness (pass@4) with 95\% bootstrap CI (1000 resamples on the final test split), inference latency, and peak inference VRAM.}
\label{tab:appA-05-dense-only--neutral}
\end{table}

\clearpage

\subsection*{\texttt{06\_dense\_only\_\_explicit\_grounded}}

\begin{table}[H]
\centering
\footnotesize
\setlength{\tabcolsep}{3pt}
\resizebox{\textwidth}{!}{
\begin{tabular}{lccccc}
\toprule
config & F1 [95\% CI] & grnd@4 [95\% CI] & corr@4 [95\% CI] & lat. (s) & inf. VRAM (GB) \\
\midrule
\texttt{3B baseline} & 0.532 [0.508, 0.555] & 0.734 [0.702, 0.766] & 0.727 [0.697, 0.757] & 0.579 & 11.676 \\
\texttt{3B r4 full\_attention} & 0.559 [0.537, 0.583] & 0.783 [0.754, 0.811] & 0.799 [0.772, 0.825] & 0.778 & 12.695 \\
\texttt{3B r4 qv\_only} & 0.566 [0.544, 0.587] & 0.804 [0.776, 0.831] & 0.810 [0.782, 0.837] & 0.636 & 12.689 \\
\texttt{3B r8 full\_attention} & 0.569 [0.546, 0.590] & 0.792 [0.764, 0.819] & 0.804 [0.776, 0.831] & 0.792 & 12.723 \\
\texttt{3B r8 qv\_only} & 0.571 [0.548, 0.593] & 0.783 [0.755, 0.809] & 0.799 [0.772, 0.827] & 0.660 & 12.697 \\
\texttt{3B r16 full\_attention} & 0.577 [0.556, 0.598] & 0.790 [0.762, 0.818] & 0.803 [0.776, 0.829] & 0.827 & 12.740 \\
\texttt{3B r16 qv\_only} & 0.574 [0.551, 0.597] & 0.790 [0.761, 0.819] & 0.795 [0.768, 0.822] & 0.628 & 12.723 \\
\texttt{3B r32 full\_attention} & 0.567 [0.544, 0.590] & 0.790 [0.759, 0.818] & 0.803 [0.776, 0.829] & 0.789 & 12.754 \\
\texttt{3B r32 qv\_only} & 0.576 [0.552, 0.598] & 0.782 [0.752, 0.811] & 0.792 [0.764, 0.820] & 0.659 & 12.740 \\
\texttt{3B r64 full\_attention} & 0.573 [0.550, 0.595] & 0.795 [0.769, 0.822] & 0.809 [0.781, 0.836] & 0.804 & 12.945 \\
\texttt{3B r64 qv\_only} & 0.598 [0.574, 0.621] & 0.783 [0.755, 0.814] & 0.792 [0.762, 0.820] & 0.668 & 12.781 \\
\texttt{8B baseline} & 0.586 [0.560, 0.611] & 0.769 [0.740, 0.800] & 0.767 [0.739, 0.795] & 0.732 & 20.797 \\
\texttt{8B r4 full\_attention} & 0.602 [0.581, 0.625] & 0.823 [0.794, 0.850] & 0.828 [0.803, 0.855] & 0.810 & 21.863 \\
\texttt{8B r4 qv\_only} & 0.602 [0.580, 0.623] & 0.813 [0.783, 0.839] & 0.823 [0.799, 0.847] & 0.700 & 21.855 \\
\texttt{8B r8 full\_attention} & 0.601 [0.581, 0.624] & 0.820 [0.792, 0.848] & 0.827 [0.800, 0.851] & 0.786 & 21.852 \\
\texttt{8B r8 qv\_only} & 0.606 [0.582, 0.627] & 0.823 [0.796, 0.850] & 0.832 [0.806, 0.856] & 0.688 & 21.863 \\
\texttt{8B r16 full\_attention} & 0.601 [0.579, 0.625] & 0.815 [0.790, 0.841] & 0.819 [0.792, 0.847] & 0.821 & 21.877 \\
\texttt{8B r16 qv\_only} & 0.610 [0.588, 0.633] & 0.827 [0.799, 0.854] & 0.832 [0.805, 0.856] & 0.679 & 21.852 \\
\texttt{8B r32 full\_attention} & 0.601 [0.579, 0.622] & 0.801 [0.773, 0.829] & 0.808 [0.778, 0.836] & 0.871 & 21.953 \\
\texttt{8B r32 qv\_only} & 0.610 [0.588, 0.632] & 0.814 [0.785, 0.841] & 0.820 [0.792, 0.846] & 0.680 & 21.877 \\
\texttt{8B r64 full\_attention} & 0.603 [0.580, 0.624] & 0.809 [0.781, 0.838] & 0.818 [0.790, 0.843] & 0.825 & 22.156 \\
\texttt{8B r64 qv\_only} & 0.623 [0.600, 0.645] & 0.824 [0.797, 0.850] & 0.827 [0.800, 0.851] & 0.661 & 21.953 \\
\bottomrule
\end{tabular}
}
\caption{Per-configuration metrics for the \texttt{06\_dense\_only\_\_explicit\_grounded} regime: F1, groundedness, and correctness (pass@4) with 95\% bootstrap CI (1000 resamples on the final test split), inference latency, and peak inference VRAM.}
\label{tab:appA-06-dense-only--explicit-grounded}
\end{table}

\clearpage

\subsection*{\texttt{07\_sparse\_only\_\_neutral}}

\begin{table}[H]
\centering
\footnotesize
\setlength{\tabcolsep}{3pt}
\resizebox{\textwidth}{!}{
\begin{tabular}{lccccc}
\toprule
config & F1 [95\% CI] & grnd@4 [95\% CI] & corr@4 [95\% CI] & lat. (s) & inf. VRAM (GB) \\
\midrule
\texttt{3B baseline} & 0.532 [0.507, 0.555] & 0.713 [0.680, 0.744] & 0.716 [0.683, 0.746] & 0.523 & 11.643 \\
\texttt{3B r4 full\_attention} & 0.573 [0.552, 0.594] & 0.790 [0.762, 0.819] & 0.803 [0.776, 0.828] & 0.746 & 12.674 \\
\texttt{3B r4 qv\_only} & 0.571 [0.549, 0.592] & 0.804 [0.775, 0.831] & 0.809 [0.781, 0.839] & 0.613 & 12.668 \\
\texttt{3B r8 full\_attention} & 0.579 [0.557, 0.601] & 0.796 [0.766, 0.824] & 0.809 [0.780, 0.836] & 0.731 & 12.701 \\
\texttt{3B r8 qv\_only} & 0.574 [0.552, 0.596] & 0.795 [0.766, 0.823] & 0.809 [0.781, 0.836] & 0.671 & 12.676 \\
\texttt{3B r16 full\_attention} & 0.586 [0.564, 0.608] & 0.809 [0.781, 0.836] & 0.815 [0.789, 0.842] & 0.756 & 12.719 \\
\texttt{3B r16 qv\_only} & 0.581 [0.559, 0.604] & 0.796 [0.768, 0.822] & 0.803 [0.776, 0.828] & 0.619 & 12.701 \\
\texttt{3B r32 full\_attention} & 0.579 [0.556, 0.601] & 0.794 [0.767, 0.822] & 0.813 [0.785, 0.839] & 0.897 & 12.732 \\
\texttt{3B r32 qv\_only} & 0.589 [0.567, 0.612] & 0.791 [0.763, 0.819] & 0.800 [0.771, 0.827] & 0.641 & 12.719 \\
\texttt{3B r64 full\_attention} & 0.575 [0.552, 0.600] & 0.796 [0.769, 0.824] & 0.809 [0.781, 0.834] & 0.791 & 12.924 \\
\texttt{3B r64 qv\_only} & 0.597 [0.573, 0.621] & 0.791 [0.764, 0.819] & 0.797 [0.769, 0.824] & 0.595 & 12.760 \\
\texttt{8B baseline} & 0.588 [0.564, 0.613] & 0.763 [0.734, 0.794] & 0.757 [0.725, 0.786] & 0.666 & 20.396 \\
\texttt{8B r4 full\_attention} & 0.603 [0.580, 0.626] & 0.827 [0.800, 0.854] & 0.832 [0.805, 0.857] & 0.801 & 21.871 \\
\texttt{8B r4 qv\_only} & 0.610 [0.588, 0.630] & 0.820 [0.791, 0.847] & 0.825 [0.799, 0.850] & 0.694 & 21.828 \\
\texttt{8B r8 full\_attention} & 0.599 [0.576, 0.620] & 0.817 [0.787, 0.843] & 0.820 [0.794, 0.846] & 0.758 & 21.859 \\
\texttt{8B r8 qv\_only} & 0.606 [0.584, 0.627] & 0.823 [0.795, 0.850] & 0.829 [0.804, 0.854] & 0.641 & 21.836 \\
\texttt{8B r16 full\_attention} & 0.604 [0.583, 0.627] & 0.808 [0.778, 0.834] & 0.818 [0.791, 0.845] & 0.794 & 21.885 \\
\texttt{8B r16 qv\_only} & 0.615 [0.593, 0.636] & 0.825 [0.800, 0.851] & 0.831 [0.804, 0.856] & 0.654 & 21.824 \\
\texttt{8B r32 full\_attention} & 0.602 [0.580, 0.622] & 0.797 [0.768, 0.826] & 0.804 [0.773, 0.832] & 0.842 & 21.961 \\
\texttt{8B r32 qv\_only} & 0.612 [0.590, 0.635] & 0.811 [0.785, 0.837] & 0.824 [0.800, 0.851] & 0.637 & 21.850 \\
\texttt{8B r64 full\_attention} & 0.606 [0.582, 0.627] & 0.815 [0.786, 0.842] & 0.818 [0.790, 0.845] & 0.827 & 22.164 \\
\texttt{8B r64 qv\_only} & 0.617 [0.595, 0.640] & 0.820 [0.792, 0.848] & 0.823 [0.799, 0.847] & 0.661 & 21.926 \\
\bottomrule
\end{tabular}
}
\caption{Per-configuration metrics for the \texttt{07\_sparse\_only\_\_neutral} regime: F1, groundedness, and correctness (pass@4) with 95\% bootstrap CI (1000 resamples on the final test split), inference latency, and peak inference VRAM.}
\label{tab:appA-07-sparse-only--neutral}
\end{table}

\clearpage

\subsection*{\texttt{08\_sparse\_only\_\_explicit\_grounded}}

\begin{table}[H]
\centering
\footnotesize
\setlength{\tabcolsep}{3pt}
\resizebox{\textwidth}{!}{
\begin{tabular}{lccccc}
\toprule
config & F1 [95\% CI] & grnd@4 [95\% CI] & corr@4 [95\% CI] & lat. (s) & inf. VRAM (GB) \\
\midrule
\texttt{3B baseline} & 0.533 [0.508, 0.559] & 0.721 [0.689, 0.752] & 0.717 [0.687, 0.750] & 0.513 & 11.643 \\
\texttt{3B r4 full\_attention} & 0.561 [0.539, 0.584] & 0.781 [0.750, 0.810] & 0.790 [0.762, 0.817] & 0.749 & 12.695 \\
\texttt{3B r4 qv\_only} & 0.561 [0.540, 0.582] & 0.800 [0.771, 0.827] & 0.808 [0.778, 0.837] & 0.623 & 12.689 \\
\texttt{3B r8 full\_attention} & 0.569 [0.546, 0.591] & 0.791 [0.763, 0.818] & 0.804 [0.777, 0.829] & 0.751 & 12.723 \\
\texttt{3B r8 qv\_only} & 0.569 [0.549, 0.591] & 0.777 [0.746, 0.806] & 0.796 [0.768, 0.823] & 0.669 & 12.697 \\
\texttt{3B r16 full\_attention} & 0.575 [0.552, 0.598] & 0.787 [0.757, 0.814] & 0.797 [0.769, 0.827] & 0.779 & 12.740 \\
\texttt{3B r16 qv\_only} & 0.571 [0.550, 0.593] & 0.780 [0.749, 0.806] & 0.791 [0.761, 0.818] & 0.624 & 12.723 \\
\texttt{3B r32 full\_attention} & 0.572 [0.550, 0.593] & 0.797 [0.771, 0.825] & 0.805 [0.778, 0.831] & 0.806 & 12.754 \\
\texttt{3B r32 qv\_only} & 0.582 [0.558, 0.604] & 0.782 [0.754, 0.810] & 0.791 [0.763, 0.818] & 0.733 & 12.740 \\
\texttt{3B r64 full\_attention} & 0.576 [0.553, 0.599] & 0.795 [0.766, 0.822] & 0.806 [0.777, 0.833] & 0.768 & 12.945 \\
\texttt{3B r64 qv\_only} & 0.593 [0.568, 0.614] & 0.776 [0.744, 0.804] & 0.785 [0.757, 0.813] & 0.605 & 12.781 \\
\texttt{8B baseline} & 0.596 [0.573, 0.618] & 0.771 [0.743, 0.800] & 0.768 [0.739, 0.799] & 0.670 & 20.396 \\
\texttt{8B r4 full\_attention} & 0.602 [0.578, 0.625] & 0.824 [0.797, 0.848] & 0.836 [0.808, 0.861] & 0.809 & 21.863 \\
\texttt{8B r4 qv\_only} & 0.602 [0.579, 0.623] & 0.819 [0.791, 0.843] & 0.831 [0.804, 0.856] & 0.706 & 21.855 \\
\texttt{8B r8 full\_attention} & 0.601 [0.581, 0.623] & 0.822 [0.792, 0.847] & 0.827 [0.800, 0.854] & 0.802 & 21.852 \\
\texttt{8B r8 qv\_only} & 0.606 [0.585, 0.627] & 0.822 [0.794, 0.848] & 0.831 [0.801, 0.860] & 0.653 & 21.863 \\
\texttt{8B r16 full\_attention} & 0.601 [0.580, 0.623] & 0.829 [0.803, 0.855] & 0.834 [0.808, 0.860] & 0.766 & 21.877 \\
\texttt{8B r16 qv\_only} & 0.610 [0.588, 0.633] & 0.823 [0.795, 0.850] & 0.831 [0.804, 0.856] & 0.665 & 21.852 \\
\texttt{8B r32 full\_attention} & 0.601 [0.578, 0.622] & 0.806 [0.777, 0.833] & 0.806 [0.780, 0.833] & 0.842 & 21.953 \\
\texttt{8B r32 qv\_only} & 0.610 [0.587, 0.631] & 0.813 [0.785, 0.841] & 0.818 [0.789, 0.846] & 0.654 & 21.877 \\
\texttt{8B r64 full\_attention} & 0.603 [0.578, 0.624] & 0.813 [0.782, 0.838] & 0.817 [0.791, 0.845] & 0.821 & 22.156 \\
\texttt{8B r64 qv\_only} & 0.623 [0.600, 0.645] & 0.828 [0.800, 0.854] & 0.832 [0.805, 0.856] & 0.646 & 21.953 \\
\bottomrule
\end{tabular}
}
\caption{Per-configuration metrics for the \texttt{08\_sparse\_only\_\_explicit\_grounded} regime: F1, groundedness, and correctness (pass@4) with 95\% bootstrap CI (1000 resamples on the final test split), inference latency, and peak inference VRAM.}
\label{tab:appA-08-sparse-only--explicit-grounded}
\end{table}

\clearpage

\subsection*{\texttt{09\_hybrid\_bm25\_\_neutral}}

\begin{table}[H]
\centering
\footnotesize
\setlength{\tabcolsep}{3pt}
\resizebox{\textwidth}{!}{
\begin{tabular}{lccccc}
\toprule
config & F1 [95\% CI] & grnd@4 [95\% CI] & corr@4 [95\% CI] & lat. (s) & inf. VRAM (GB) \\
\midrule
\texttt{3B baseline} & 0.543 [0.519, 0.568] & 0.729 [0.697, 0.759] & 0.721 [0.690, 0.752] & 0.617 & 11.789 \\
\texttt{3B r4 full\_attention} & 0.572 [0.550, 0.595] & 0.790 [0.762, 0.815] & 0.801 [0.772, 0.829] & 0.753 & 12.674 \\
\texttt{3B r4 qv\_only} & 0.569 [0.547, 0.592] & 0.797 [0.768, 0.824] & 0.803 [0.776, 0.829] & 0.635 & 12.668 \\
\texttt{3B r8 full\_attention} & 0.579 [0.557, 0.601] & 0.801 [0.772, 0.831] & 0.814 [0.785, 0.839] & 0.752 & 12.701 \\
\texttt{3B r8 qv\_only} & 0.571 [0.549, 0.591] & 0.797 [0.768, 0.824] & 0.804 [0.776, 0.831] & 0.628 & 12.676 \\
\texttt{3B r16 full\_attention} & 0.587 [0.564, 0.610] & 0.801 [0.773, 0.829] & 0.811 [0.783, 0.837] & 0.774 & 12.719 \\
\texttt{3B r16 qv\_only} & 0.577 [0.553, 0.601] & 0.785 [0.757, 0.813] & 0.797 [0.771, 0.827] & 0.624 & 12.701 \\
\texttt{3B r32 full\_attention} & 0.578 [0.556, 0.600] & 0.801 [0.773, 0.828] & 0.817 [0.787, 0.845] & 0.751 & 12.732 \\
\texttt{3B r32 qv\_only} & 0.593 [0.570, 0.617] & 0.792 [0.763, 0.822] & 0.803 [0.775, 0.833] & 0.652 & 12.719 \\
\texttt{3B r64 full\_attention} & 0.576 [0.553, 0.598] & 0.797 [0.771, 0.825] & 0.810 [0.781, 0.837] & 0.767 & 12.924 \\
\texttt{3B r64 qv\_only} & 0.597 [0.573, 0.621] & 0.786 [0.755, 0.815] & 0.792 [0.764, 0.820] & 0.673 & 12.760 \\
\texttt{8B baseline} & 0.596 [0.574, 0.620] & 0.778 [0.748, 0.806] & 0.778 [0.748, 0.806] & 0.769 & 20.768 \\
\texttt{8B r4 full\_attention} & 0.603 [0.581, 0.623] & 0.824 [0.797, 0.850] & 0.834 [0.808, 0.859] & 0.811 & 21.871 \\
\texttt{8B r4 qv\_only} & 0.610 [0.588, 0.633] & 0.819 [0.791, 0.847] & 0.829 [0.804, 0.856] & 0.675 & 21.828 \\
\texttt{8B r8 full\_attention} & 0.599 [0.576, 0.621] & 0.811 [0.785, 0.838] & 0.817 [0.790, 0.845] & 0.766 & 21.859 \\
\texttt{8B r8 qv\_only} & 0.606 [0.586, 0.627] & 0.827 [0.801, 0.852] & 0.836 [0.808, 0.861] & 0.646 & 21.836 \\
\texttt{8B r16 full\_attention} & 0.604 [0.581, 0.625] & 0.810 [0.782, 0.836] & 0.808 [0.780, 0.836] & 0.806 & 21.885 \\
\texttt{8B r16 qv\_only} & 0.615 [0.592, 0.636] & 0.818 [0.791, 0.845] & 0.824 [0.797, 0.848] & 0.660 & 21.824 \\
\texttt{8B r32 full\_attention} & 0.602 [0.581, 0.623] & 0.797 [0.768, 0.824] & 0.806 [0.777, 0.833] & 0.829 & 21.961 \\
\texttt{8B r32 qv\_only} & 0.612 [0.591, 0.635] & 0.819 [0.792, 0.846] & 0.825 [0.797, 0.852] & 0.649 & 21.850 \\
\texttt{8B r64 full\_attention} & 0.606 [0.583, 0.627] & 0.814 [0.787, 0.841] & 0.822 [0.796, 0.848] & 0.866 & 22.164 \\
\texttt{8B r64 qv\_only} & 0.617 [0.597, 0.639] & 0.815 [0.787, 0.842] & 0.824 [0.797, 0.851] & 0.659 & 21.926 \\
\bottomrule
\end{tabular}
}
\caption{Per-configuration metrics for the \texttt{09\_hybrid\_bm25\_\_neutral} regime: F1, groundedness, and correctness (pass@4) with 95\% bootstrap CI (1000 resamples on the final test split), inference latency, and peak inference VRAM.}
\label{tab:appA-09-hybrid-bm25--neutral}
\end{table}

\clearpage

\subsection*{\texttt{10\_hybrid\_bm25\_\_explicit\_grounded}}

\begin{table}[H]
\centering
\footnotesize
\setlength{\tabcolsep}{3pt}
\resizebox{\textwidth}{!}{
\begin{tabular}{lccccc}
\toprule
config & F1 [95\% CI] & grnd@4 [95\% CI] & corr@4 [95\% CI] & lat. (s) & inf. VRAM (GB) \\
\midrule
\texttt{3B baseline} & 0.543 [0.520, 0.567] & 0.735 [0.704, 0.766] & 0.729 [0.696, 0.758] & 0.615 & 11.789 \\
\texttt{3B r4 full\_attention} & 0.561 [0.538, 0.584] & 0.778 [0.752, 0.809] & 0.796 [0.769, 0.824] & 0.775 & 12.695 \\
\texttt{3B r4 qv\_only} & 0.561 [0.540, 0.585] & 0.799 [0.771, 0.825] & 0.806 [0.778, 0.833] & 0.656 & 12.689 \\
\texttt{3B r8 full\_attention} & 0.569 [0.547, 0.591] & 0.796 [0.768, 0.824] & 0.810 [0.782, 0.837] & 0.805 & 12.723 \\
\texttt{3B r8 qv\_only} & 0.569 [0.548, 0.590] & 0.778 [0.748, 0.806] & 0.791 [0.762, 0.819] & 0.667 & 12.697 \\
\texttt{3B r16 full\_attention} & 0.575 [0.554, 0.598] & 0.781 [0.753, 0.809] & 0.794 [0.762, 0.822] & 0.776 & 12.740 \\
\texttt{3B r16 qv\_only} & 0.571 [0.548, 0.596] & 0.786 [0.754, 0.814] & 0.794 [0.764, 0.820] & 0.642 & 12.723 \\
\texttt{3B r32 full\_attention} & 0.572 [0.549, 0.593] & 0.800 [0.771, 0.829] & 0.811 [0.783, 0.838] & 0.841 & 12.754 \\
\texttt{3B r32 qv\_only} & 0.582 [0.559, 0.604] & 0.780 [0.750, 0.808] & 0.791 [0.763, 0.818] & 0.737 & 12.740 \\
\texttt{3B r64 full\_attention} & 0.576 [0.551, 0.598] & 0.795 [0.767, 0.824] & 0.809 [0.782, 0.837] & 0.783 & 12.945 \\
\texttt{3B r64 qv\_only} & 0.593 [0.570, 0.615] & 0.780 [0.750, 0.809] & 0.792 [0.766, 0.822] & 0.593 & 12.781 \\
\texttt{8B baseline} & 0.603 [0.580, 0.625] & 0.781 [0.755, 0.808] & 0.781 [0.753, 0.810] & 0.734 & 20.768 \\
\texttt{8B r4 full\_attention} & 0.602 [0.580, 0.625] & 0.820 [0.795, 0.846] & 0.829 [0.804, 0.856] & 0.804 & 21.863 \\
\texttt{8B r4 qv\_only} & 0.602 [0.579, 0.623] & 0.808 [0.781, 0.834] & 0.818 [0.791, 0.843] & 0.714 & 21.855 \\
\texttt{8B r8 full\_attention} & 0.601 [0.580, 0.623] & 0.820 [0.792, 0.846] & 0.822 [0.792, 0.848] & 0.795 & 21.852 \\
\texttt{8B r8 qv\_only} & 0.606 [0.585, 0.627] & 0.832 [0.808, 0.857] & 0.842 [0.815, 0.869] & 0.675 & 21.863 \\
\texttt{8B r16 full\_attention} & 0.601 [0.581, 0.624] & 0.827 [0.800, 0.850] & 0.832 [0.806, 0.857] & 0.842 & 21.877 \\
\texttt{8B r16 qv\_only} & 0.610 [0.590, 0.631] & 0.834 [0.808, 0.860] & 0.838 [0.811, 0.861] & 0.691 & 21.852 \\
\texttt{8B r32 full\_attention} & 0.601 [0.579, 0.621] & 0.804 [0.776, 0.829] & 0.808 [0.780, 0.834] & 0.850 & 21.953 \\
\texttt{8B r32 qv\_only} & 0.610 [0.589, 0.633] & 0.808 [0.782, 0.836] & 0.813 [0.783, 0.838] & 0.672 & 21.877 \\
\texttt{8B r64 full\_attention} & 0.603 [0.580, 0.624] & 0.810 [0.785, 0.838] & 0.817 [0.790, 0.842] & 0.820 & 22.156 \\
\texttt{8B r64 qv\_only} & 0.623 [0.603, 0.645] & 0.823 [0.795, 0.847] & 0.829 [0.803, 0.855] & 0.664 & 21.953 \\
\bottomrule
\end{tabular}
}
\caption{Per-configuration metrics for the \texttt{10\_hybrid\_bm25\_\_explicit\_grounded} regime: F1, groundedness, and correctness (pass@4) with 95\% bootstrap CI (1000 resamples on the final test split), inference latency, and peak inference VRAM.}
\label{tab:appA-10-hybrid-bm25--explicit-grounded}
\end{table}

\section{Per-Regime Plots}
\label{app:plots}

This appendix contains the detailed plots for each of the 10 ablation regimes
referenced in Section~\ref{sec:results}. For every regime we show the main
two-dimensional projections used in the trade-off analysis. Plots with an
$F_1$ axis (\texttt{f1\_vs\_latency}, \texttt{f1\_vs\_inference\_vram},
\texttt{f1\_vs\_groundedness\_pass4}) are drawn with vertical error bars
corresponding to 95\% bootstrap confidence intervals on the final test split
($n = 785$, $1{,}000$ resamples). Analogously, for the judge metrics
\texttt{groundedness\_pass@4} and \texttt{correctness\_pass@4} (per-example
binomial indicators of passing the Likert-score threshold $\geq 4$) we
compute 95\% bootstrap CIs; the \texttt{groundedness\_pass4\_vs\_latency}
plot has vertical error bars on the \texttt{groundedness} axis, while
\texttt{f1\_vs\_groundedness\_pass4} has two-sided error bars (on both axes).
The point CI values are mirrored in the corresponding tables of
Appendix~\ref{app:tables}. The \texttt{f1\_vs\_inference\_vram} plot is
drawn as two adjacent panels (separate zoom on the 3B and 8B families),
because under a single axis the configuration points cluster into two narrow
bands of about 0.3~GB and overlap with each other; separate panels allow
the within-cluster structure to be distinguished. The
\texttt{groundedness\_pass4\_vs\_inference\_vram} projection is not included
in the appendix: in it the spread along the VRAM axis within each family
does not exceed 0.3--0.4~GB, while the corresponding \texttt{groundedness}
spread fully lies inside the 95\% CI band, so the plot carries no
information beyond \texttt{f1\_vs\_inference\_vram}.

\clearpage
\subsection*{\texttt{01\_base\_\_neutral}}
Regime: \texttt{base + neutral}.

\medskip
\noindent\begin{minipage}{\linewidth}
\textbf{F1 vs Latency}\par\nopagebreak\vspace{-1.5ex}
\begin{center}
\includegraphics[width=0.88\textwidth]{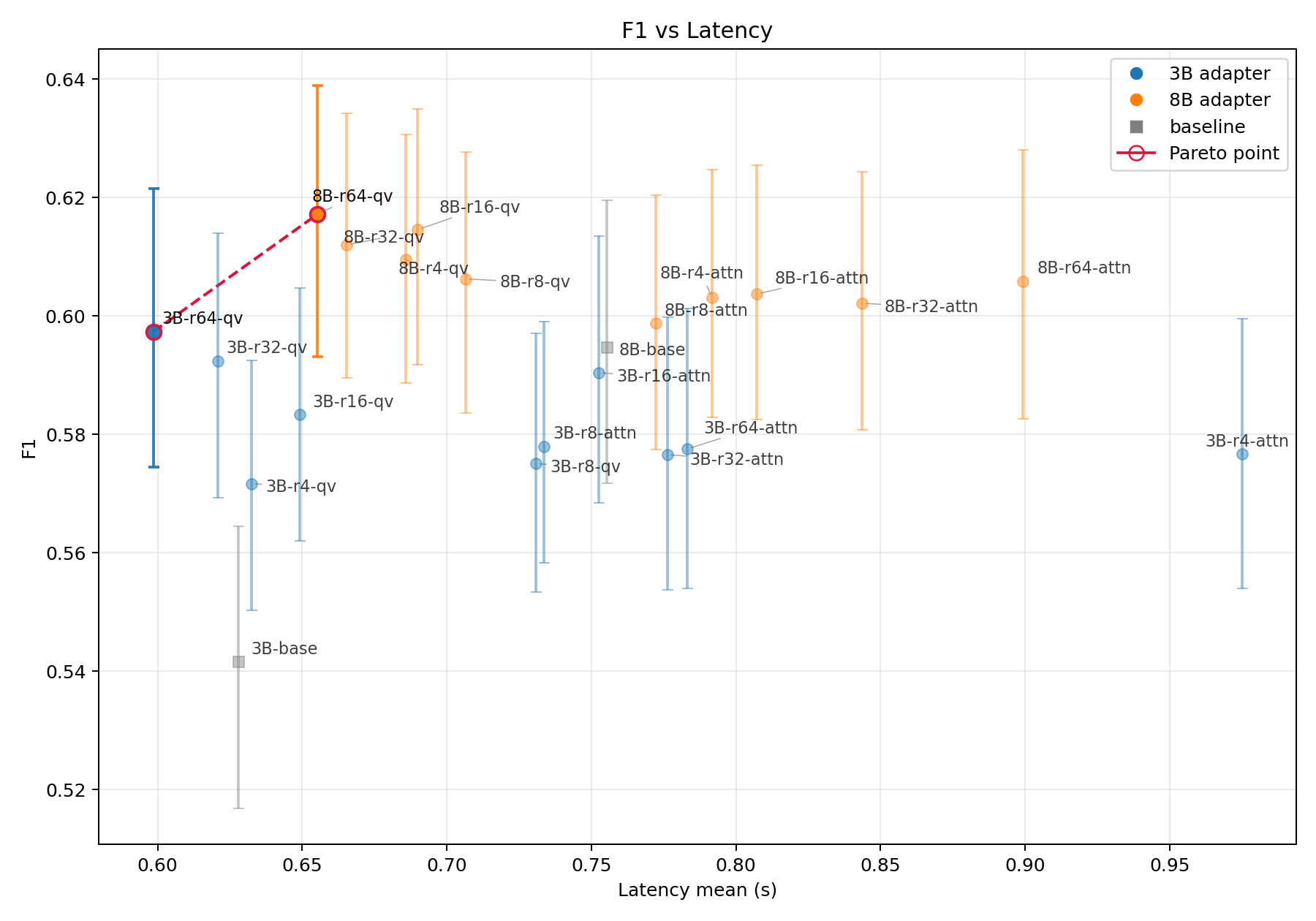}
\end{center}
\end{minipage}

\medskip
\noindent\begin{minipage}{\linewidth}
\textbf{F1 vs Inference VRAM}\par\nopagebreak\vspace{-1.5ex}
\begin{center}
\includegraphics[width=0.88\textwidth]{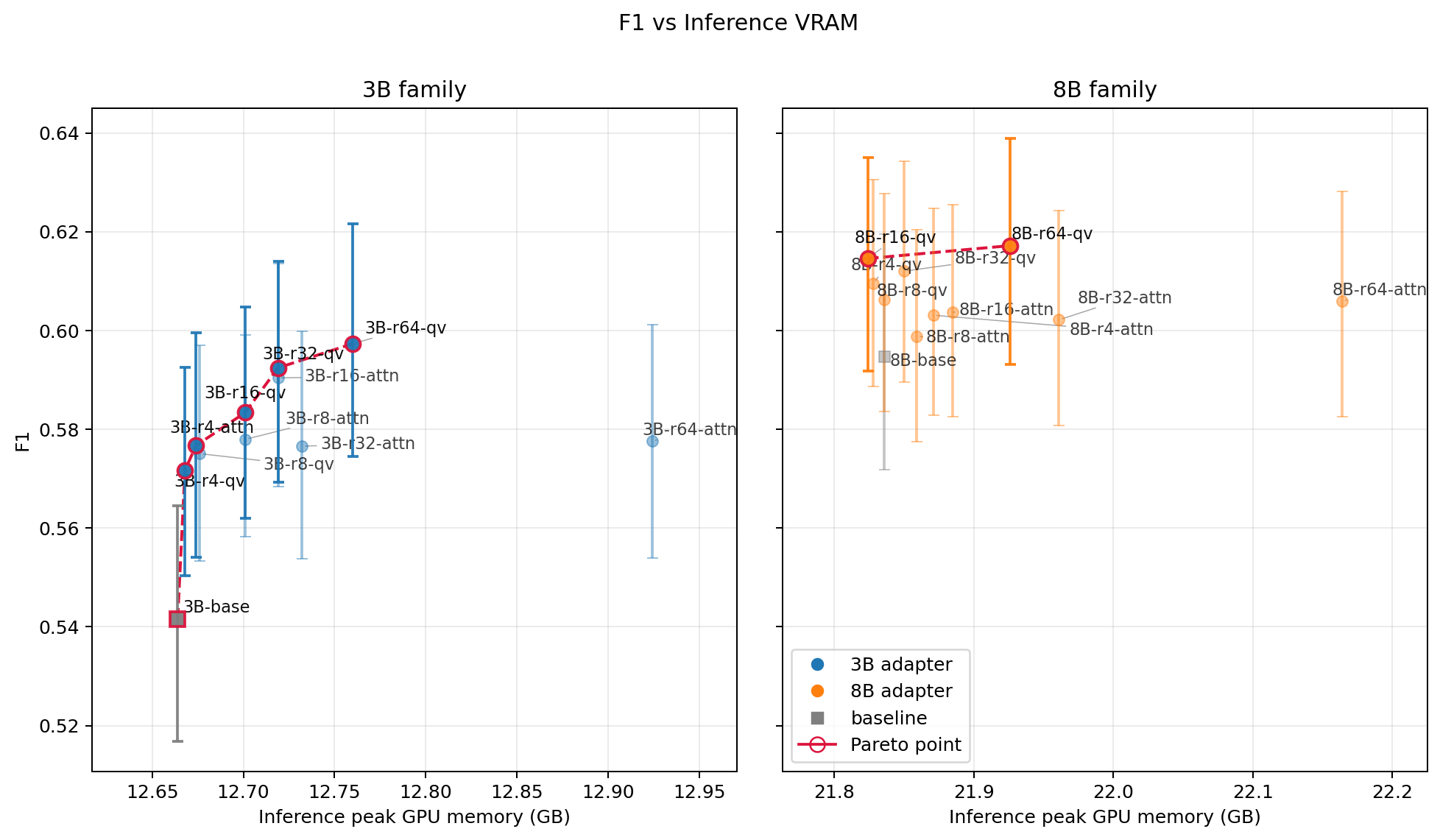}
\end{center}
\end{minipage}

\medskip
\noindent\begin{minipage}{\linewidth}
\textbf{F1 vs Groundedness pass@4}\par\nopagebreak\vspace{-1.5ex}
\begin{center}
\includegraphics[width=0.88\textwidth]{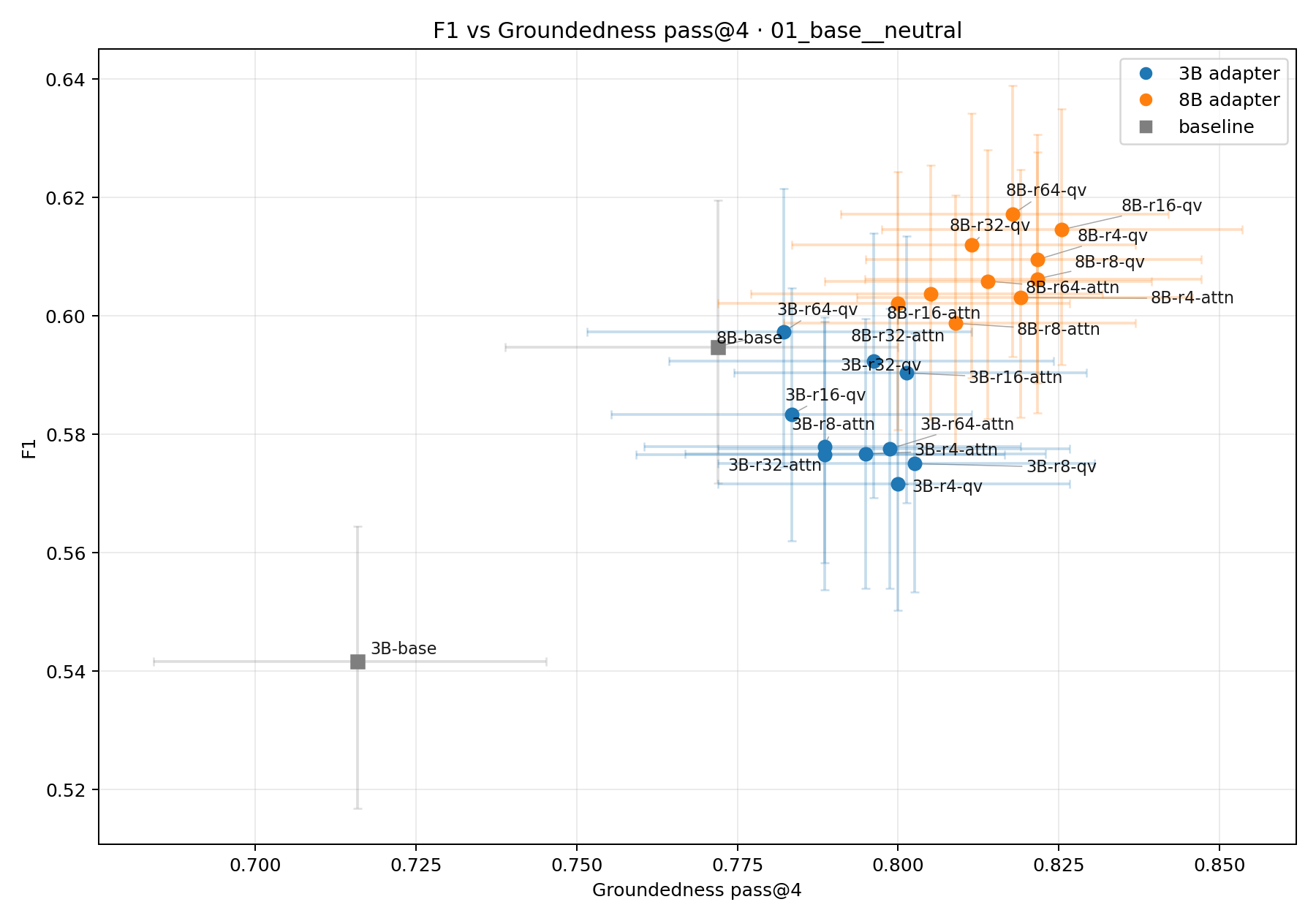}
\end{center}
\end{minipage}

\medskip
\noindent\begin{minipage}{\linewidth}
\textbf{Groundedness pass@4 vs Latency}\par\nopagebreak\vspace{-1.5ex}
\begin{center}
\includegraphics[width=0.88\textwidth]{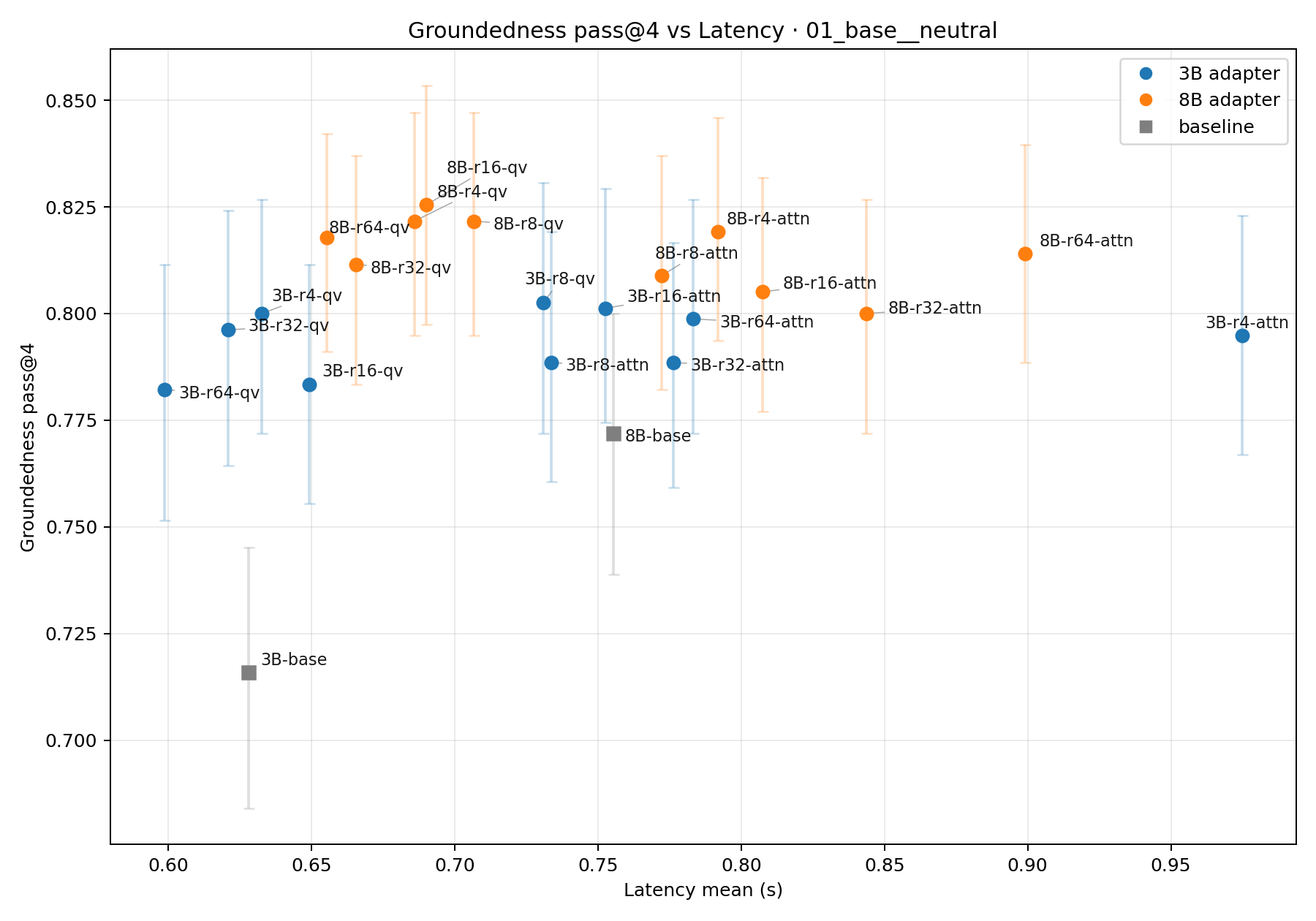}
\end{center}
\end{minipage}

\clearpage
\subsection*{\texttt{02\_base\_\_explicit\_grounded}}
Regime: \texttt{base + explicit\_grounded}.

\medskip
\noindent\begin{minipage}{\linewidth}
\textbf{F1 vs Latency}\par\nopagebreak\vspace{-1.5ex}
\begin{center}
\includegraphics[width=0.88\textwidth]{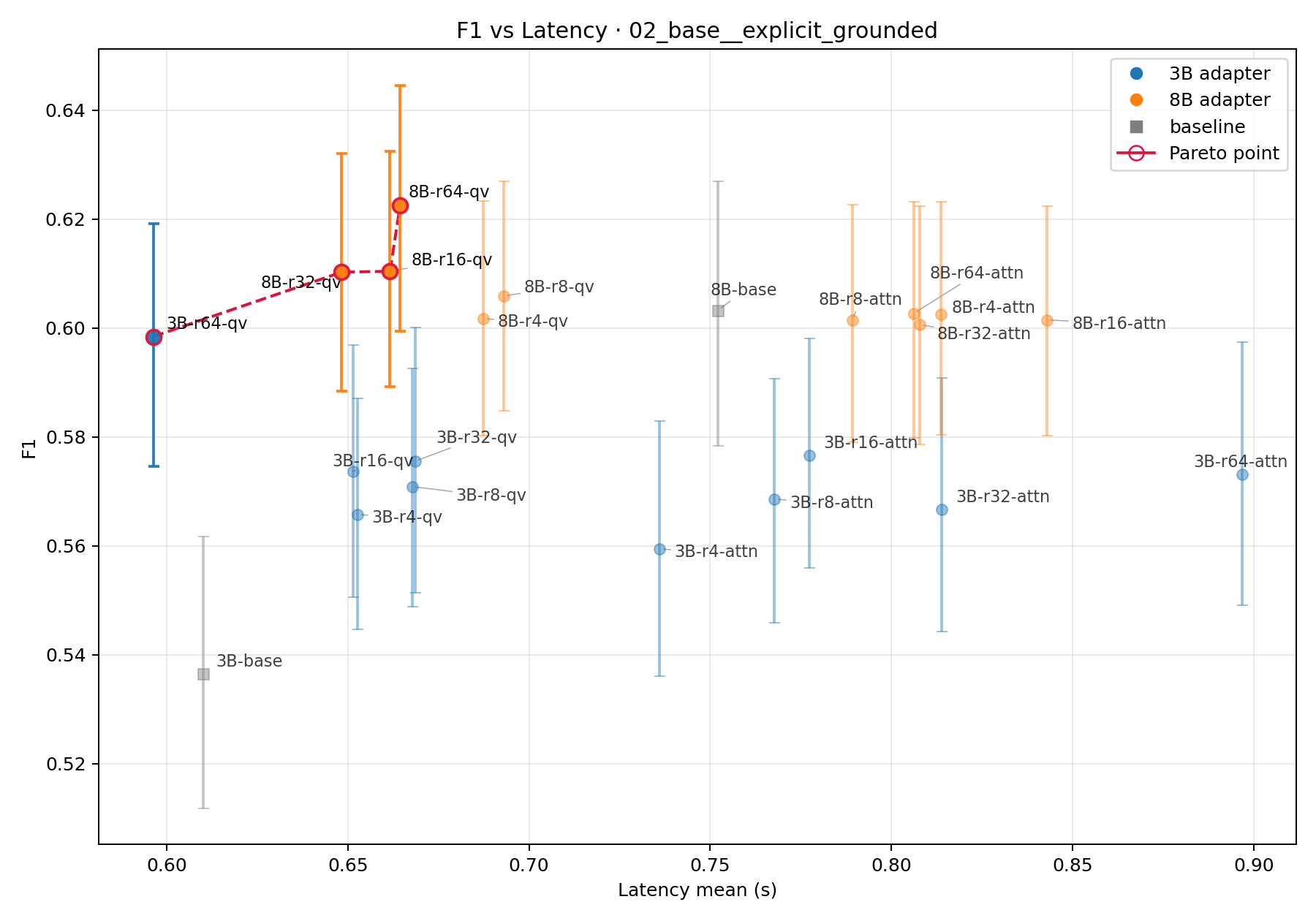}
\end{center}
\end{minipage}

\medskip
\noindent\begin{minipage}{\linewidth}
\textbf{F1 vs Inference VRAM}\par\nopagebreak\vspace{-1.5ex}
\begin{center}
\includegraphics[width=0.88\textwidth]{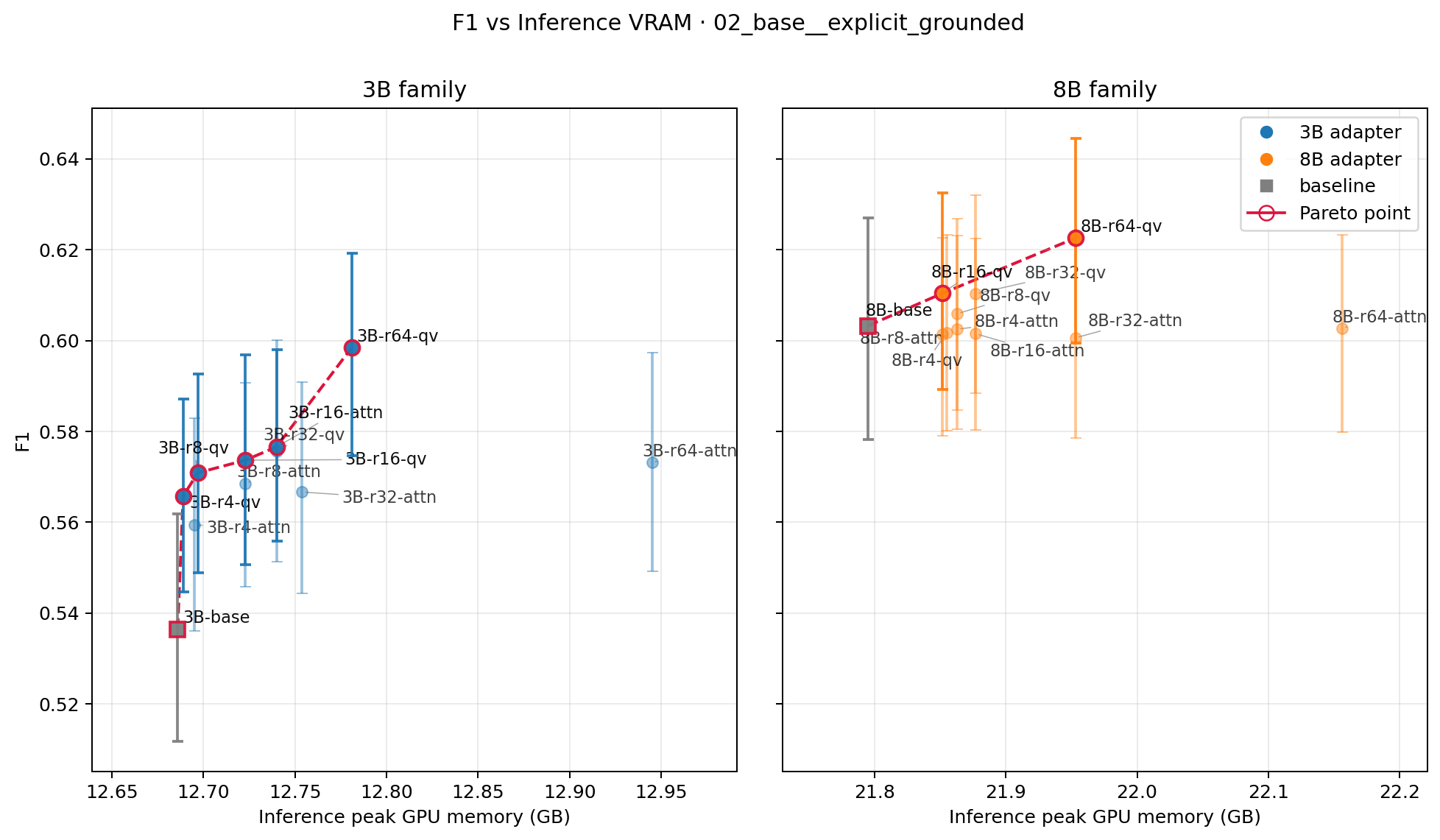}
\end{center}
\end{minipage}

\medskip
\noindent\begin{minipage}{\linewidth}
\textbf{F1 vs Groundedness pass@4}\par\nopagebreak\vspace{-1.5ex}
\begin{center}
\includegraphics[width=0.88\textwidth]{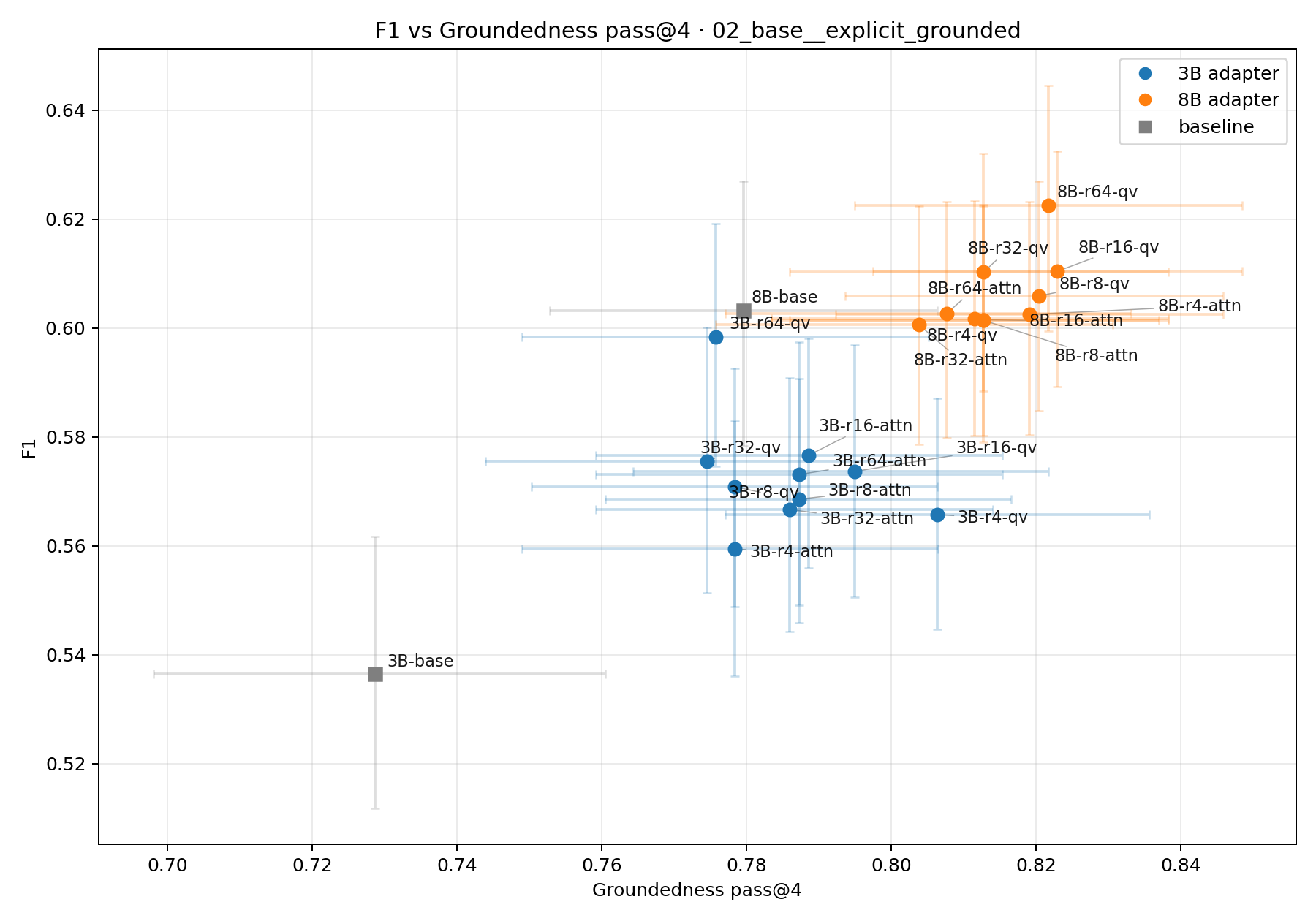}
\end{center}
\end{minipage}

\medskip
\noindent\begin{minipage}{\linewidth}
\textbf{Groundedness pass@4 vs Latency}\par\nopagebreak\vspace{-1.5ex}
\begin{center}
\includegraphics[width=0.88\textwidth]{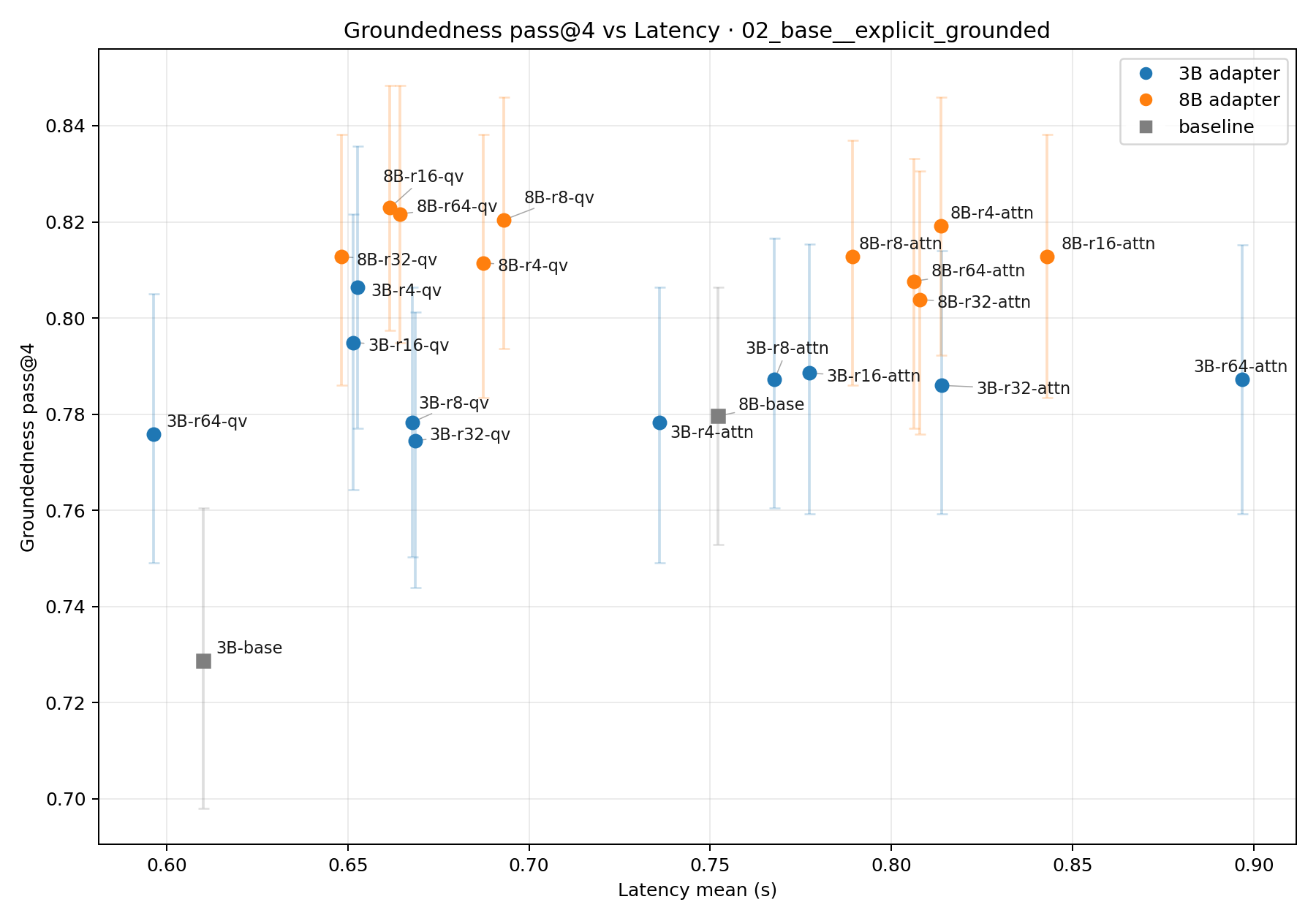}
\end{center}
\end{minipage}

\clearpage
\subsection*{\texttt{03\_reranker\_off\_\_neutral}}
Regime: \texttt{reranker\_off + neutral}.

\medskip
\noindent\begin{minipage}{\linewidth}
\textbf{F1 vs Latency}\par\nopagebreak\vspace{-1.5ex}
\begin{center}
\includegraphics[width=0.88\textwidth]{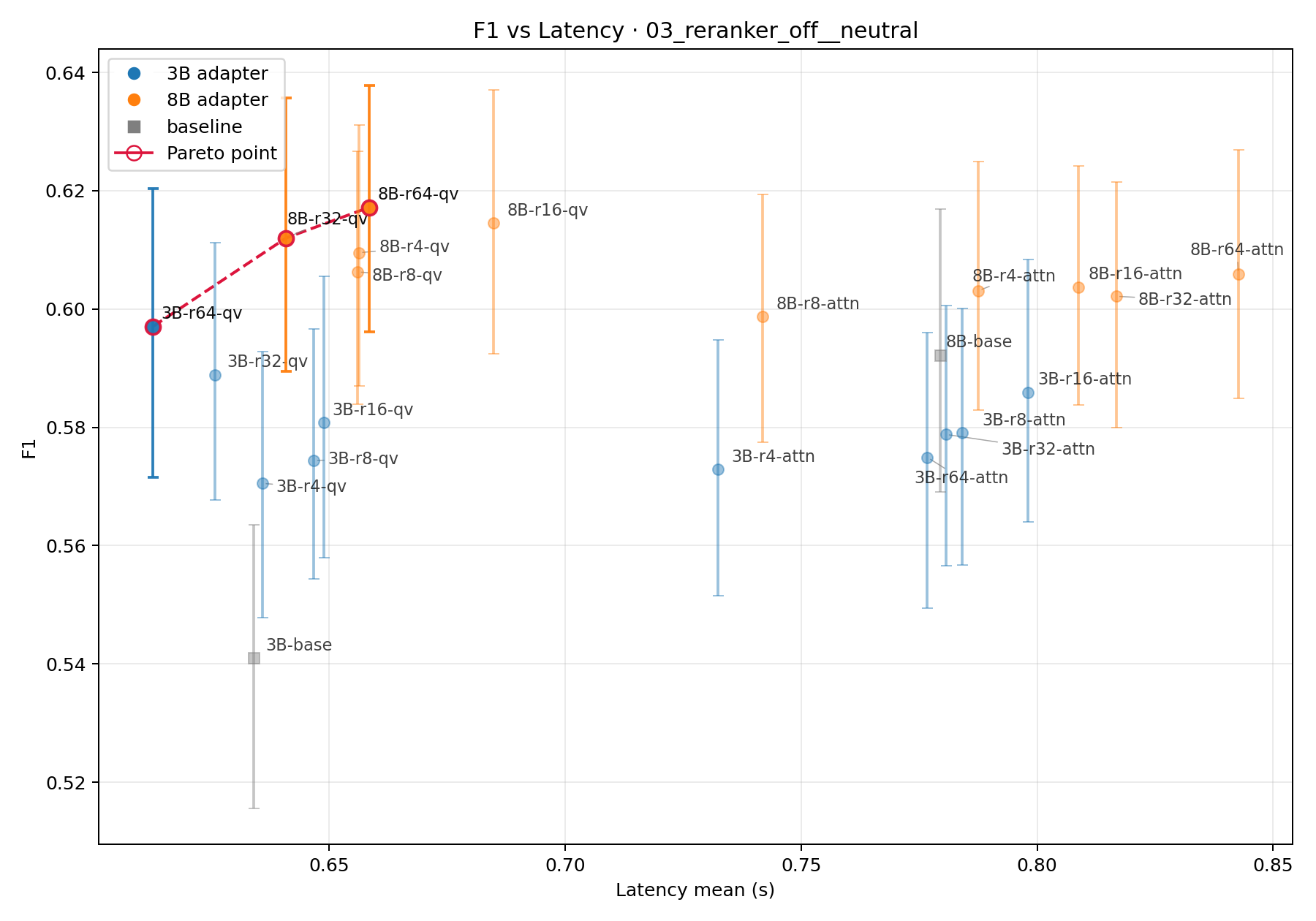}
\end{center}
\end{minipage}

\medskip
\noindent\begin{minipage}{\linewidth}
\textbf{F1 vs Inference VRAM}\par\nopagebreak\vspace{-1.5ex}
\begin{center}
\includegraphics[width=0.88\textwidth]{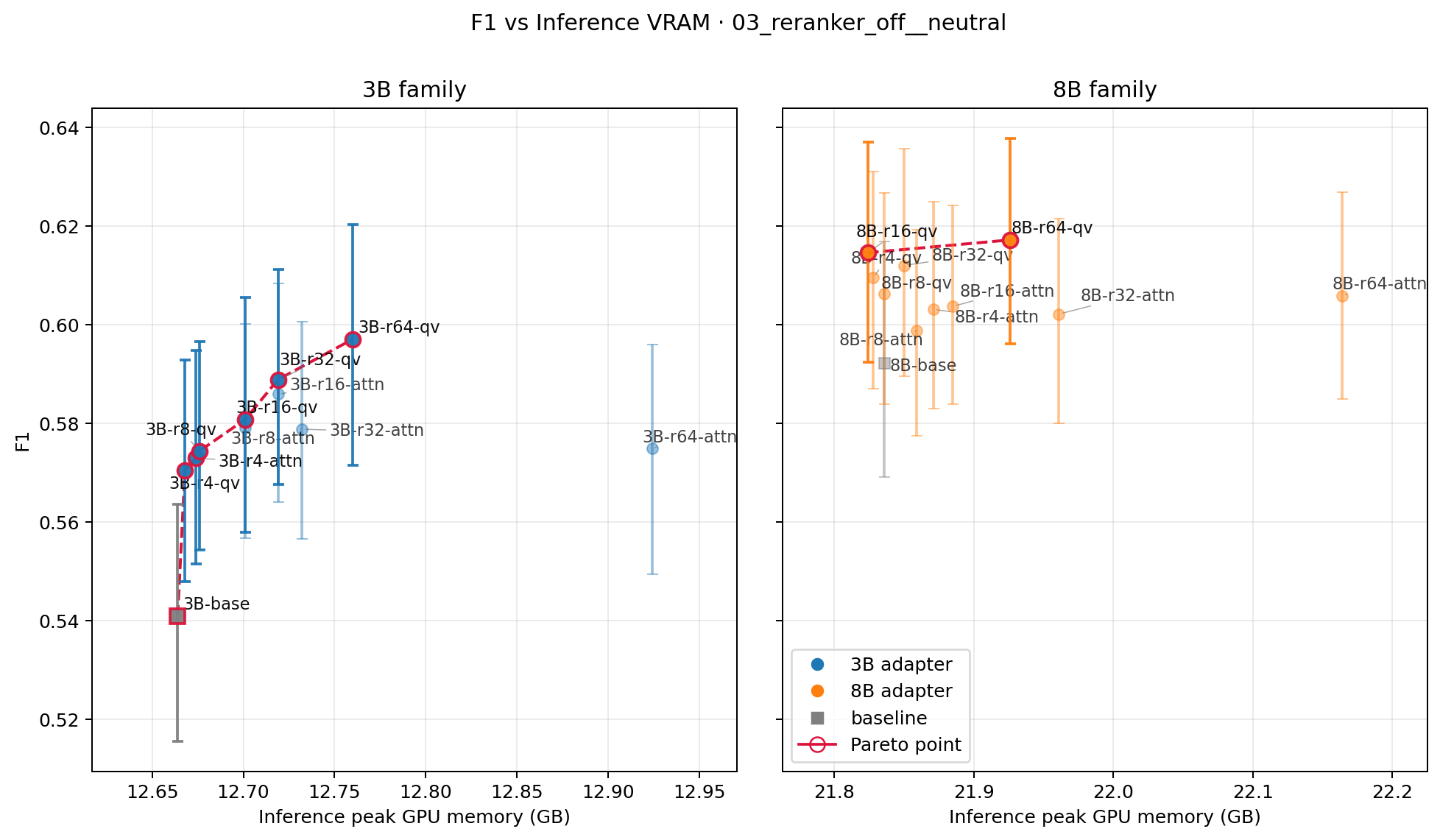}
\end{center}
\end{minipage}

\medskip
\noindent\begin{minipage}{\linewidth}
\textbf{F1 vs Groundedness pass@4}\par\nopagebreak\vspace{-1.5ex}
\begin{center}
\includegraphics[width=0.88\textwidth]{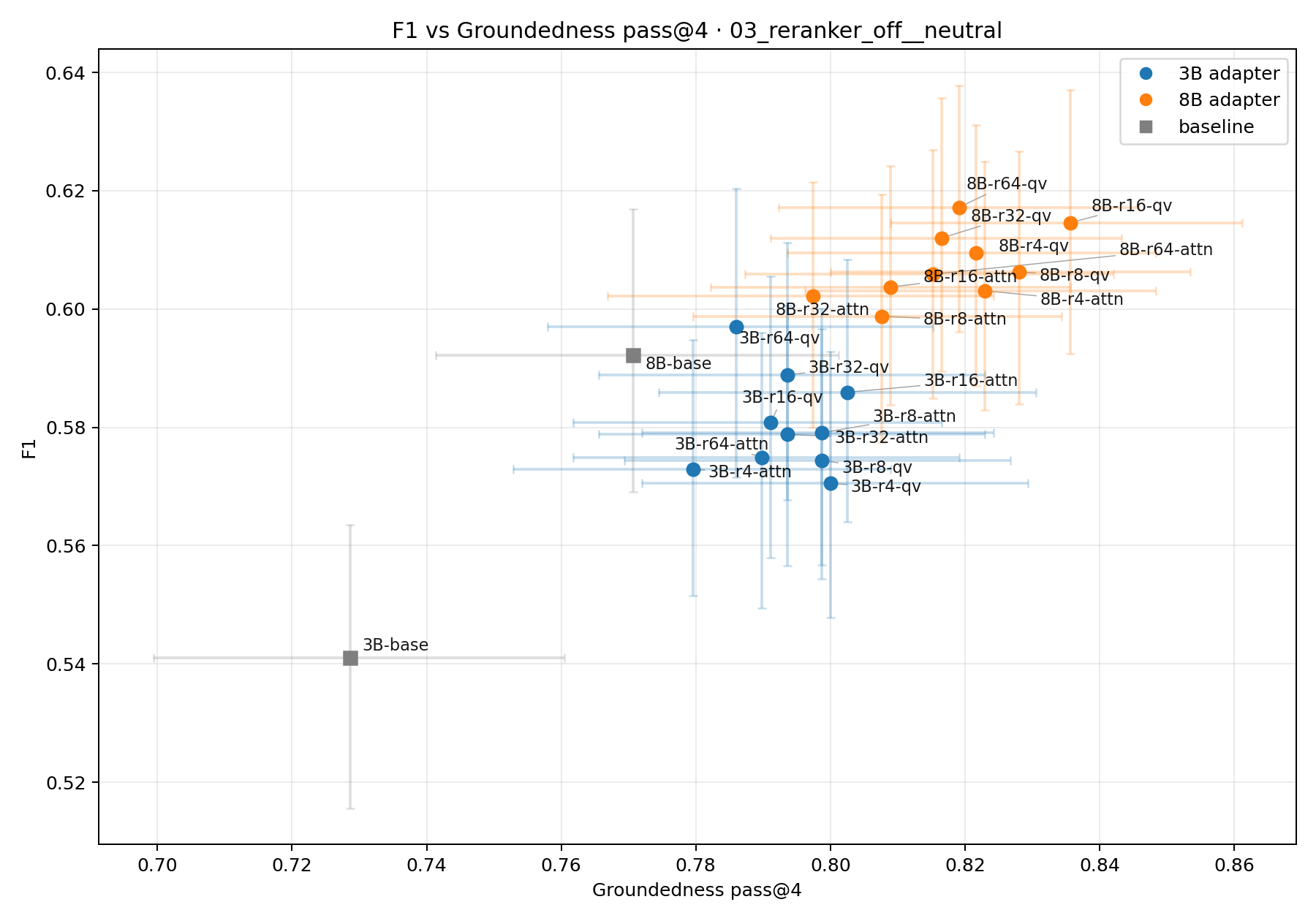}
\end{center}
\end{minipage}

\medskip
\noindent\begin{minipage}{\linewidth}
\textbf{Groundedness pass@4 vs Latency}\par\nopagebreak\vspace{-1.5ex}
\begin{center}
\includegraphics[width=0.88\textwidth]{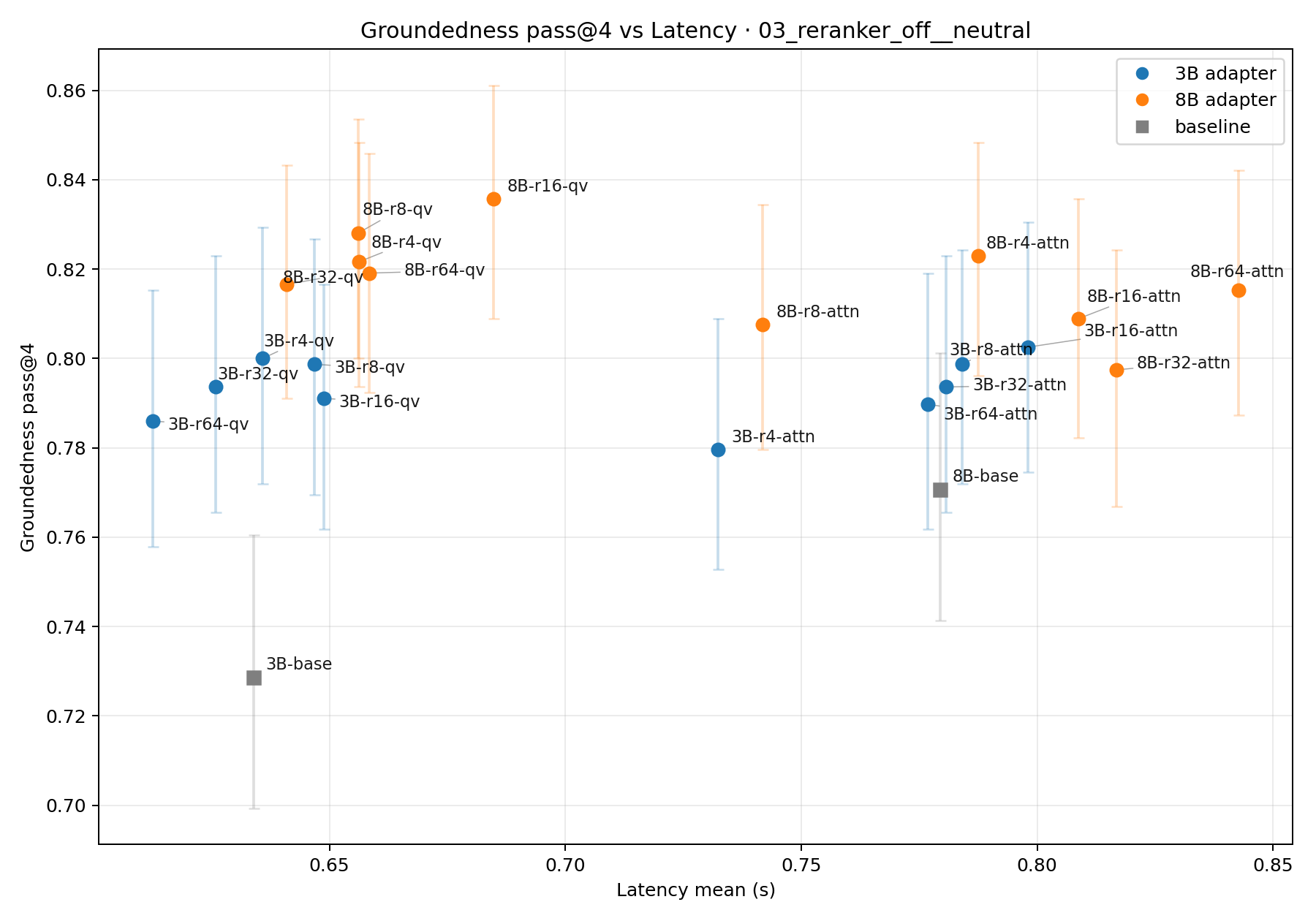}
\end{center}
\end{minipage}

\clearpage
\subsection*{\texttt{04\_reranker\_off\_\_explicit\_grounded}}
Regime: \texttt{reranker\_off + explicit\_grounded}.

\medskip
\noindent\begin{minipage}{\linewidth}
\textbf{F1 vs Latency}\par\nopagebreak\vspace{-1.5ex}
\begin{center}
\includegraphics[width=0.88\textwidth]{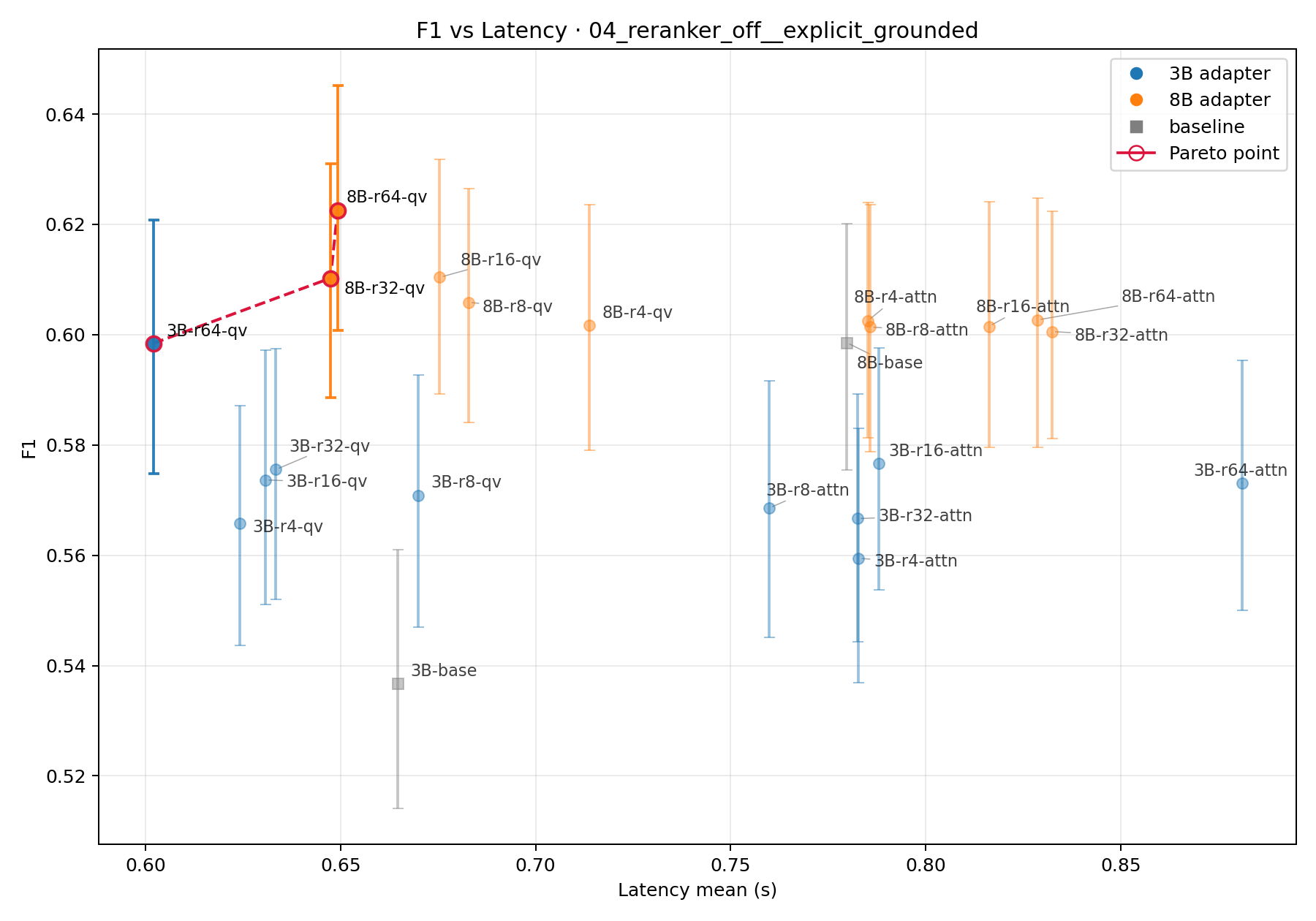}
\end{center}
\end{minipage}

\medskip
\noindent\begin{minipage}{\linewidth}
\textbf{F1 vs Inference VRAM}\par\nopagebreak\vspace{-1.5ex}
\begin{center}
\includegraphics[width=0.88\textwidth]{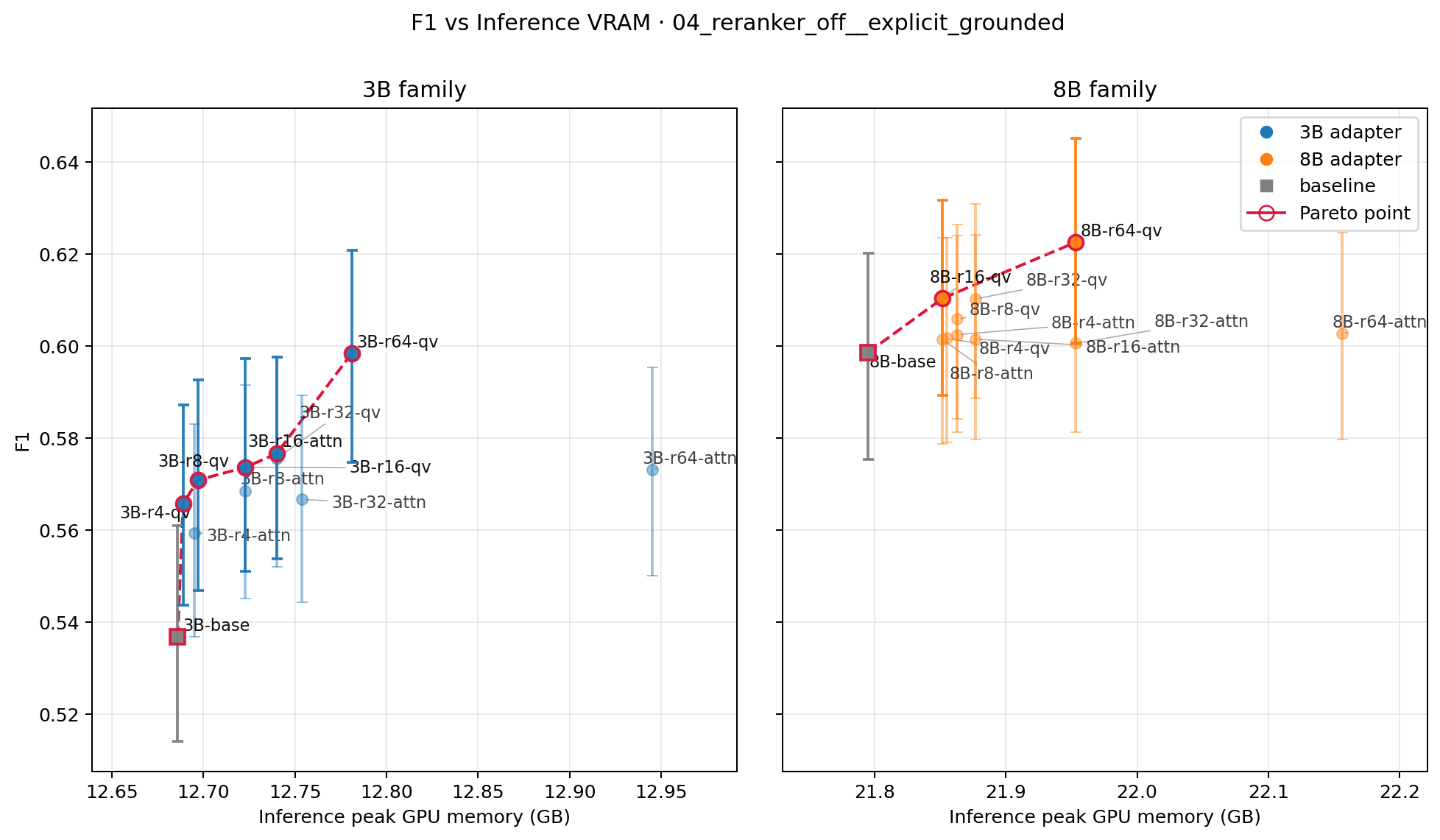}
\end{center}
\end{minipage}

\medskip
\noindent\begin{minipage}{\linewidth}
\textbf{F1 vs Groundedness pass@4}\par\nopagebreak\vspace{-1.5ex}
\begin{center}
\includegraphics[width=0.88\textwidth]{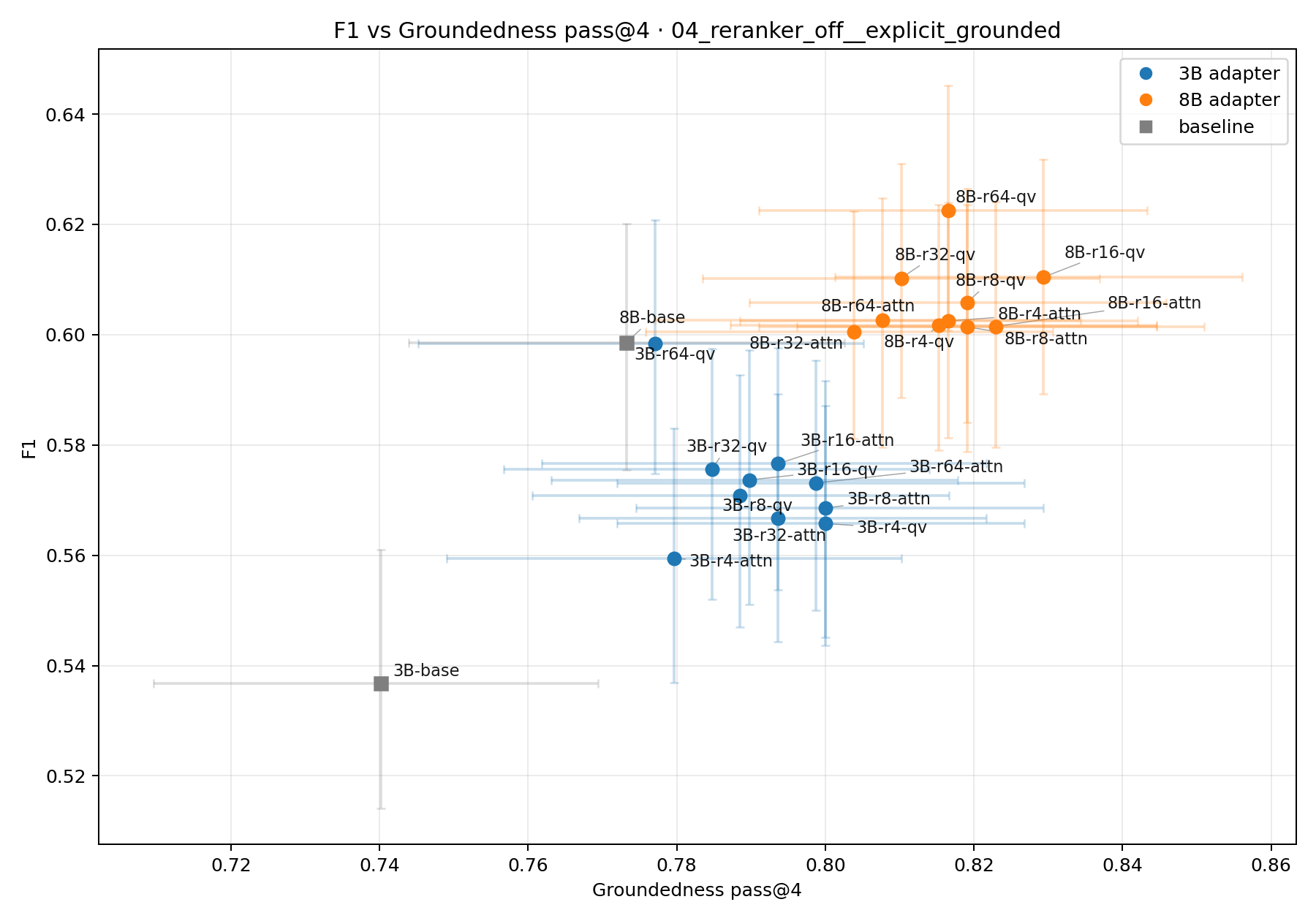}
\end{center}
\end{minipage}

\medskip
\noindent\begin{minipage}{\linewidth}
\textbf{Groundedness pass@4 vs Latency}\par\nopagebreak\vspace{-1.5ex}
\begin{center}
\includegraphics[width=0.88\textwidth]{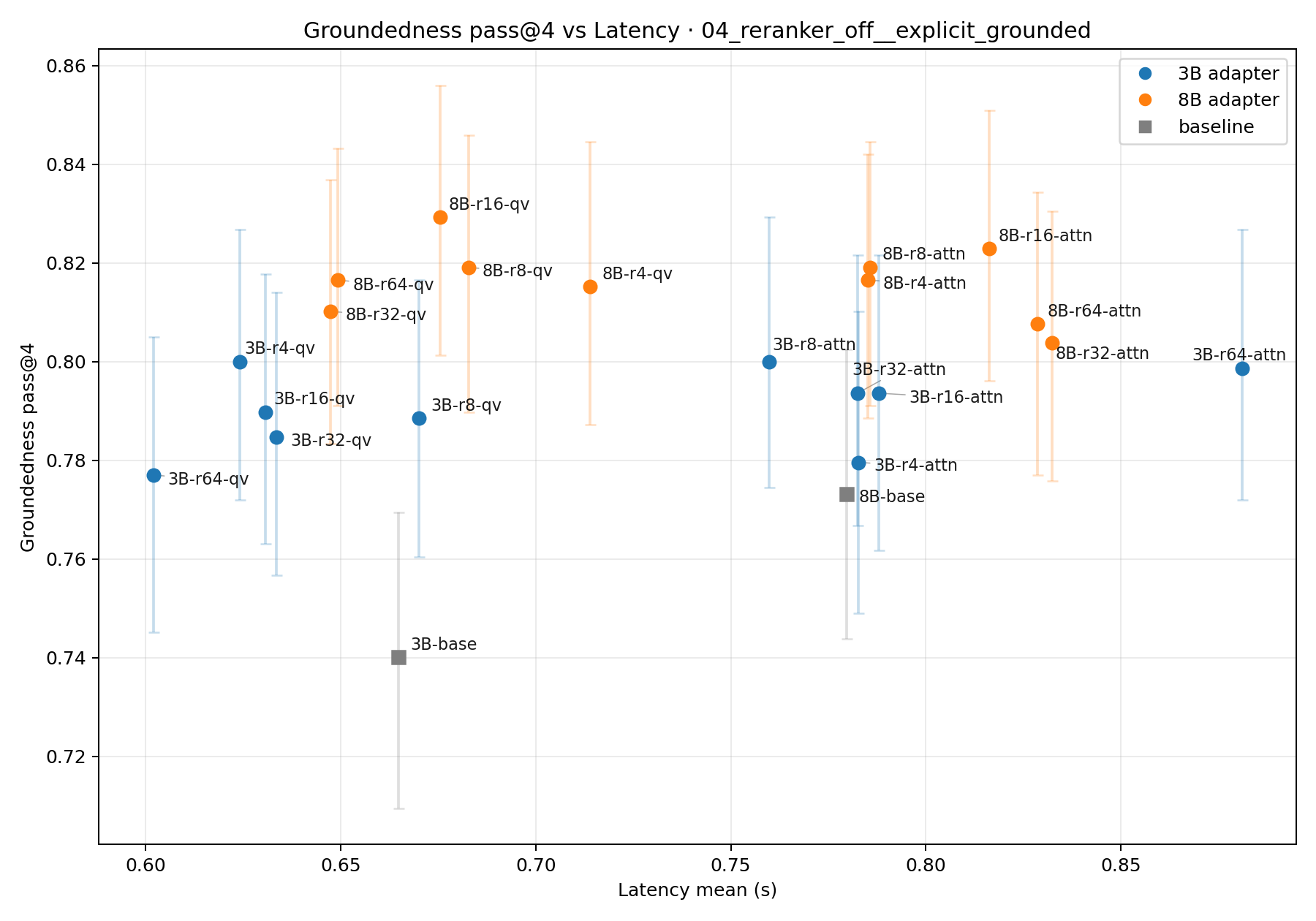}
\end{center}
\end{minipage}

\clearpage
\subsection*{\texttt{05\_dense\_only\_\_neutral}}
Regime: \texttt{dense\_only + neutral}.

\medskip
\noindent\begin{minipage}{\linewidth}
\textbf{F1 vs Latency}\par\nopagebreak\vspace{-1.5ex}
\begin{center}
\includegraphics[width=0.88\textwidth]{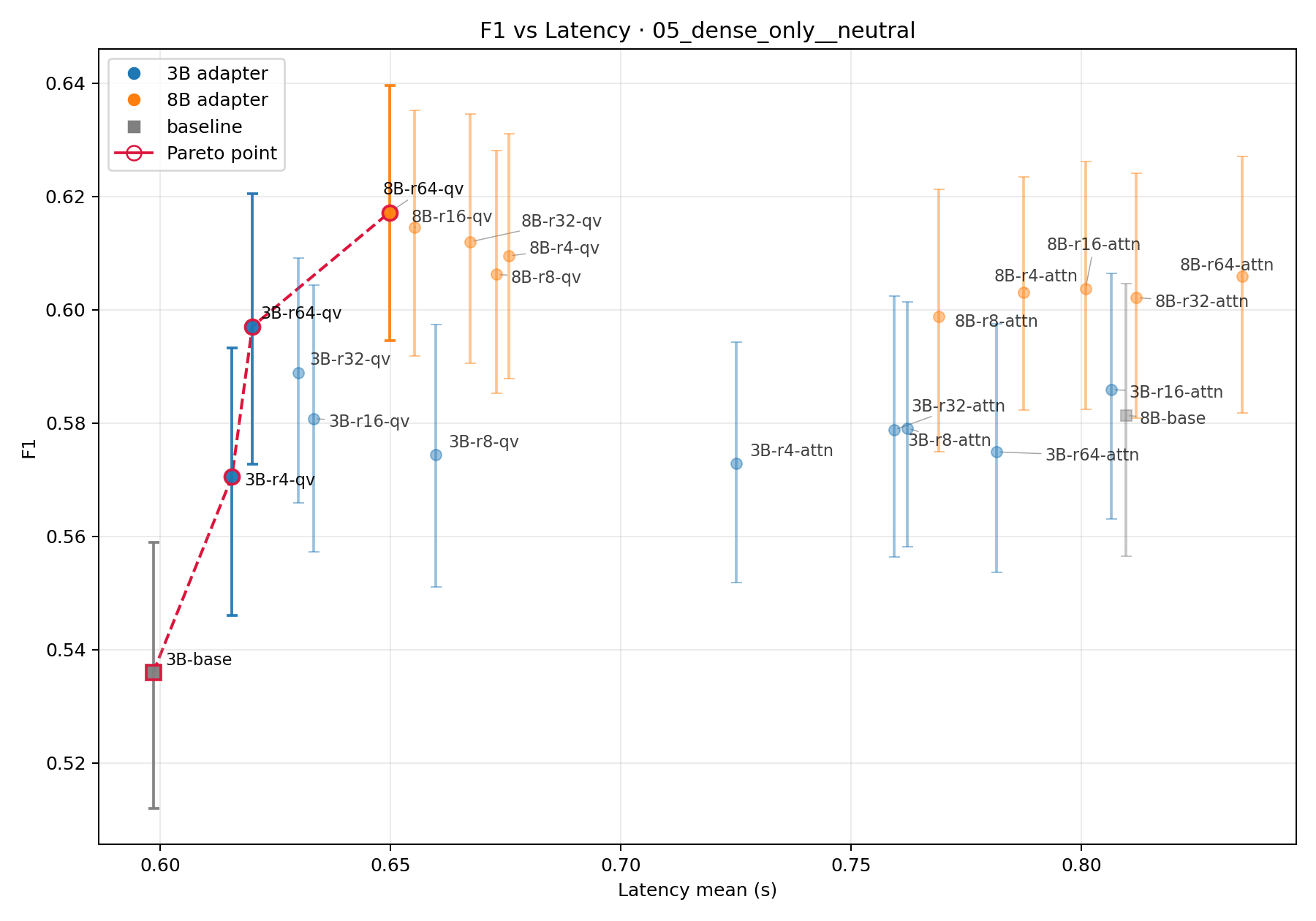}
\end{center}
\end{minipage}

\medskip
\noindent\begin{minipage}{\linewidth}
\textbf{F1 vs Inference VRAM}\par\nopagebreak\vspace{-1.5ex}
\begin{center}
\includegraphics[width=0.88\textwidth]{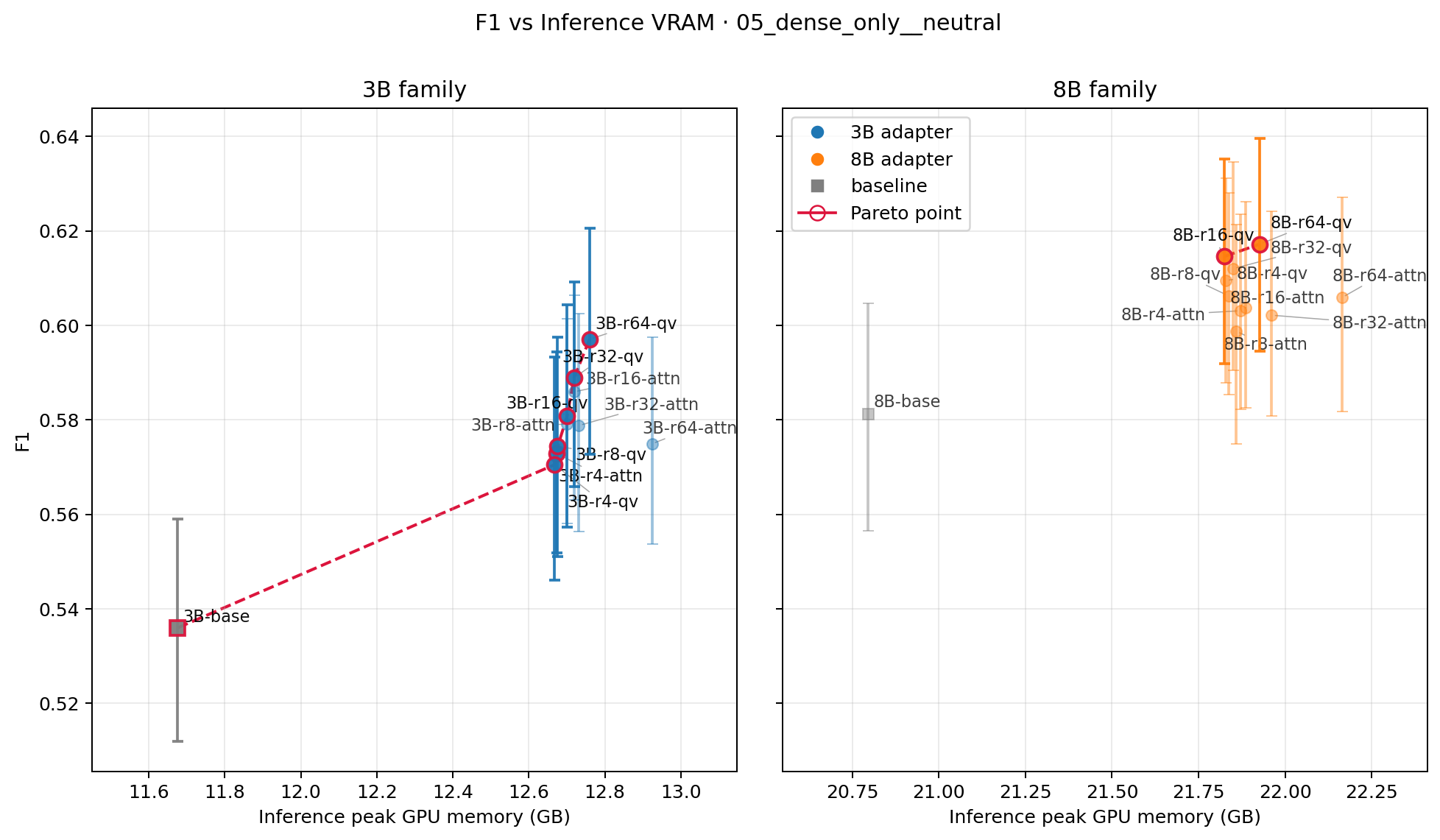}
\end{center}
\end{minipage}

\medskip
\noindent\begin{minipage}{\linewidth}
\textbf{F1 vs Groundedness pass@4}\par\nopagebreak\vspace{-1.5ex}
\begin{center}
\includegraphics[width=0.88\textwidth]{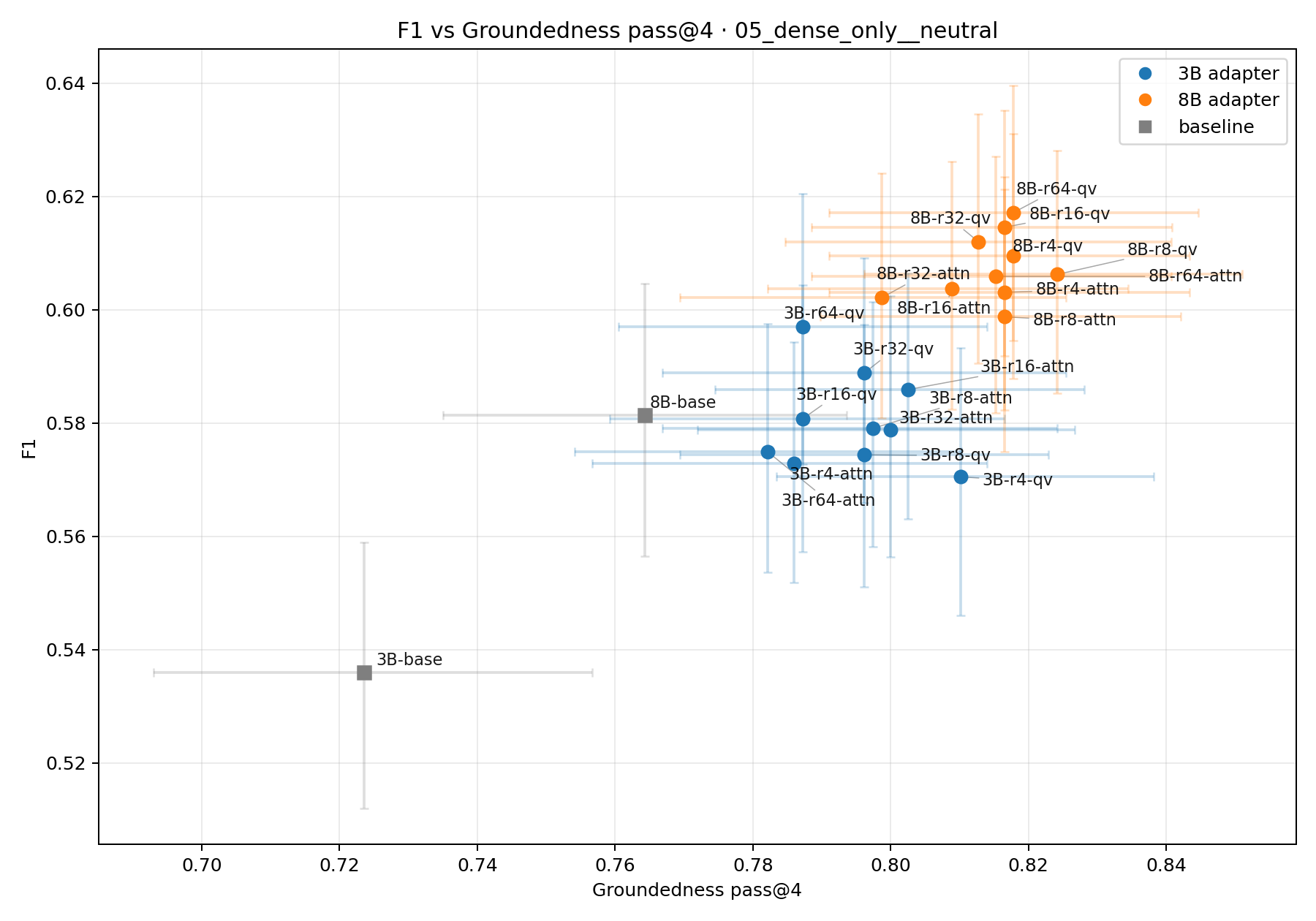}
\end{center}
\end{minipage}

\medskip
\noindent\begin{minipage}{\linewidth}
\textbf{Groundedness pass@4 vs Latency}\par\nopagebreak\vspace{-1.5ex}
\begin{center}
\includegraphics[width=0.88\textwidth]{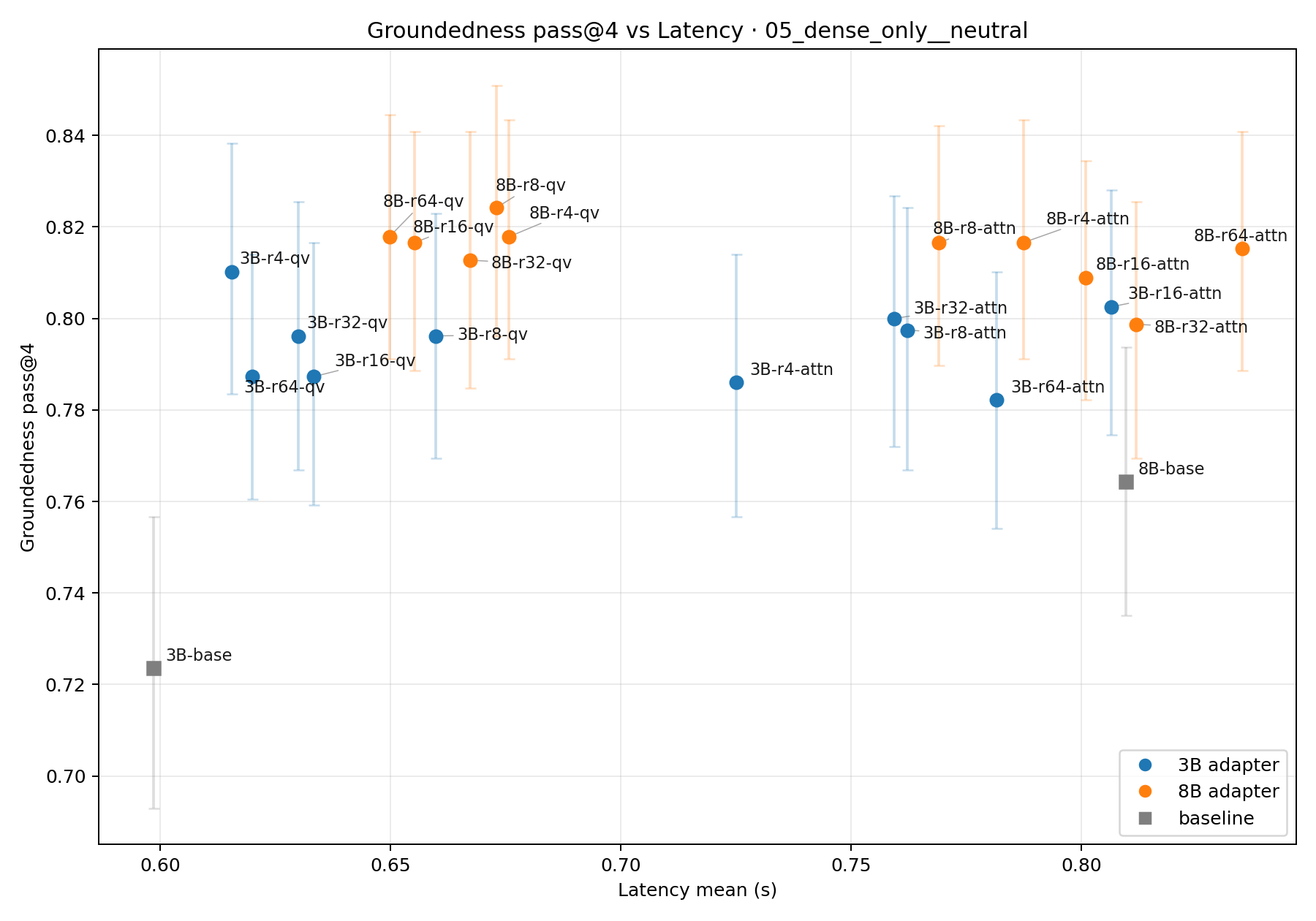}
\end{center}
\end{minipage}

\clearpage
\subsection*{\texttt{06\_dense\_only\_\_explicit\_grounded}}
Regime: \texttt{dense\_only + explicit\_grounded}.

\medskip
\noindent\begin{minipage}{\linewidth}
\textbf{F1 vs Latency}\par\nopagebreak\vspace{-1.5ex}
\begin{center}
\includegraphics[width=0.88\textwidth]{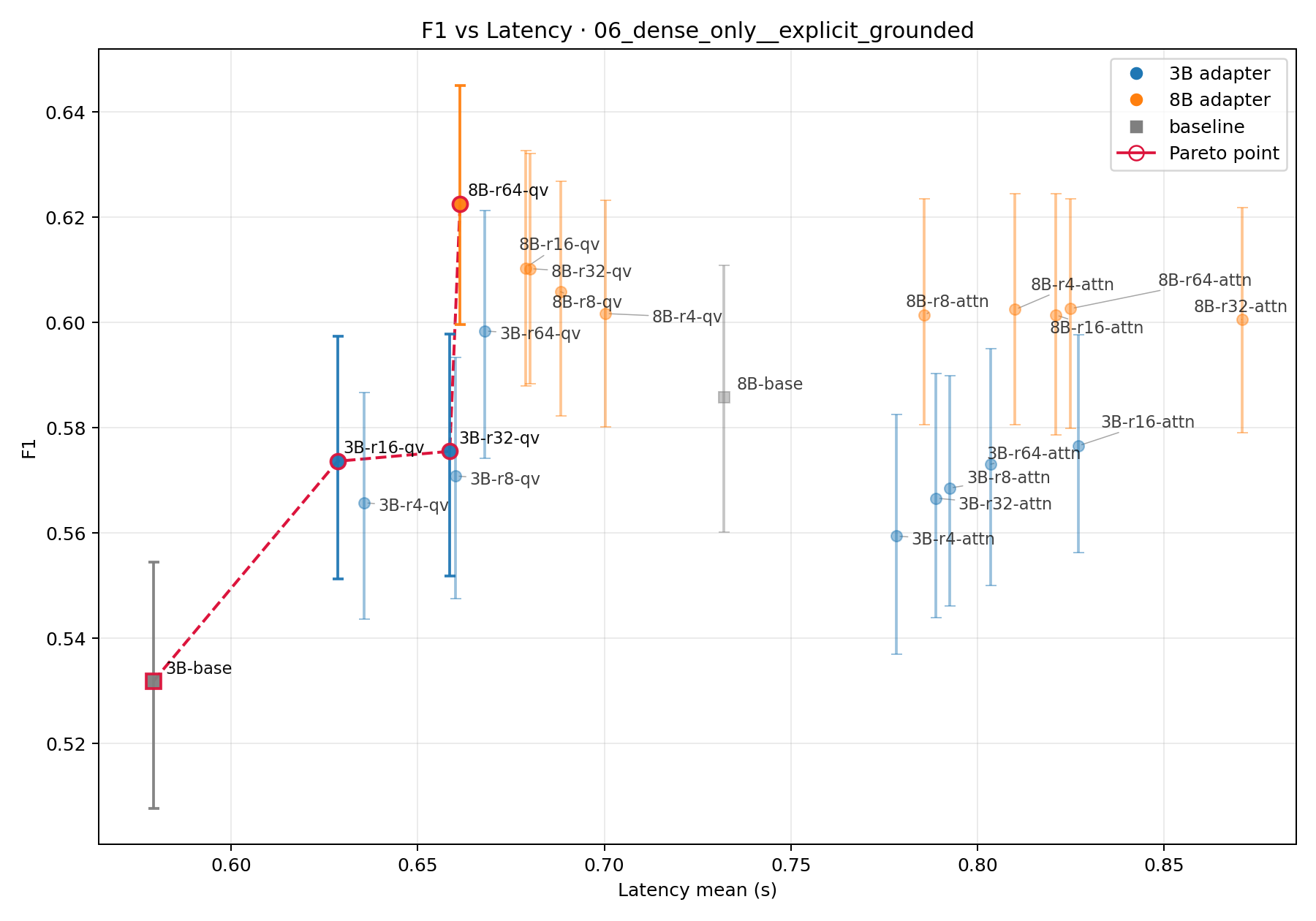}
\end{center}
\end{minipage}

\medskip
\noindent\begin{minipage}{\linewidth}
\textbf{F1 vs Inference VRAM}\par\nopagebreak\vspace{-1.5ex}
\begin{center}
\includegraphics[width=0.88\textwidth]{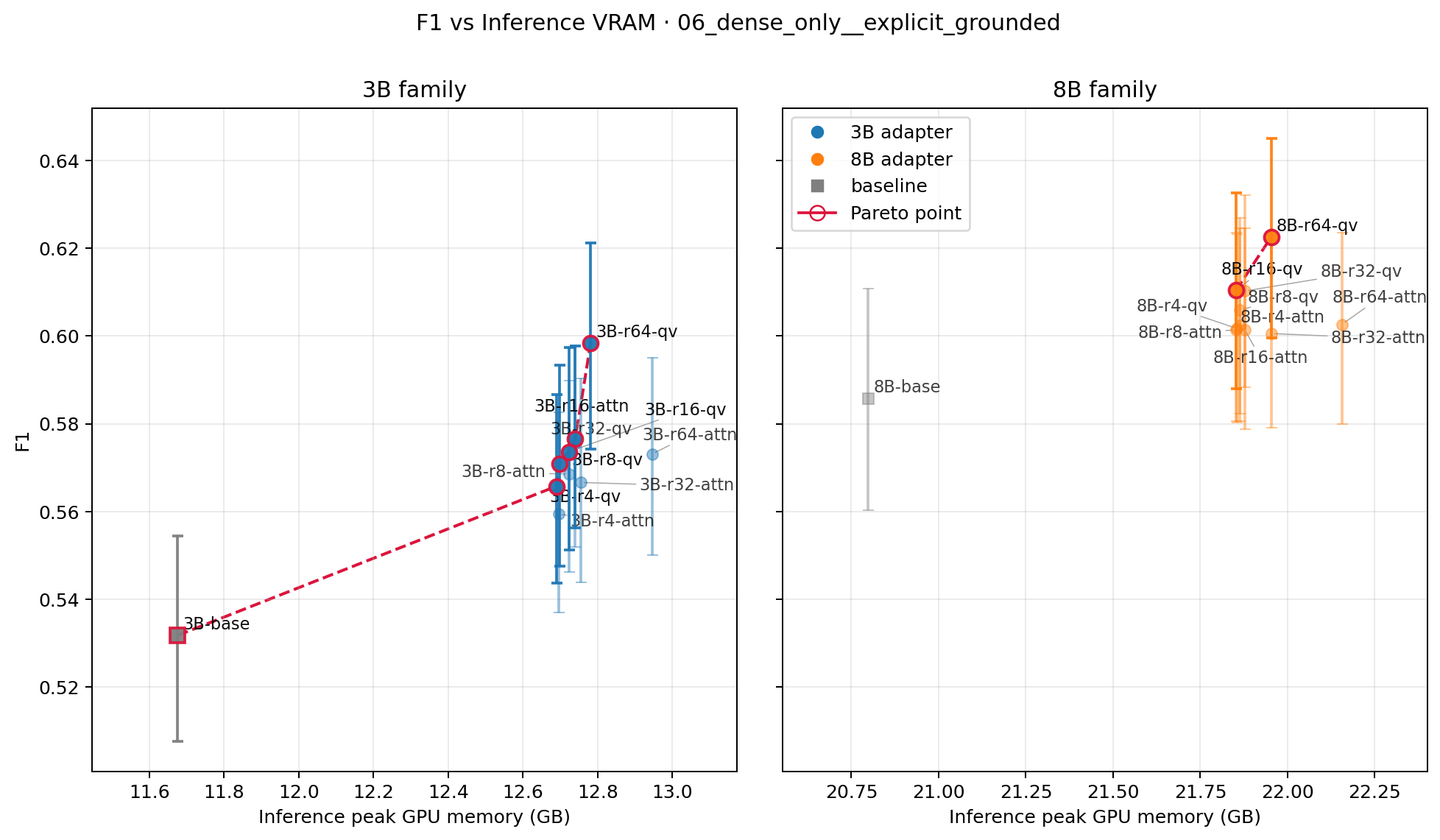}
\end{center}
\end{minipage}

\medskip
\noindent\begin{minipage}{\linewidth}
\textbf{F1 vs Groundedness pass@4}\par\nopagebreak\vspace{-1.5ex}
\begin{center}
\includegraphics[width=0.88\textwidth]{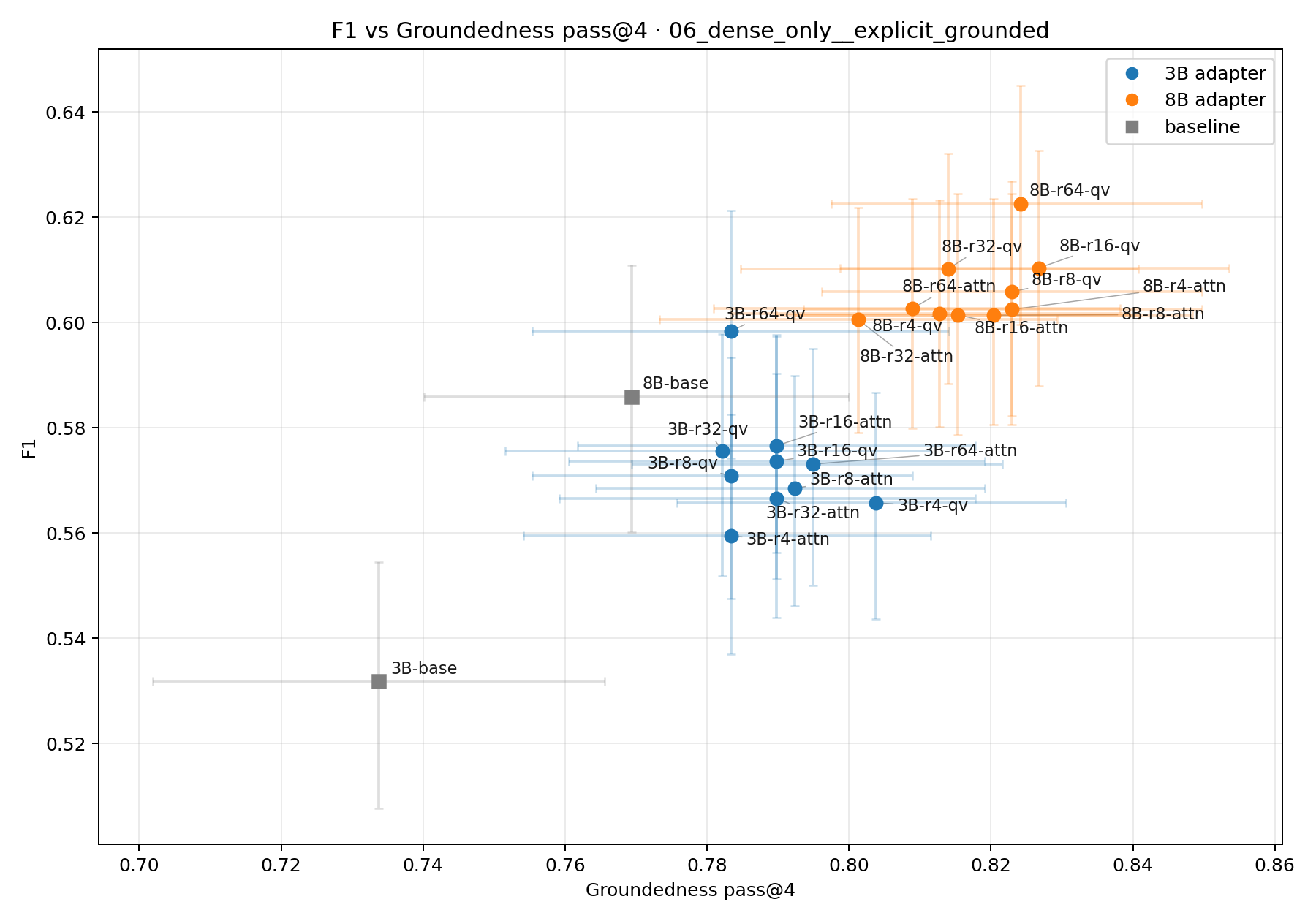}
\end{center}
\end{minipage}

\medskip
\noindent\begin{minipage}{\linewidth}
\textbf{Groundedness pass@4 vs Latency}\par\nopagebreak\vspace{-1.5ex}
\begin{center}
\includegraphics[width=0.88\textwidth]{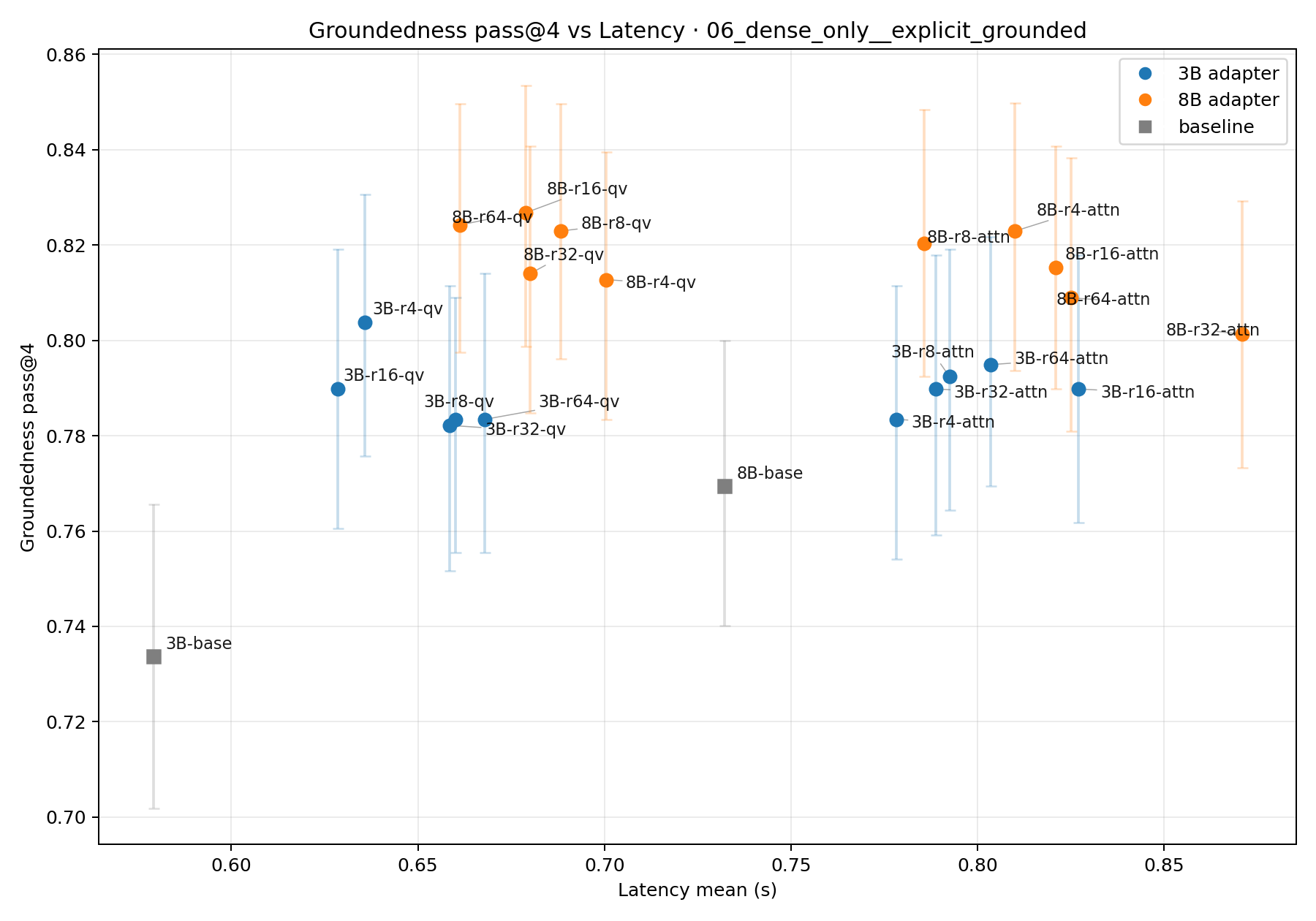}
\end{center}
\end{minipage}

\clearpage
\subsection*{\texttt{07\_sparse\_only\_\_neutral}}
Regime: \texttt{sparse\_only + neutral}.

\medskip
\noindent\begin{minipage}{\linewidth}
\textbf{F1 vs Latency}\par\nopagebreak\vspace{-1.5ex}
\begin{center}
\includegraphics[width=0.88\textwidth]{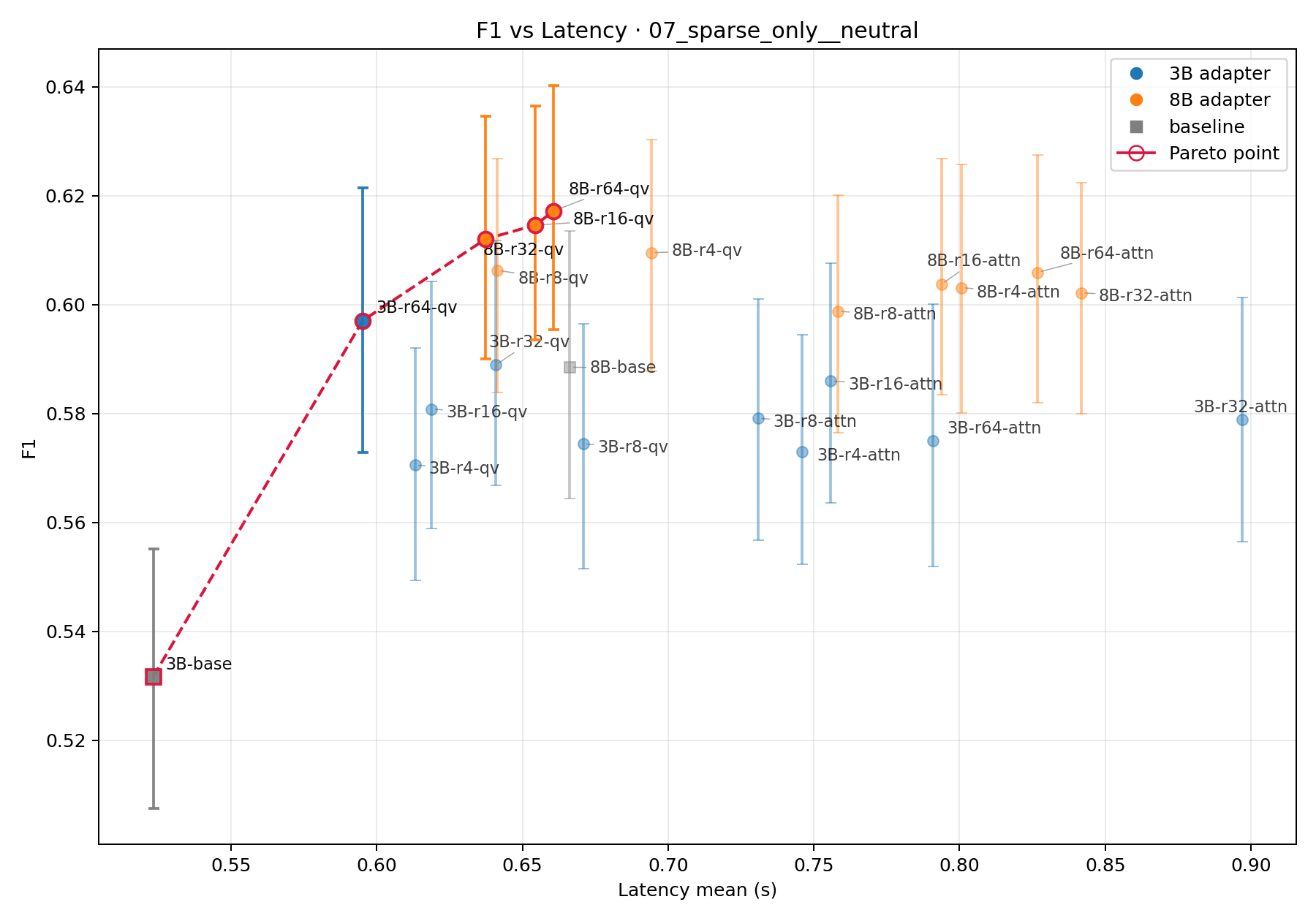}
\end{center}
\end{minipage}

\medskip
\noindent\begin{minipage}{\linewidth}
\textbf{F1 vs Inference VRAM}\par\nopagebreak\vspace{-1.5ex}
\begin{center}
\includegraphics[width=0.88\textwidth]{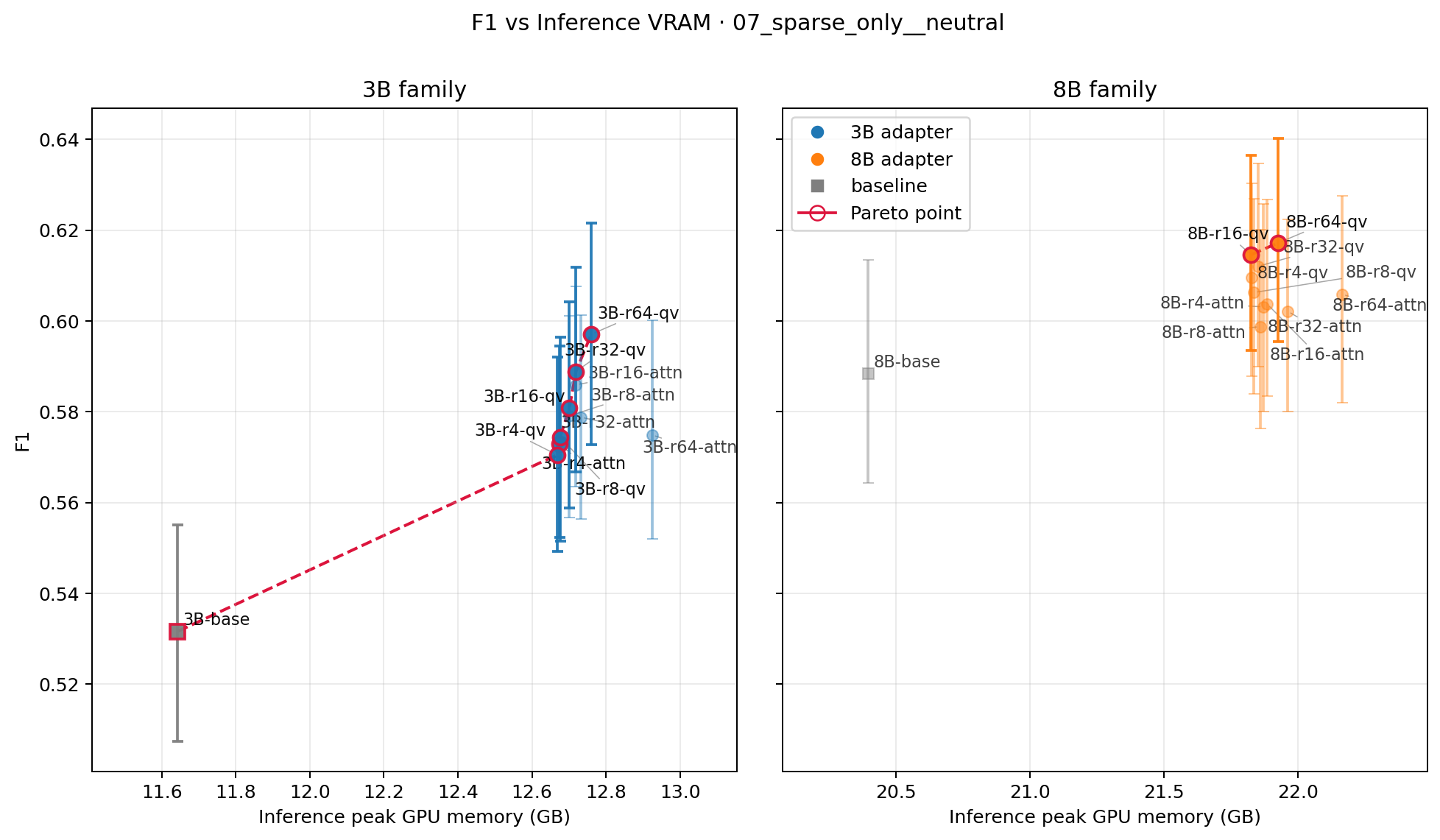}
\end{center}
\end{minipage}

\medskip
\noindent\begin{minipage}{\linewidth}
\textbf{F1 vs Groundedness pass@4}\par\nopagebreak\vspace{-1.5ex}
\begin{center}
\includegraphics[width=0.88\textwidth]{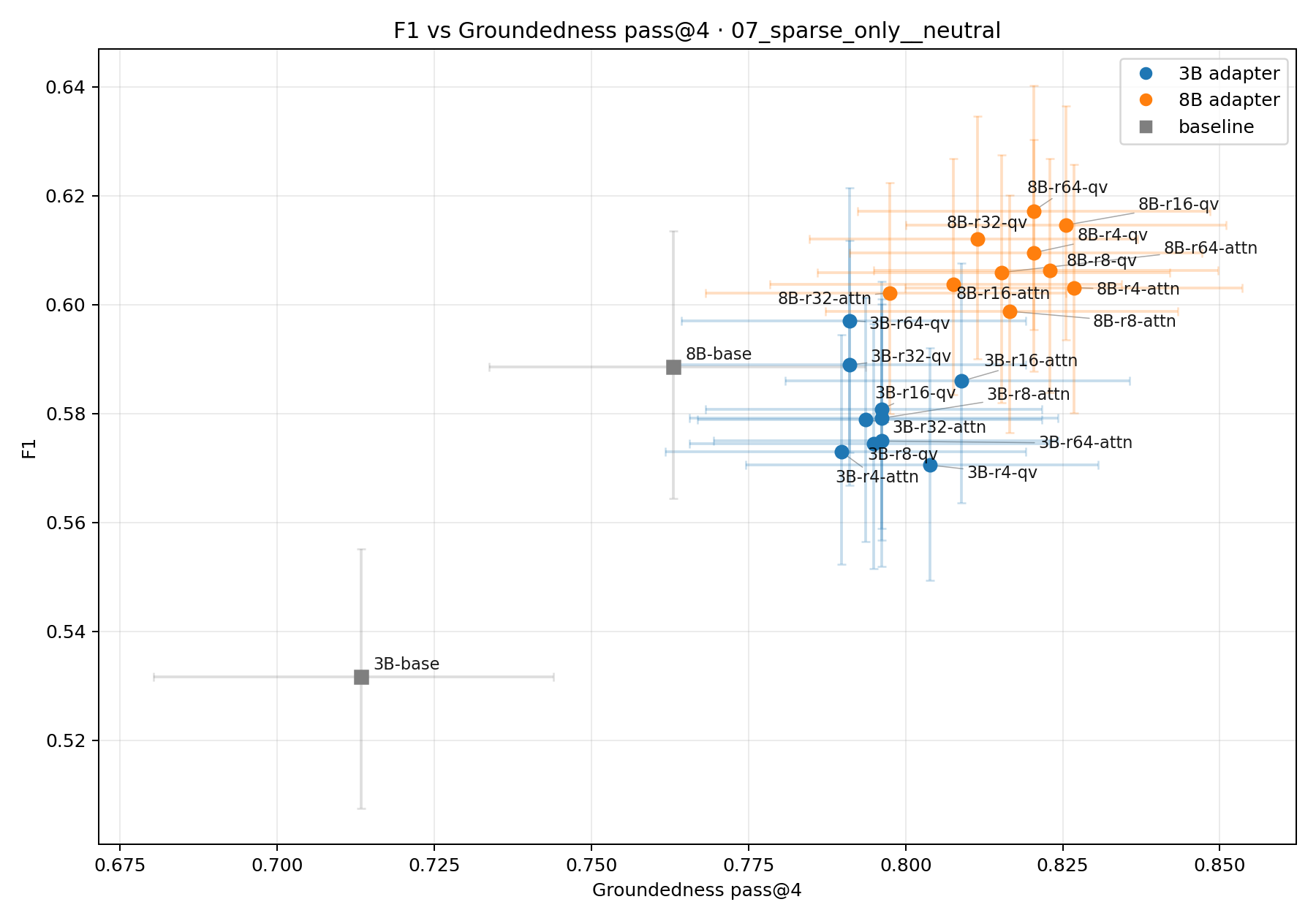}
\end{center}
\end{minipage}

\medskip
\noindent\begin{minipage}{\linewidth}
\textbf{Groundedness pass@4 vs Latency}\par\nopagebreak\vspace{-1.5ex}
\begin{center}
\includegraphics[width=0.88\textwidth]{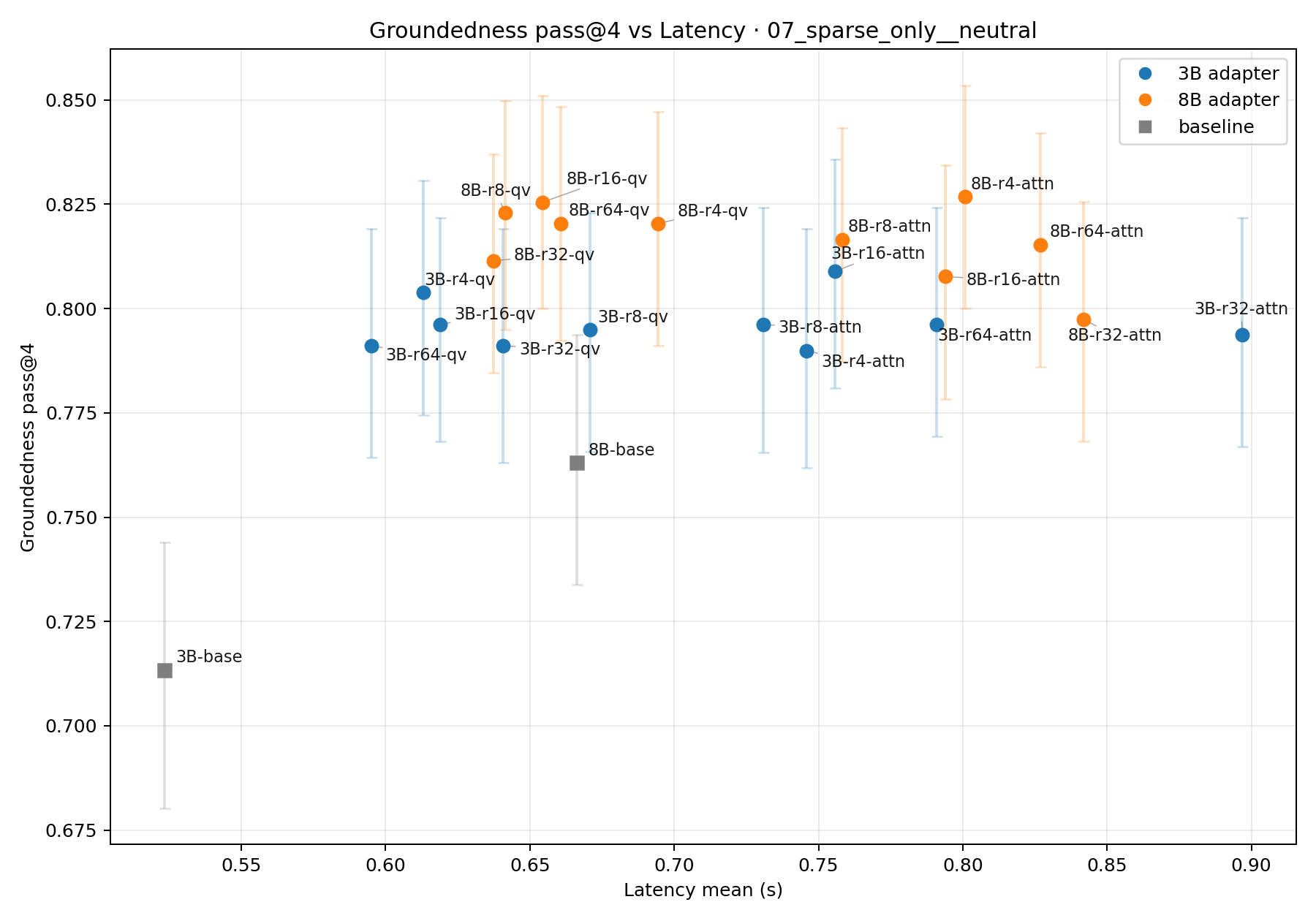}
\end{center}
\end{minipage}

\clearpage
\subsection*{\texttt{08\_sparse\_only\_\_explicit\_grounded}}
Regime: \texttt{sparse\_only + explicit\_grounded}.

\medskip
\noindent\begin{minipage}{\linewidth}
\textbf{F1 vs Latency}\par\nopagebreak\vspace{-1.5ex}
\begin{center}
\includegraphics[width=0.88\textwidth]{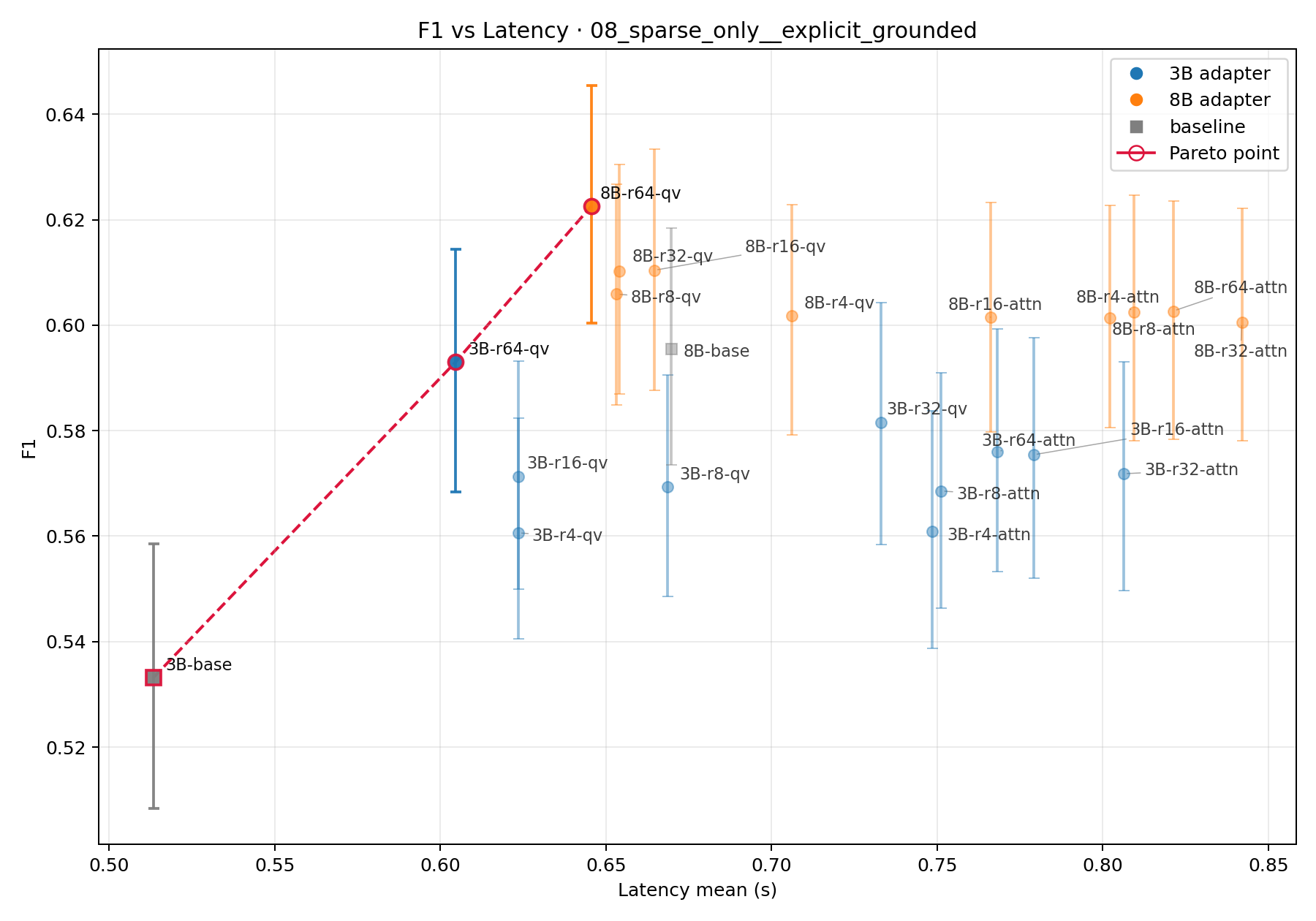}
\end{center}
\end{minipage}

\medskip
\noindent\begin{minipage}{\linewidth}
\textbf{F1 vs Inference VRAM}\par\nopagebreak\vspace{-1.5ex}
\begin{center}
\includegraphics[width=0.88\textwidth]{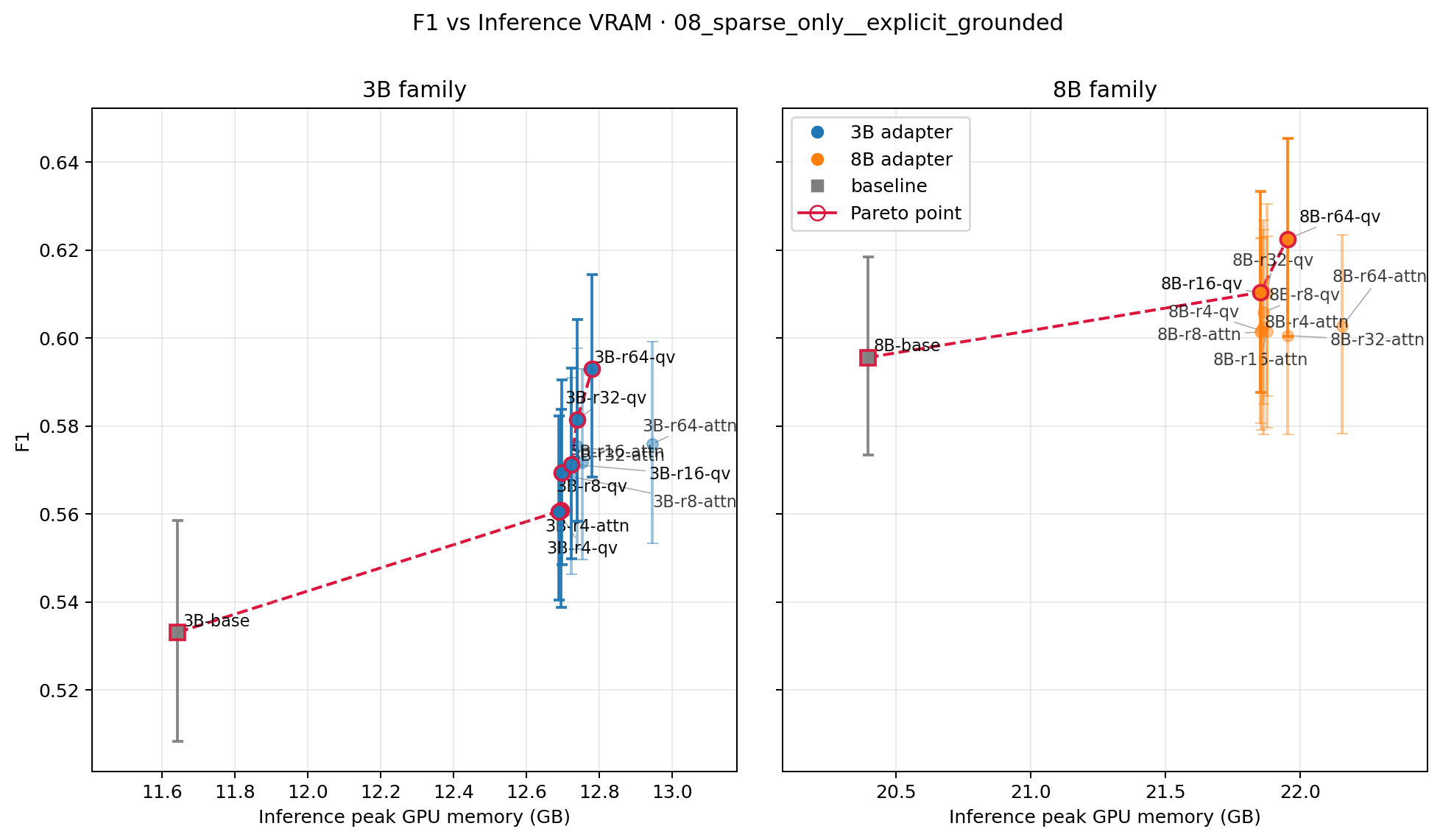}
\end{center}
\end{minipage}

\medskip
\noindent\begin{minipage}{\linewidth}
\textbf{F1 vs Groundedness pass@4}\par\nopagebreak\vspace{-1.5ex}
\begin{center}
\includegraphics[width=0.88\textwidth]{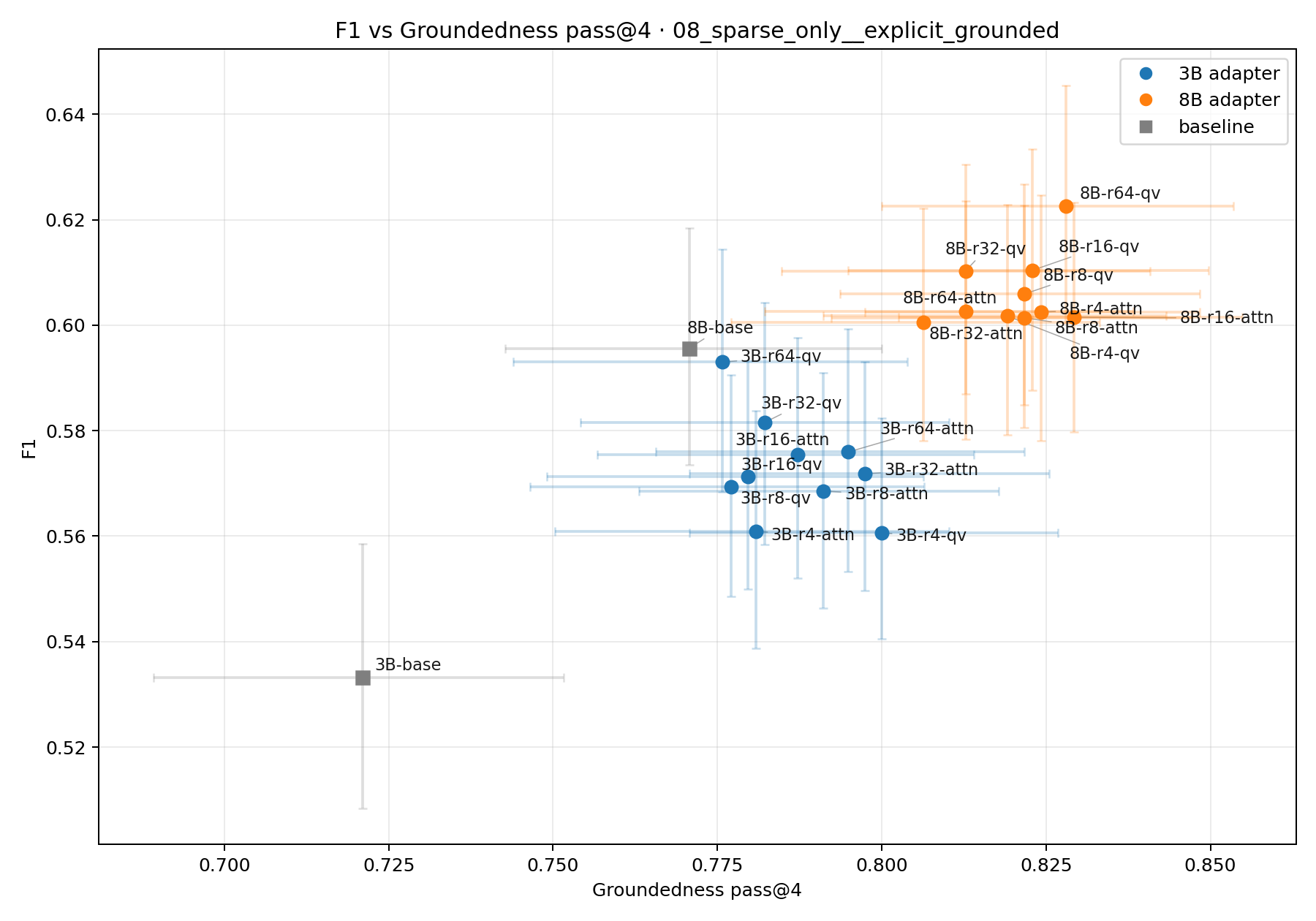}
\end{center}
\end{minipage}

\medskip
\noindent\begin{minipage}{\linewidth}
\textbf{Groundedness pass@4 vs Latency}\par\nopagebreak\vspace{-1.5ex}
\begin{center}
\includegraphics[width=0.88\textwidth]{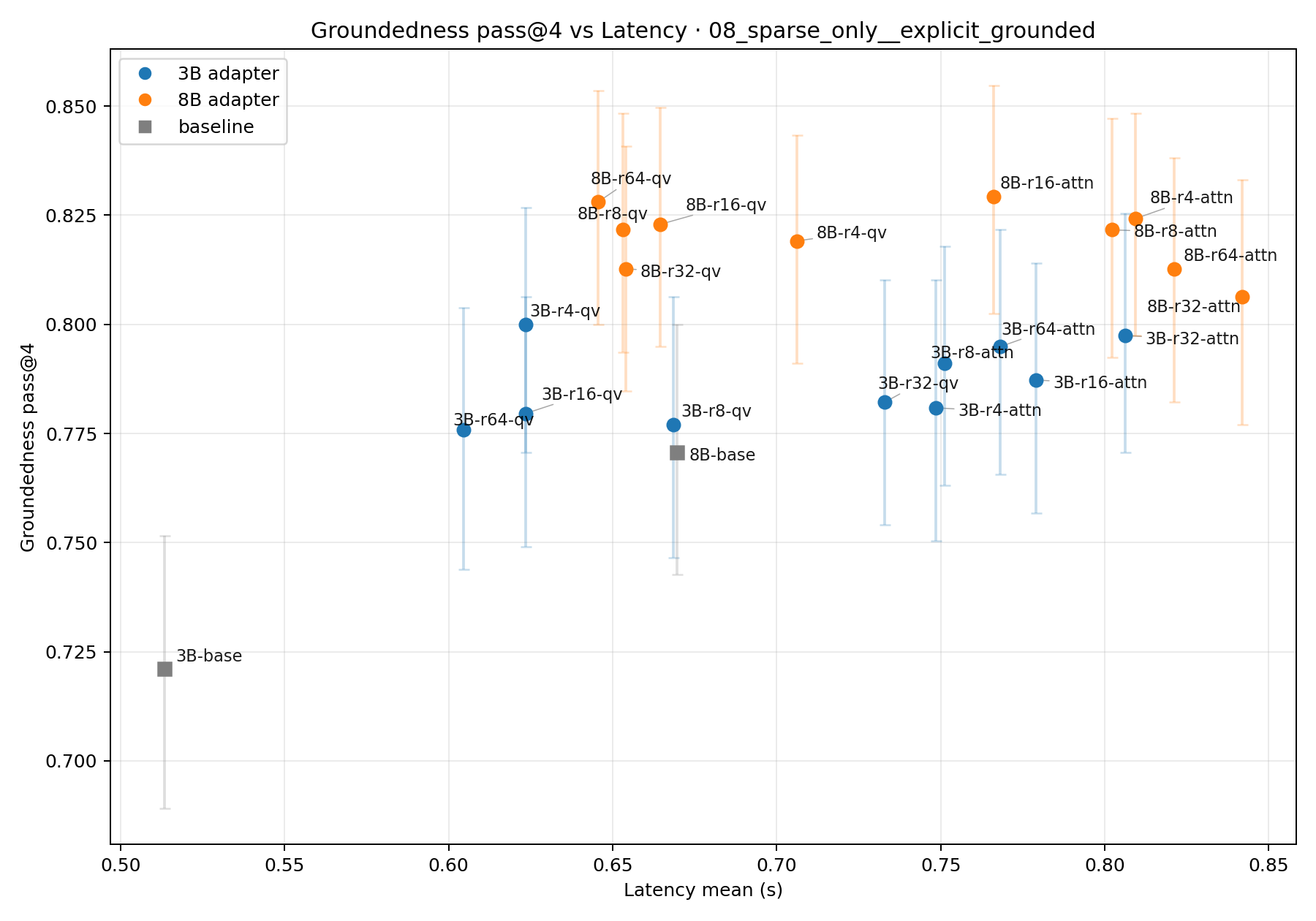}
\end{center}
\end{minipage}

\clearpage
\subsection*{\texttt{09\_hybrid\_bm25\_\_neutral}}
Regime: \texttt{hybrid\_bm25 + neutral}.

\medskip
\noindent\begin{minipage}{\linewidth}
\textbf{F1 vs Latency}\par\nopagebreak\vspace{-1.5ex}
\begin{center}
\includegraphics[width=0.88\textwidth]{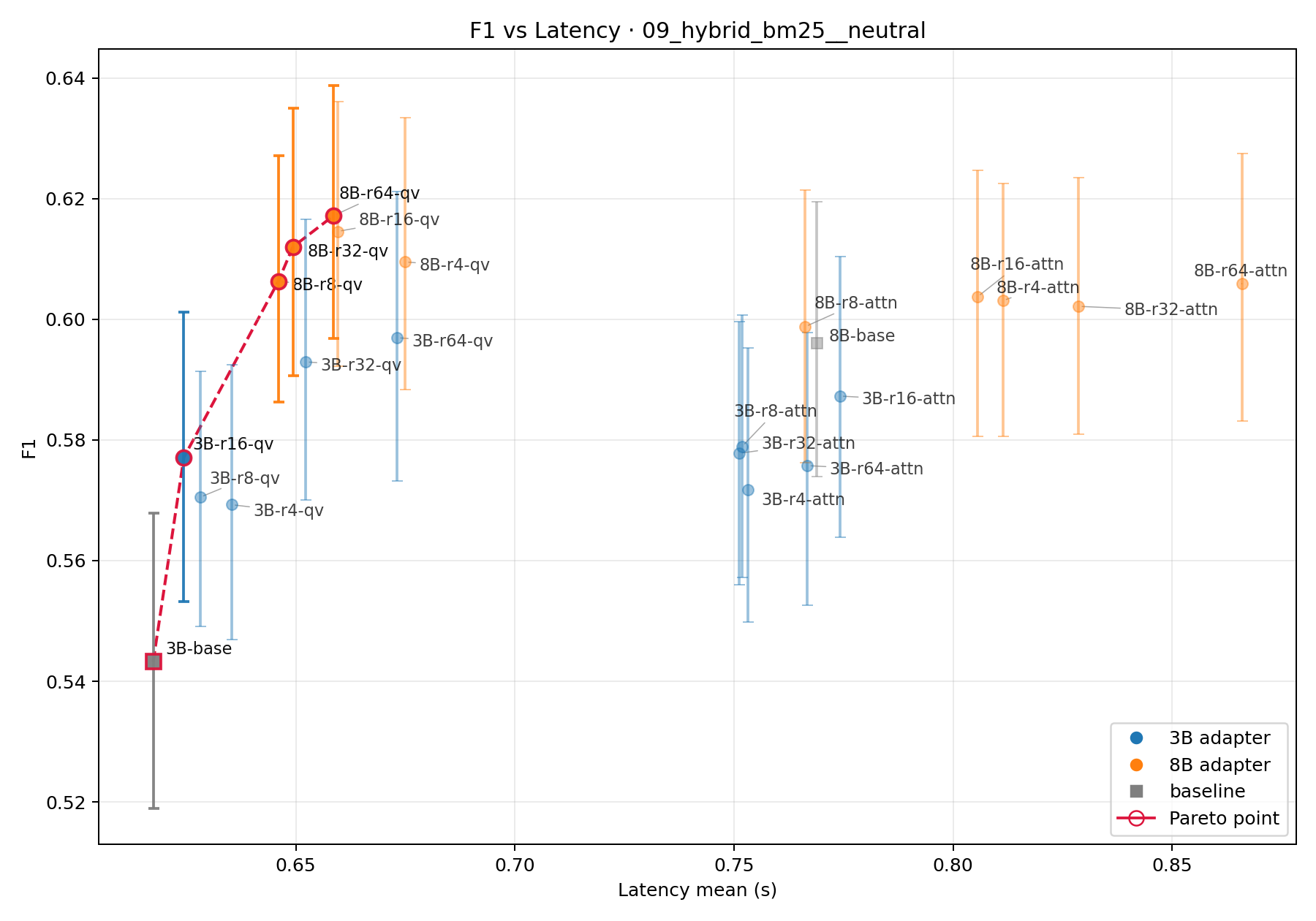}
\end{center}
\end{minipage}

\medskip
\noindent\begin{minipage}{\linewidth}
\textbf{F1 vs Inference VRAM}\par\nopagebreak\vspace{-1.5ex}
\begin{center}
\includegraphics[width=0.88\textwidth]{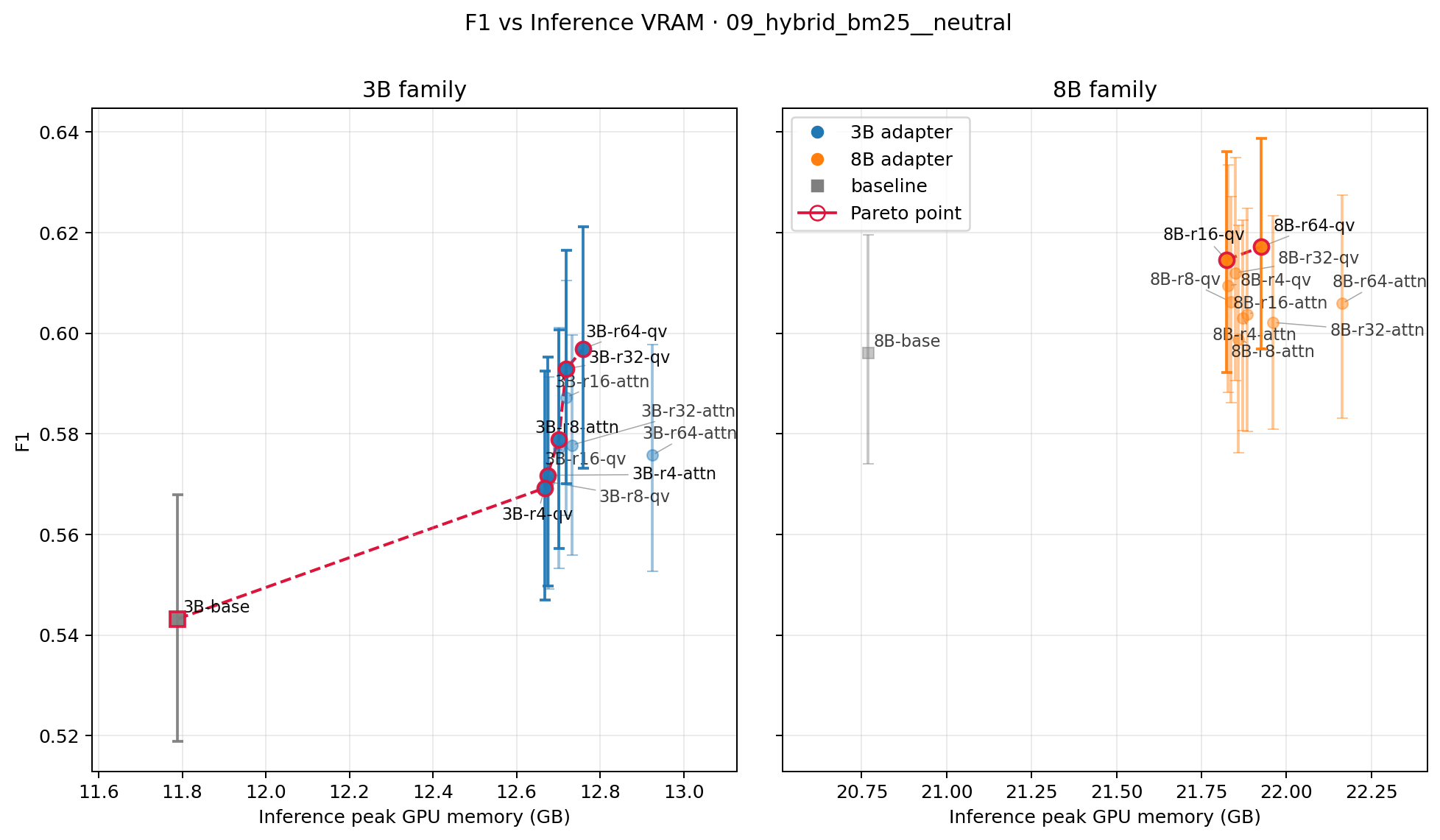}
\end{center}
\end{minipage}

\medskip
\noindent\begin{minipage}{\linewidth}
\textbf{F1 vs Groundedness pass@4}\par\nopagebreak\vspace{-1.5ex}
\begin{center}
\includegraphics[width=0.88\textwidth]{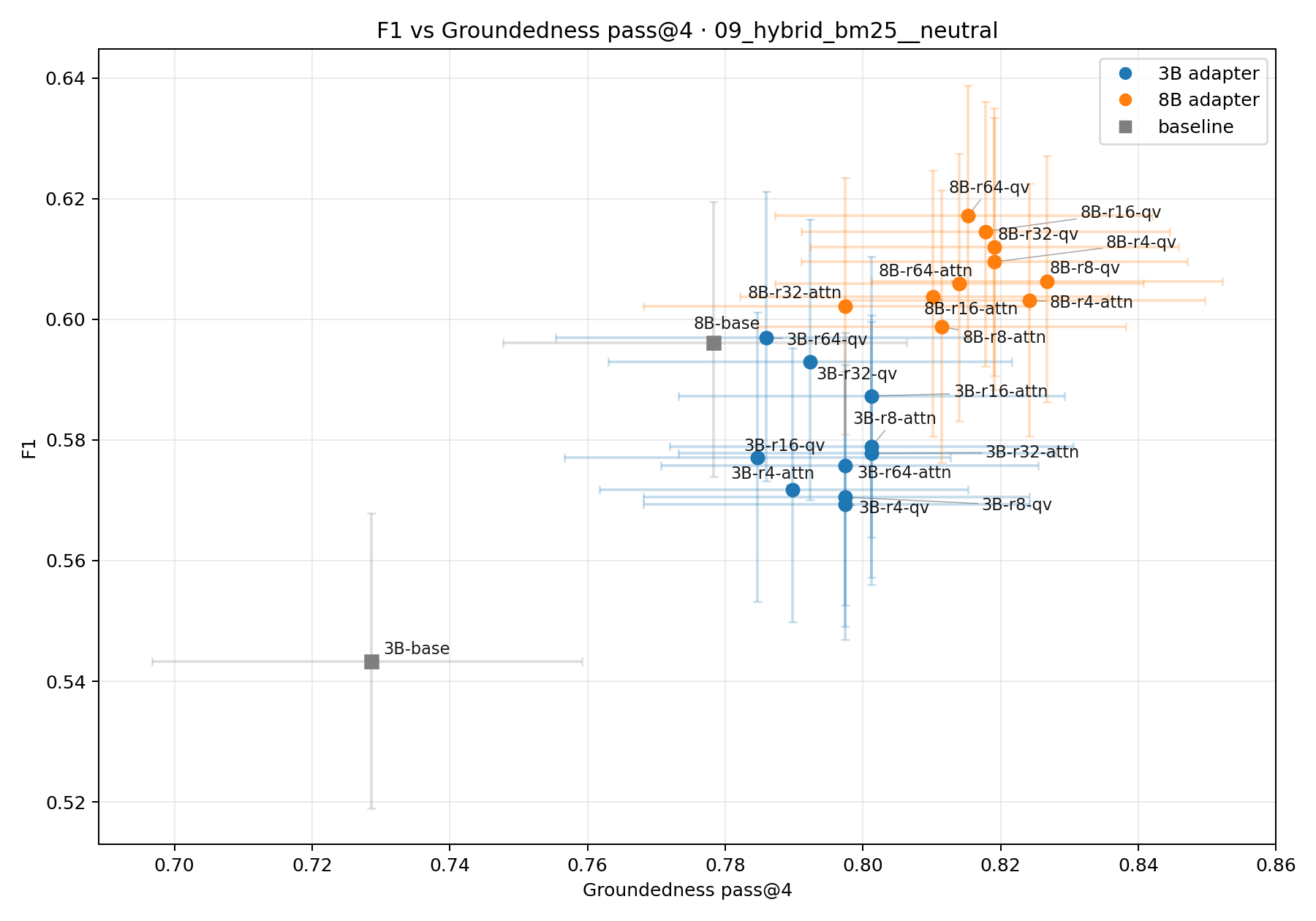}
\end{center}
\end{minipage}

\medskip
\noindent\begin{minipage}{\linewidth}
\textbf{Groundedness pass@4 vs Latency}\par\nopagebreak\vspace{-1.5ex}
\begin{center}
\includegraphics[width=0.88\textwidth]{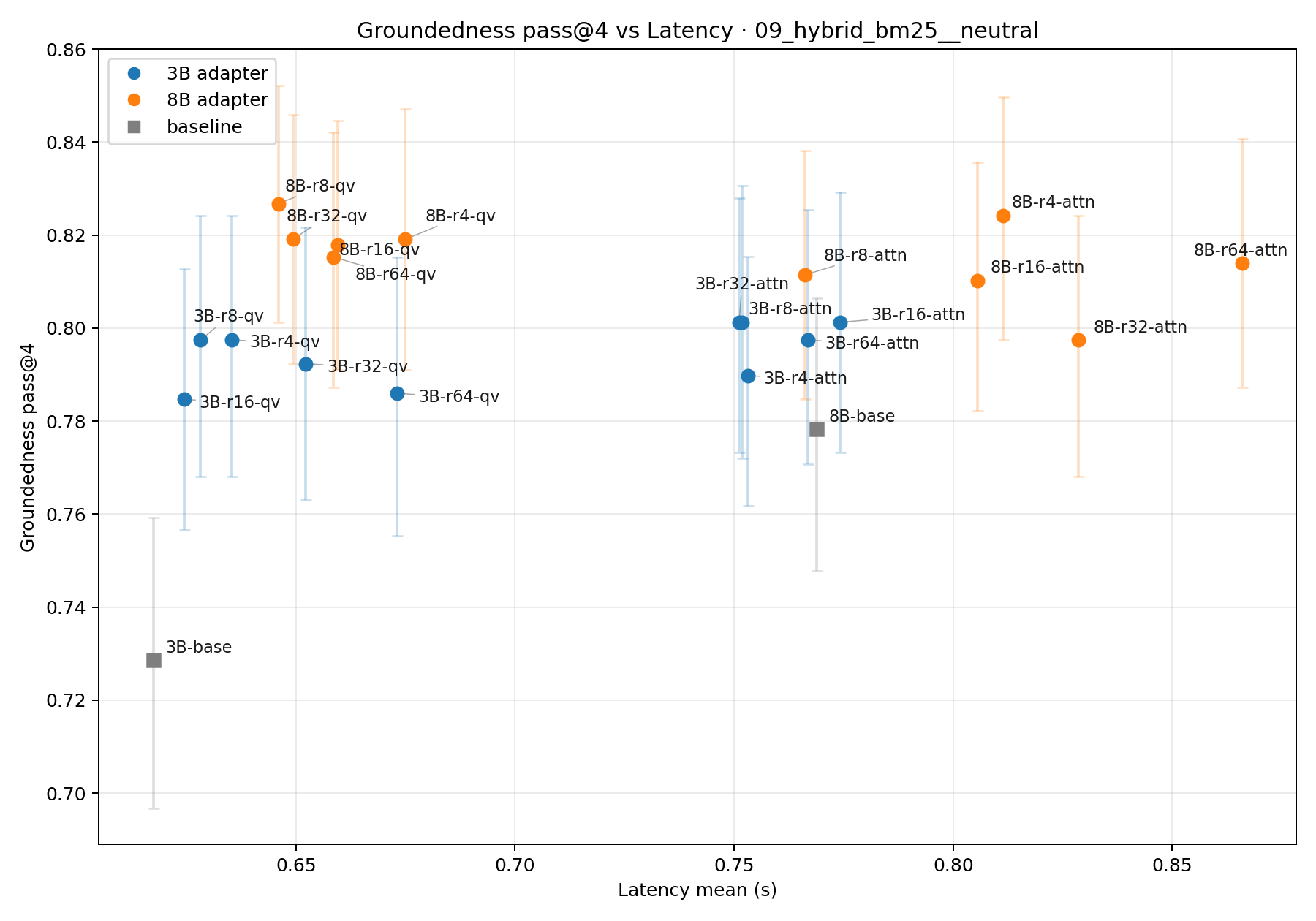}
\end{center}
\end{minipage}

\clearpage
\subsection*{\texttt{10\_hybrid\_bm25\_\_explicit\_grounded}}
Regime: \texttt{hybrid\_bm25 + explicit\_grounded}.

\medskip
\noindent\begin{minipage}{\linewidth}
\textbf{F1 vs Latency}\par\nopagebreak\vspace{-1.5ex}
\begin{center}
\includegraphics[width=0.88\textwidth]{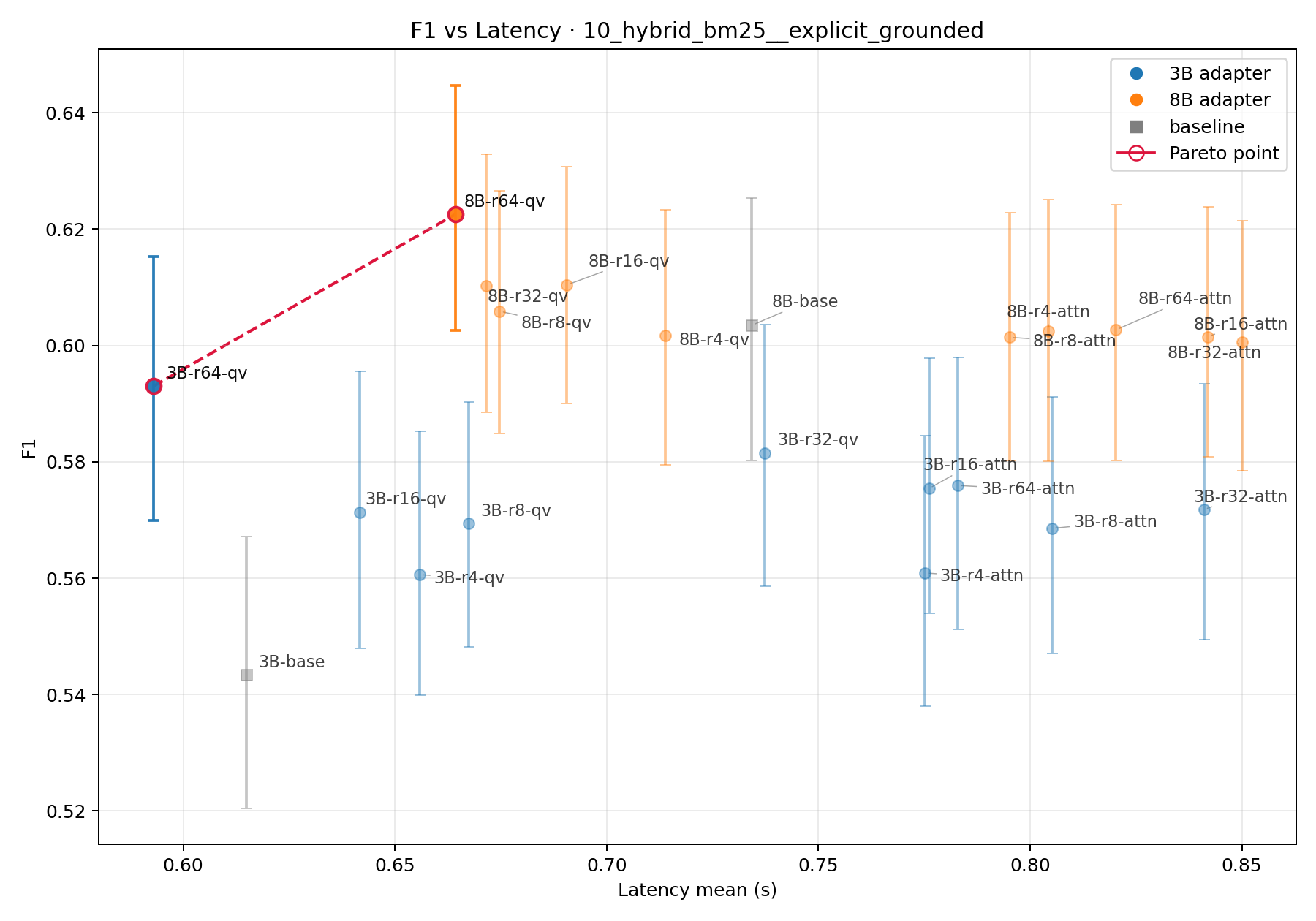}
\end{center}
\end{minipage}

\medskip
\noindent\begin{minipage}{\linewidth}
\textbf{F1 vs Inference VRAM}\par\nopagebreak\vspace{-1.5ex}
\begin{center}
\includegraphics[width=0.88\textwidth]{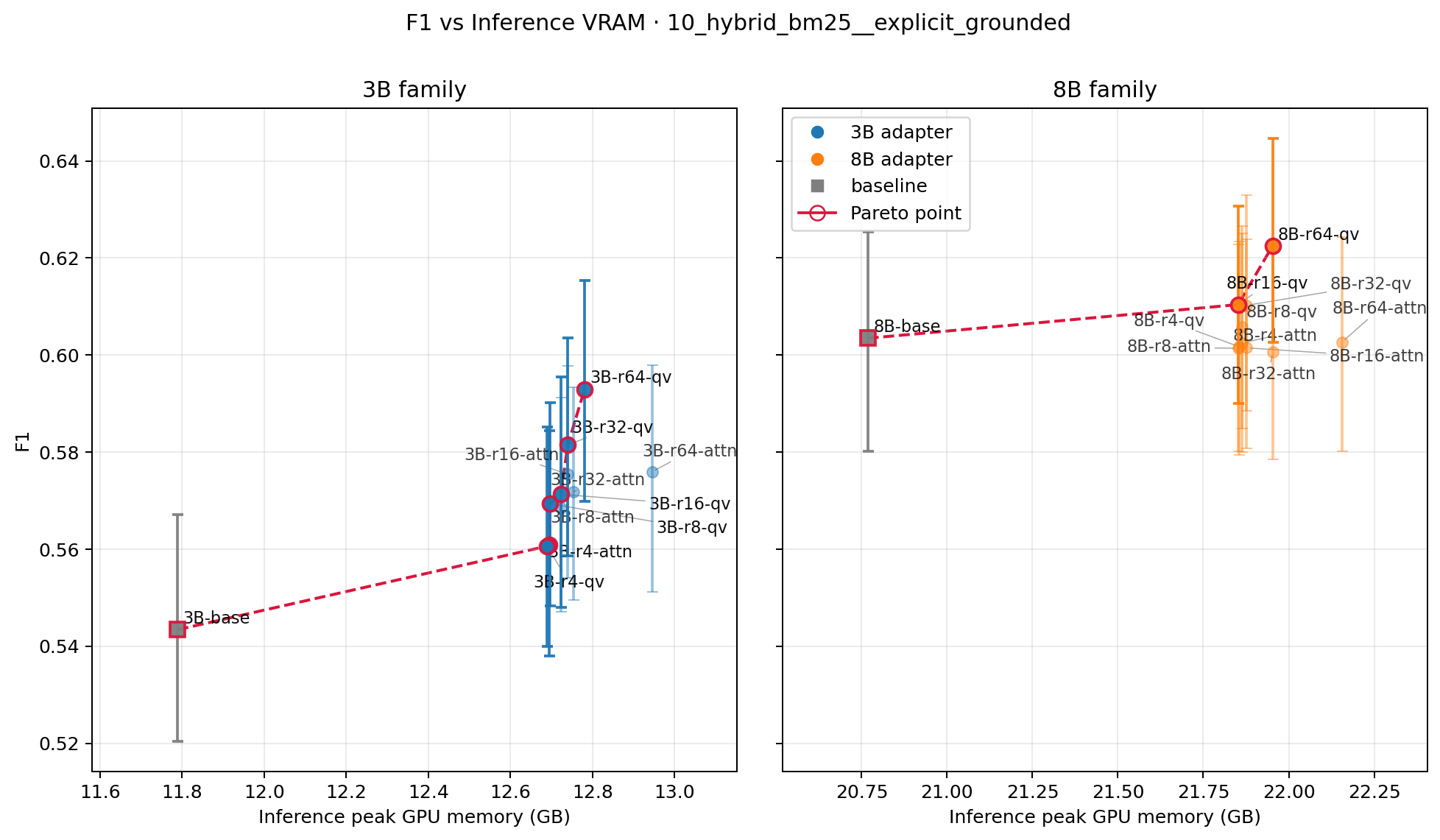}
\end{center}
\end{minipage}

\medskip
\noindent\begin{minipage}{\linewidth}
\textbf{F1 vs Groundedness pass@4}\par\nopagebreak\vspace{-1.5ex}
\begin{center}
\includegraphics[width=0.88\textwidth]{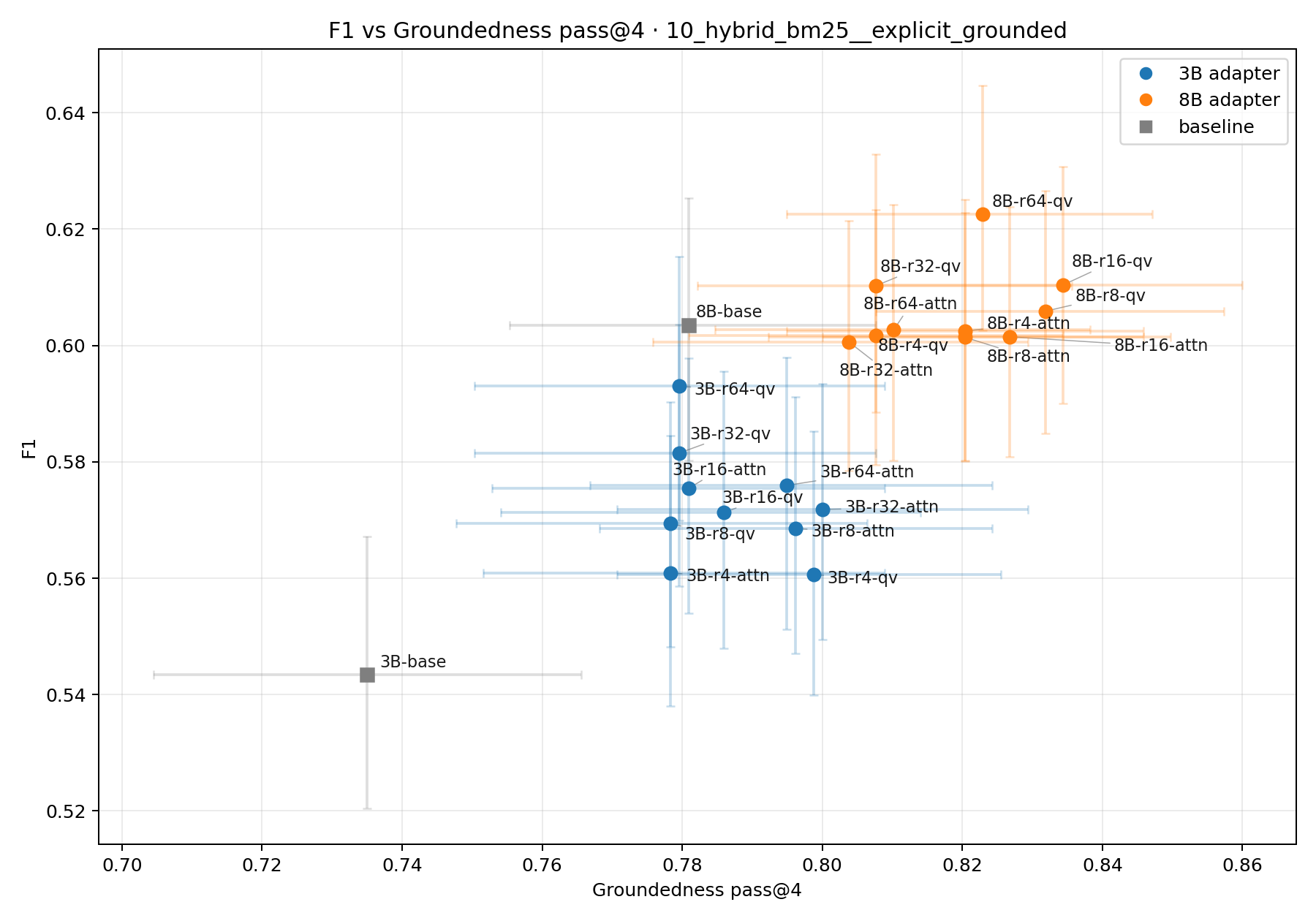}
\end{center}
\end{minipage}

\medskip
\noindent\begin{minipage}{\linewidth}
\textbf{Groundedness pass@4 vs Latency}\par\nopagebreak\vspace{-1.5ex}
\begin{center}
\includegraphics[width=0.88\textwidth]{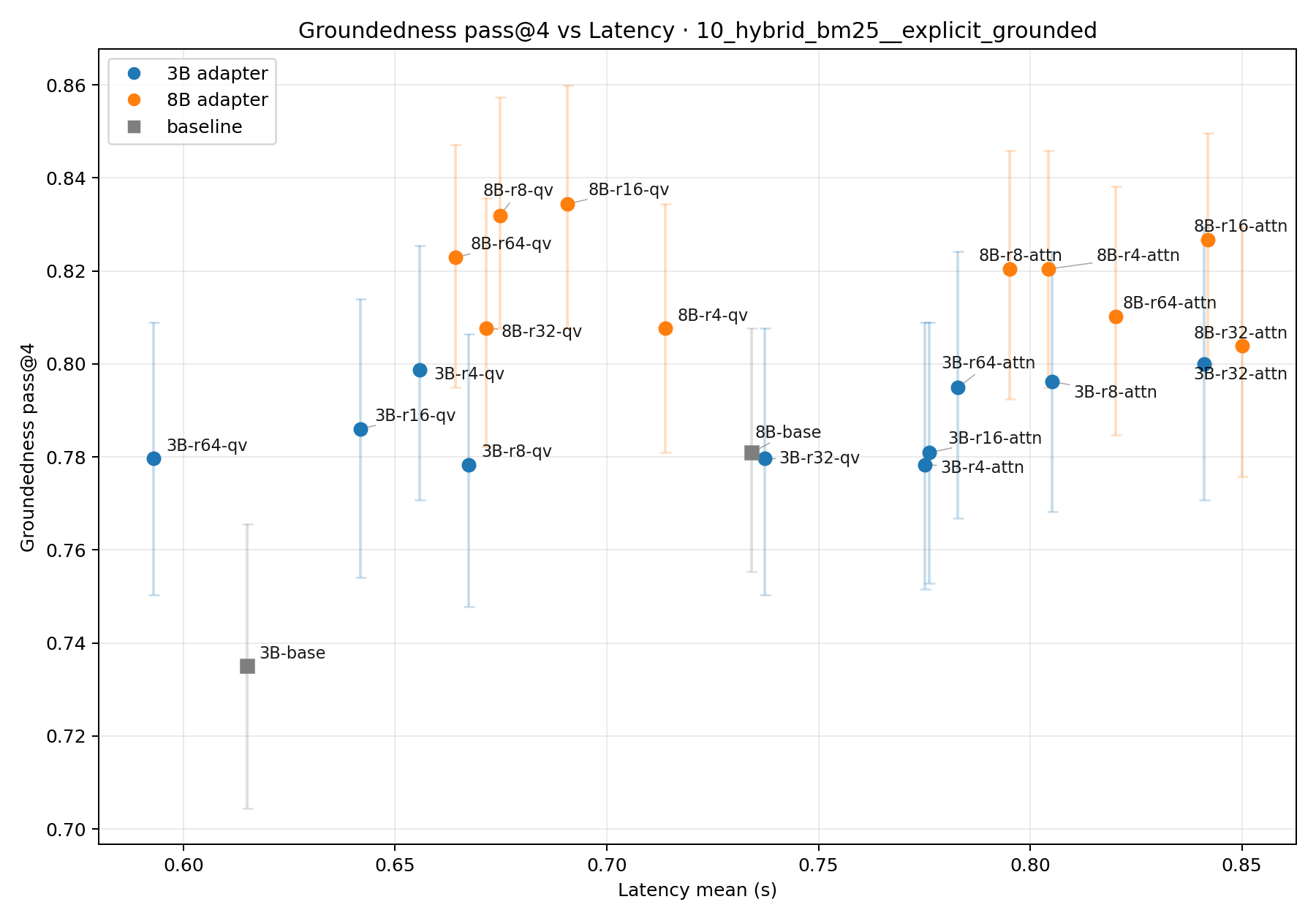}
\end{center}
\end{minipage}

\end{document}